\documentclass[10pt,twocolumn,letterpaper]{article}

\usepackage{cvpr}              

\usepackage{times}
\usepackage{epsfig}
\usepackage{graphicx}
\usepackage{amsmath}
\usepackage{amssymb}
\usepackage{tabularx}
\usepackage{mathtools}
\usepackage{algorithm}
\usepackage{algpseudocode}
\usepackage{bm}
\usepackage{wrapfig}
\usepackage[accsupp]{axessibility} 

\usepackage{multirow}
\usepackage{makecell, tabularx}
\newcolumntype{Y}{>{\centering\arraybackslash}X}
\usepackage{rotating}
\usepackage[export]{adjustbox}
\usepackage{pgfplots}
\pgfplotsset{compat=1.18}

\definecolor{cvprblue}{rgb}{0.21,0.49,0.74}
\usepackage[pagebackref,breaklinks,colorlinks,citecolor=cvprblue]{hyperref}

\def\paperID{} 
\def\confName{CVPR}
\def\confYear{2024}

\title{PAPR in Motion: Seamless Point-level 3D Scene Interpolation
}

\author{Shichong Peng, Yanshu Zhang, Ke Li \\
APEX Lab\\
School of Computing Science\\
Simon Fraser University \\
{\tt\small \{shichong\_peng,yanshu\_zhang,keli\}@sfu.ca}}

\begin{document}

\maketitle
\begin{abstract}
We propose the problem of point-level 3D scene interpolation, which aims to simultaneously reconstruct a 3D scene in two states from multiple views, synthesize smooth point-level interpolations between them, and render the scene from novel viewpoints, all without any supervision between the states. The primary challenge is on achieving a smooth transition between states that may involve significant and non-rigid changes. To address these challenges, we introduce ``PAPR in Motion'', a novel approach that builds upon the recent Proximity Attention Point Rendering (PAPR) technique, which can deform a point cloud to match a significantly different shape and render a visually coherent scene even after non-rigid deformations. Our approach is specifically designed to maintain the temporal consistency of the geometric structure by introducing various regularization techniques for PAPR. The result is a method that can effectively bridge large scene changes and produce visually coherent and temporally smooth interpolations in both geometry and appearance. Evaluation across diverse motion types demonstrates that ``PAPR in Motion'' outperforms the leading neural renderer for dynamic scenes. For more results and code, please visit our 
project website at \footnotesize \url{https://niopeng.github.io/PAPR-in-Motion/}
\normalsize.

\end{abstract}    
\section{Introduction}
\label{sec:intro}

\begin{figure*}[ht]
    \centering
    \includegraphics[width=\linewidth]{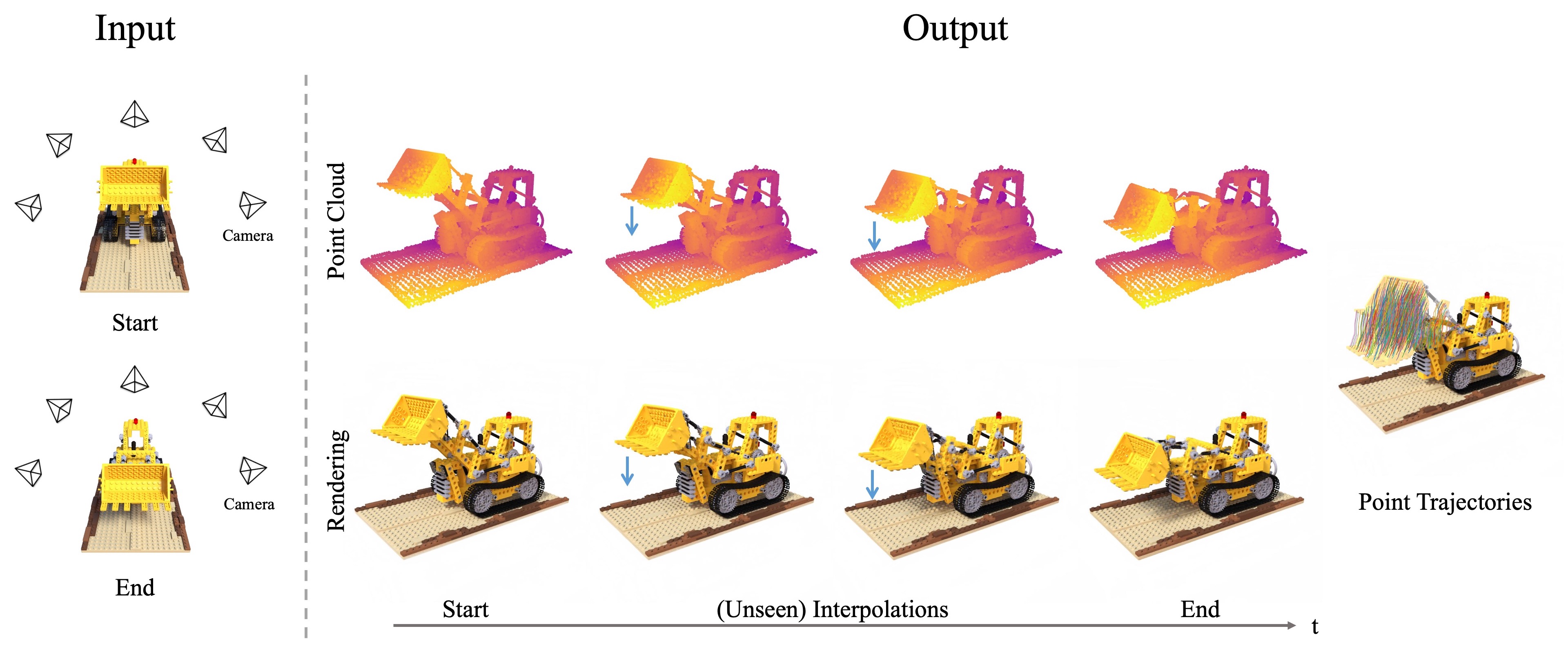}
    \caption{
    We introduce the novel task of point-level 3D scene interpolation: Given multi-view RGB images of a scene at two distinct states, the start state and the end state, our goal is to achieve a seamless point-level interpolation between these states without any intermediate supervision. Our method, PAPR in Motion, can effectively synthesize a plausible geometry for any intermediate state, represented by a point cloud (top row), and renders the corresponding scene appearance (bottom row). Additionally, we provide a visualization of the point interpolation trajectories (right), which illustrate the coherent motion synthesized within the scene. }
    \label{fig:what_fig}
\end{figure*}

In computer vision and graphics, it is commonly desired to visualize how a scene might evolve smoothly from a state to a very different state. If this can be done, it could open the way to many possibilities. For instance, in animation and filmmaking, the ability to create lifelike animations from just two observed states could greatly simplify production processes and reduce the time and effort required compared to traditional frame-by-frame animation.

To tackle this problem, we introduce a novel task, \emph{point-level 3D scene interpolation}. Given observations of a 3D scene at only two states from multiple views, our objective is to synthesize a smooth and plausible interpolation between scene points at the different states, so that the scene can be rendered from novel views at any intermediate state, as shown in Figure~\ref{fig:what_fig}. No observation of any intermediate state is assumed. Our task differs from traditional static scene reconstruction~\cite{kerbl20233d,Mildenhall2020NeRFRS,Zuo2022ViewSW,Yu2021PlenoxelsRF,Lassner2021PulsarES,Xu2022PointNeRFPN}, which does not handle scene changes, and also from novel view synthesis of dynamic 3D scenes~\cite{Pumarola2020DNeRFNR,Park2020NerfiesDN,Liu2023RobustDR,Park2021HyperNeRF}, which assumes the observation of the scene at all times as it changes.

Our task carries unique challenges beyond those of novel view synthesis on static or dynamic scenes. Because intermediate states are not observed, the scene geometry can change substantially between the observed states. The geometry changes can be highly complex and involve different local deformations that may be non-rigid. Additionally, these geometry changes can lead to appearance changes, for example in the cast shadows and specular reflections. 

To tackle this task, we need to be able to represent the large geometry changes between the states accurately. We can view such changes as motion. Different representations of motion are suitable for capturing different types of scene changes. Eulerian motion representations record changes at fixed scene locations, while Lagrangian motion representations track specific points through space and time. Unlike Eulerian representations, which estimate changes independently at different time steps, Lagrangian representations jointly estimate changes for the same point tracked across time. As a result, Lagrangian representations can model long-range dynamics more accurately. Since there may be substantial geometry changes between the observed states in our task, we opt for modelling scene changes with the Lagrangian approach. 
To enable this, we leverage recent advances in point-based neural renderers, which are well-suited to representing Lagrangian motion through the point-based scene representations they learn. At a high level, our method involves first training a model on the scene's initial state and then finetuning it to reconstruct the scene in its final state. The intermediate iterations of this finetuning process effectively create interpolations between the two states.

To ensure seamless and realistic interpolations, the choice of point rendering technique is crucial. We leverage the Proximity Attention Point Rendering (PAPR)~\cite{zhang2023papr} technique, which offers two key advantages: (i) its ability to deform a point cloud to match a significantly different shape, crucial for adapting to substantial scene changes; and (ii) its ability to maintain shape contiguity and visual coherence after geometry modifications such as non-rigid deformations, ensuring realistic novel view synthesis of the scene at all interpolated states.

Our method, which we dub ``PAPR in Motion'', harnesses the strengths of PAPR and incorporates regularization techniques specifically designed for our problem. We leverage the observation that nearby points tend to move together and design regularizers for PAPR that encourage local rigidity and uniform point motion. We show that our method produces smooth and realistic interpolations in both geometry and appearance across diverse and complex scene changes, encompassing synthetic and real-world scenes. PAPR in Motion consistently outperforms the leading neural renderer for dynamic scenes, Dynamic Gaussian~\cite{Luiten2023Dynamic3G}, on point-level 3D scene interpolation. These findings not only validate the effectiveness of our approach in tackling this challenging new task but also establish a new baseline.
In summary, our contributions are as follows:

\begin{enumerate}
    \item We introduce the novel task of point-level 3D scene interpolation between two distinct scene states. 
    \item We propose PAPR in Motion, a point-based approach capable of synthesizing plausible and smooth 3D scene interpolations without any supervision in between.
    \item We evaluate our approach on a diverse set of complex scene changes, and show that it significantly outperforms the leading neural renderer for dynamic scenes, Dynamic Gaussian~\cite{Luiten2023Dynamic3G}, on this task. 
\end{enumerate}

\section{Related Work}
\label{sec:related_work}

Our work introduces the novel task of point-level 3D scene interpolation, a task for which specific methods have not yet been established. However, in a broader context, our work intersects with and draws insights from related areas such as dynamic scene novel view synthesis and motion analysis and synthesis of 3D objects. 

\vspace{-1em}
\paragraph{Novel View Synthesis for Dynamic Scenes}

The introduction of neural radiance fields (NeRF)~\cite{Mildenhall2020NeRFRS} marked a significant milestone in novel view synthesis, subsequently giving rise to extensive research focused on adapting NeRF for dynamic scenes. Some studies have proposed fitting a separate model for each frame~\cite{Bansal2023NeuralPC,Xian2020SpacetimeNI}, while others have utilized Eulerian representations on a 4D space-time grid to model the scene, optimizing efficiency through techniques like planar decomposition or hash functions~\cite{Cao2023HexPlaneAF,FridovichKeil2023KPlanesER,Turki2023SUDSSU}. Alternatively, some methods define the representation in a canonical timestep and use deformation fields to map this reference to other frames~\cite{Du2020NeuralRF,Li2020NeuralSF,Liu2023RobustDR,Park2020NerfiesDN,Park2021HyperNeRF,Pumarola2020DNeRFNR,Song2023TotalReconDS,Wang2023TrackingEE,Yang2021BANMoBA}. Other research has also explored the use of pre-defined templates or skeleton tracking to guide the deformation process~\cite{Ik2023HumanRFHN,Li2022TAVATA,Weng2022HumanNeRFFR}. While effective in certain scenarios, these approaches rely on prior knowledge on the reconstructed objects, limiting their applicability to general scenes.

Another line of work in the modelling of dynamic scenes has adopted point-based representations~\cite{Zhang2022DifferentiablePR,AbouChakra2022ParticleNeRFAP,Luiten2023Dynamic3G}. Recently, the Dynamic Gaussian method~\cite{Luiten2023Dynamic3G}, which builds on the 3D Gaussian Splatting point-based representation~\cite{kerbl20233d}, has showcased impressive capabilities in long-term point tracking and novel view synthesis. By treating points as particles, it effectively creates a Lagrangian representation that naturally follows points through space and time. 

However, a common assumption in these methods, is the availability of uninterrupted observation of the scene at all time steps. In contrast, our method is designed to function without requiring constant observation, and can address scenarios where only discrete snapshots of the scene at different states are available.

\paragraph{3D Object Motion Analysis and Synthesis}

Unlike the works mentioned earlier and our work, these methods only estimate the change in object geometry, but they cannot produce the appearance of the object as it moves.
The analysis and synthesis of 3D object motion represents a long-standing area of research with considerable advances over time~\cite{Liu2018FlowNet3DLS,Abbatematteo2019LearningTG,Gadre2021ActTP,Haresh2022Articulated3H,Mo2021Where2ActFP}. The rise of data-driven methodologies has further expanded this domain, with numerous methods~\cite{Wang2019Shape2MotionJA,Yan2020RPMNetRP,Jain2020ScrewNetCA,Heppert2023CARTOCA,Liu2019MeteorNetDL,Fan2021Point4T,Zeng2022IDEANetD3} emerging to learn object motions in a supervised framework. A common limitation among these methods, however, is their reliance on 3D supervision and sometimes also on articulation annotations, which are challenging and costly to obtain at scale. Another line of research has focused on category-specific modelling, where separate models are developed for distinct object categories~\cite{Li2019CategoryLevelAO,Mu2021ASDFLD,Tseng2022CLANeRFCA,Wei2022SelfsupervisedNA}. While effective within their defined scope, these methods often face difficulties in generalizing to arbitrary novel objects. 

In a related context, recent work by Liu et al.~\cite{Liu2023PARISPR} investigates using pairs of RGB observations to analyze the motion parameters of an articulated object. Their approach, however, only focuses on part-level rigid motion of a single movable part. In contrast, our approach models point-level scene changes, accommodating both rigid and non-rigid transformations, and handles scenarios where multiple parts of an object move simultaneously.

\section{Problem Setup}
We introduce the novel task of point-level 3D scene interpolation. In this task, we consider a scene undergoing changes within a 3D space, with our inputs limited to observations from only two distinct scene states: the initial state at time $t=0$ and the final state at time $t=T$. Our objective is to generate a smooth and plausible interpolation at the point level between these two states, without observing any intermediate states, as illustrated in Figure~\ref{fig:what_fig}. This means for each point $i$ in the scene, which moves from its initial location $p_i^{0}$ at the start state to its final location $p_i^{T}$ at the end state, we aim to generate its trajectory $\{p_i^t\}_{t\in(0, T)}$ through time and space. Moreover, the synthesized images $\hat{\mathbf{I}}_t$ of the scene at any given time step $t\in[0, T]$ should also appear realistic, thereby yielding a visually coherent interpolation between the initial and final states.

We are given a set of RGB images of a 3D scene at two states from multiple views along with their associated camera poses. It is important to note the distinction between our problem setting and static 3D scene reconstruction~\cite{kerbl20233d,Mildenhall2020NeRFRS,Zuo2022ViewSW,Yu2021PlenoxelsRF,Lassner2021PulsarES,Xu2022PointNeRFPN}, which does not consider changes in the scene over time. It also differs from the classic problem of novel view synthesis of dynamic 3D scenes~\cite{Pumarola2020DNeRFNR,Park2020NerfiesDN,Liu2023RobustDR,Park2021HyperNeRF}, where all time steps between start and end states are typically observed. Our focus is on synthesizing plausible interpolations between scenes at two distinct states, rather than on reconstruction or tracking of observed motion across time. Our problem setting is inherently quite challenging due to the potential for significant scene changes between the two observed states.

\section{Method}
To achieve point-level 3D scene interpolation between two states, our approach starts with training a point-based neural rendering model on the initial state, followed by finetuning this model to match the final state. The intermediate iterations of this finetuning process generate the interpolations between the initial and final states. 
The rest of this section starts by discussing the rationale behind our choice of the point-based rendering technique we build upon, namely Proximity Attention Point Rendering (PAPR)~\cite{zhang2023papr} (Section~\ref{sec:why_papr}). This is followed by a brief overview of PAPR (Section~\ref{sec:papr}). Finally, we delineate our method, ``PAPR in Motion'', which leverages the capabilities of PAPR to generate scene interpolations (Section~\ref{sec:papr-in-motion}).

\subsection{Choice of Point-based Rendering Technique}
\label{sec:why_papr}

Our method leverages point-based representations to model long-range scene dynamics in an Lagrangian fashion. Given the challenges posed by large scene changes and non-rigid transformations, selecting an appropriate point-based renderer is critical.

Point-based neural renderers come in two flavours: splat-based renderers and attention-based renderers. Splat-based renderers~\cite{kerbl20233d,Wiles2019SynSinEV,Rckert2021ADOPAD} place primitives, e.g., ellipsoids or Gaussian kernels, at every point and check which primitives intersect with a given ray. On the other hand, attention-based renderers, like PAPR~\cite{zhang2023papr}, directly predict the intersection point between the surface and a given ray by interpolating between nearby points using attention. 

In our problem setting, since there may be large scene changes between the observed states, the initial and final point clouds should differ substantially. We opt for PAPR over splat-based renderers because the latter struggle with large scene changes. This limitation of splat-based renderers arises from the decreasing contribution of each point to a pixel as it moves away from the pixel, leading to vanishing gradients for distant points. Consequently, splat-based renderers can become stuck in point cloud configurations far from the ground truth, particularly when the initial point cloud differs substantially from it.
In contrast, instead of solely considering the absolute distance of each point to the pixel, PAPR takes the relative distances of different points to the pixel into account. This is achieved through an attention mechanism, which normalizes the contributions of all points, ensuring their total contribution always sums to 1. As a result, there will always be points that have significant contributions to each pixel, even for pixels far from all points. This prevents the gradient from vanishing even when the point cloud is far from the ground truth geometry, thereby allowing PAPR to move points at ease regardless of where they were initially. 

Furthermore, because there can be complex, non-rigid geometry changes between observed states, synthesizing realistic views after applying such deformations becomes crucial. Again, PAPR outperforms splat-based renderers in this aspect. Splat-based renderers often produce porous shapes after non-rigid transformations, as shown in \cite{zhang2023papr}. This phenomenon stems from the need to redistribute splats to maintain surface coverage after non-rigid deformations like stretching. In contrast, PAPR, by implicitly representing surfaces through point interpolation, can adapt to changes in point density, preserving shape contiguity and visually coherent appearances under non-rigid deformations.

\subsection{Overview of Proximity Attention Point Rendering (PAPR)}
\label{sec:papr}

PAPR takes a set of RGB images of a static 3D scene from multiple views and their associated camera poses as input, and jointly learns a point-based scene representation, an attention mechanism, and a differentiable renderer based on U-Net architecture.
Each point $i$ in the scene representation consists of a point location $\mathbf{p}_i\in \mathbb{R}^3$, an influence score $\tau_i\in \mathbb{R}$ and a view-independent feature vector $\mathbf{u}_i\in \mathbb{R}^d$. 

Given a camera pose, PAPR casts a ray from the camera's centre of projection through each pixel on the image plane. At each ray, PAPR first produces a ray-dependent embedding for each point. It then uses the attention mechanism to compute the relative contribution weights of these points, and combines the feature vectors of these points using the output weights. The aggregated feature map is then fed to the differentiable renderer to produce the output image $\hat{\mathbf{I}}$. 
The entire pipeline is trained end-to-end to minimize reconstruction loss between the output $\hat{\mathbf{I}}$ and ground truth image $\mathbf{I}_{gt}$, as measured by the function $\mathcal{L}_{recon}$:
\begin{align}
    \mathcal{L}_{recon} = \text{MSE}(\hat{\mathbf{I}}, \mathbf{I}_{gt}) + \lambda\cdot\text{LPIPS}(\hat{\mathbf{I}}, \mathbf{I}_{gt})
\end{align}
Here, the reconstruction loss is composed of a weighted combination of mean squared error (MSE) and the LPIPS metric~\cite{Zhang2018TheUE}. For further details and in-depth analysis of PAPR, we direct readers to the original paper~\cite{zhang2023papr}.

\begin{figure*}[t]
\footnotesize
\begin{tabularx}{\linewidth}{lYYYY|YYYY}
& \multicolumn{2}{c}{Dynamic Gaussian~\cite{Luiten2023Dynamic3G}} & \multicolumn{2}{c}{PAPR in Motion (Ours)} & \multicolumn{2}{c}{Dynamic Gaussian~\cite{Luiten2023Dynamic3G}} & \multicolumn{2}{c}{PAPR in Motion (Ours)} \\
& Point Cloud & Rendering & Point Cloud & Rendering & Point Cloud & Rendering & Point Cloud & Rendering \\
    \rotatebox[origin=c]{90}{Start} 
    & \includegraphics[width=\hsize,valign=m]{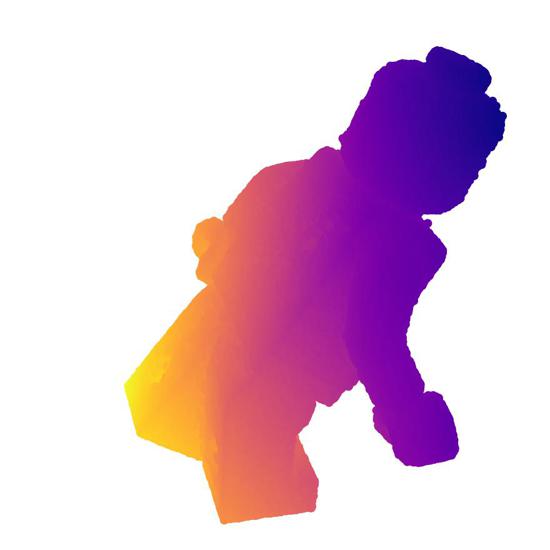}
    & \includegraphics[width=\hsize,valign=m]{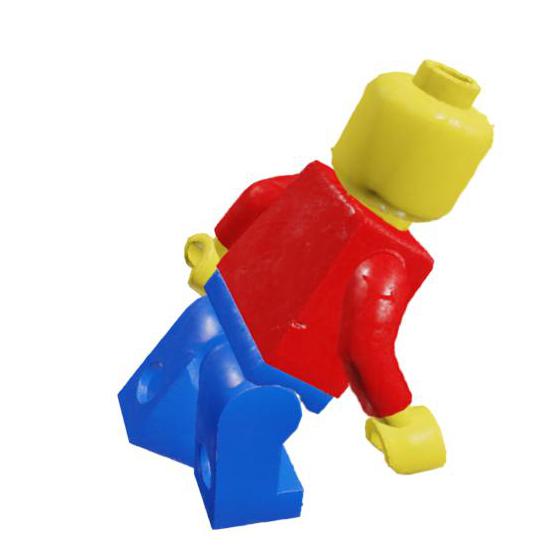}
    & \includegraphics[width=\hsize,valign=m]{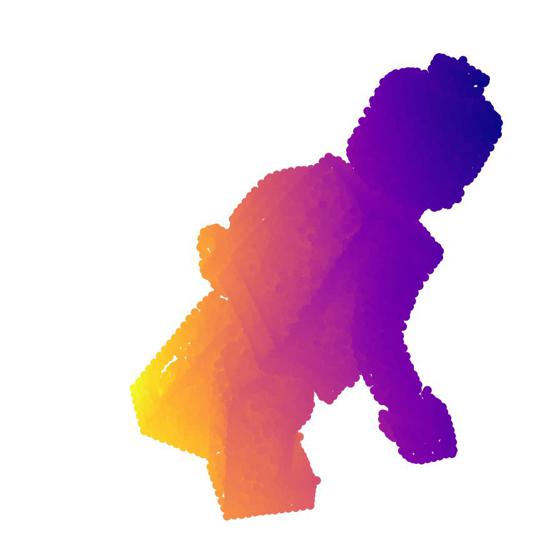}
    & \includegraphics[width=\hsize,valign=m]{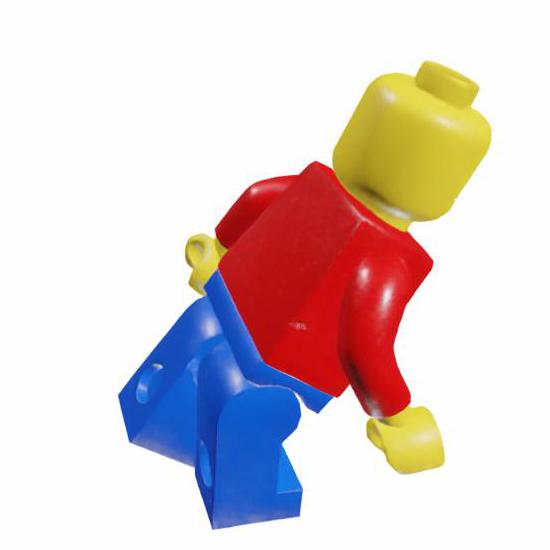}
    & \includegraphics[width=\hsize,valign=m]{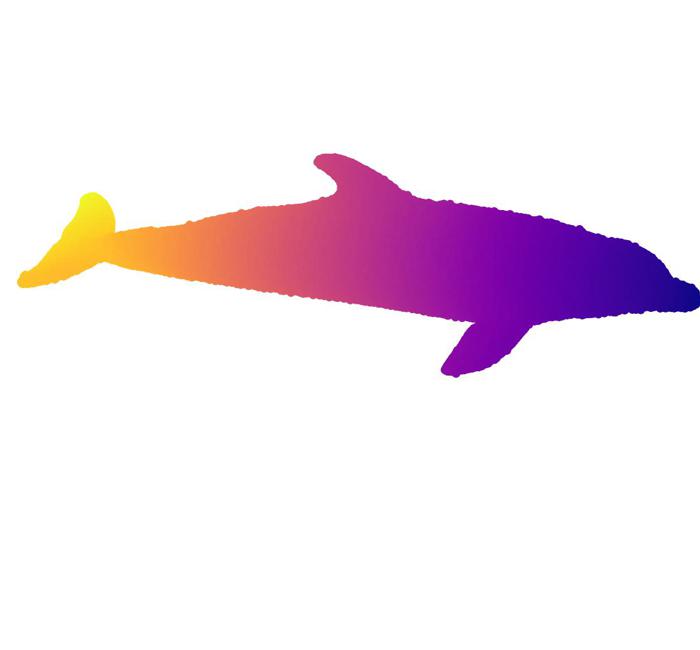}
    & \includegraphics[width=\hsize,valign=m]{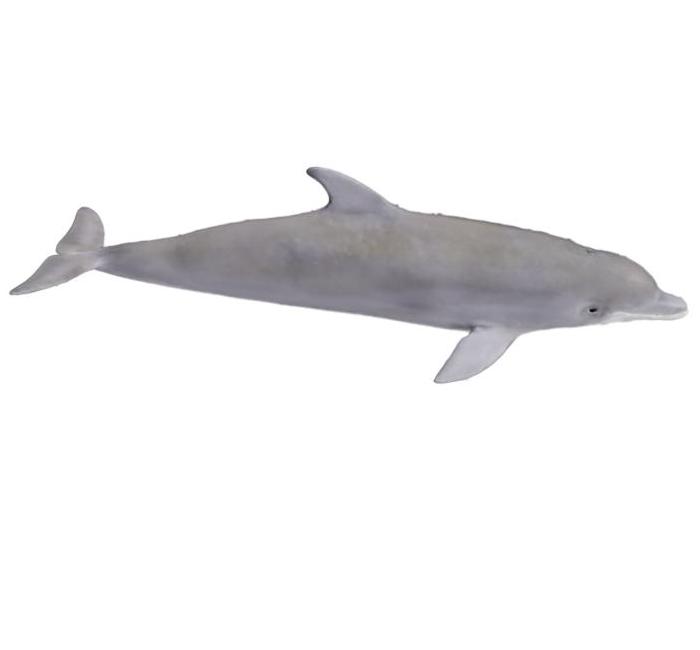}
    & \includegraphics[width=\hsize,valign=m]{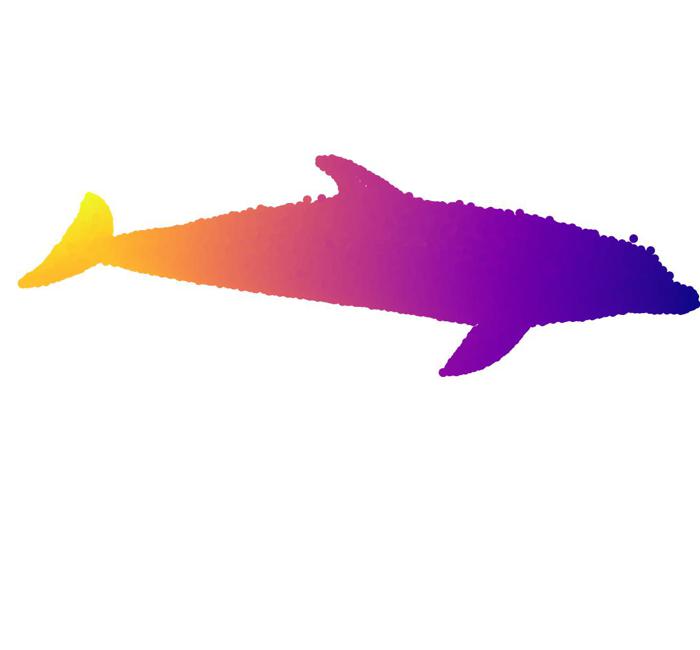}
    & \includegraphics[width=\hsize,valign=m]{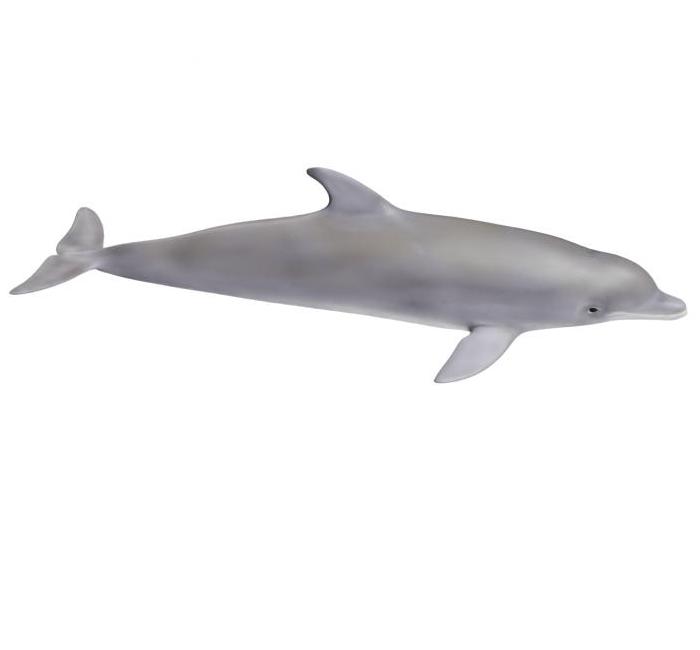}
    \\
    \multirow{2}{*}{\rotatebox[origin=c]{90}{Intermediate}}
    & \includegraphics[width=\hsize,valign=m]{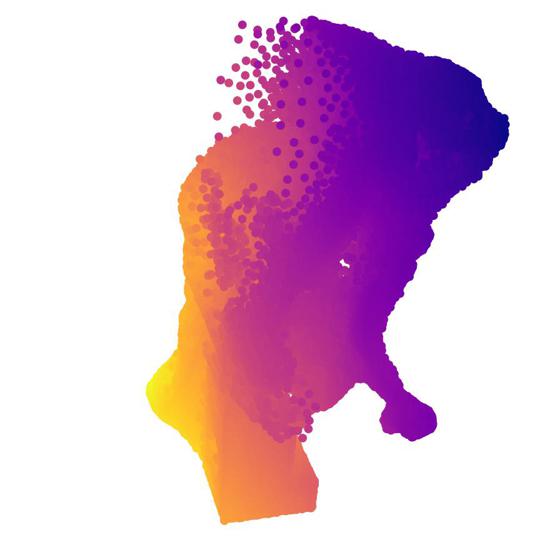}
    & \includegraphics[width=\hsize,valign=m]{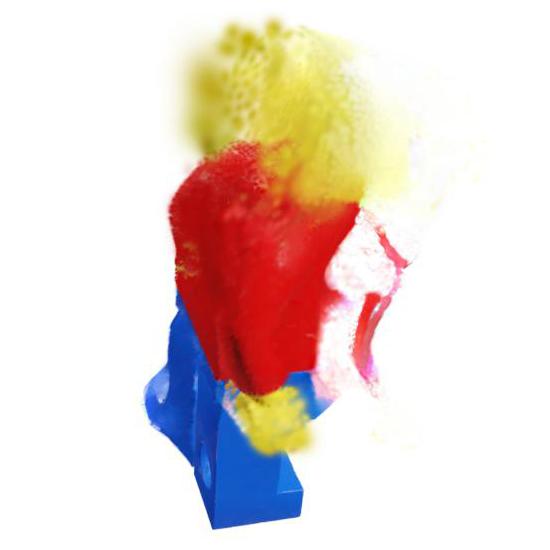}
    & \includegraphics[width=\hsize,valign=m]{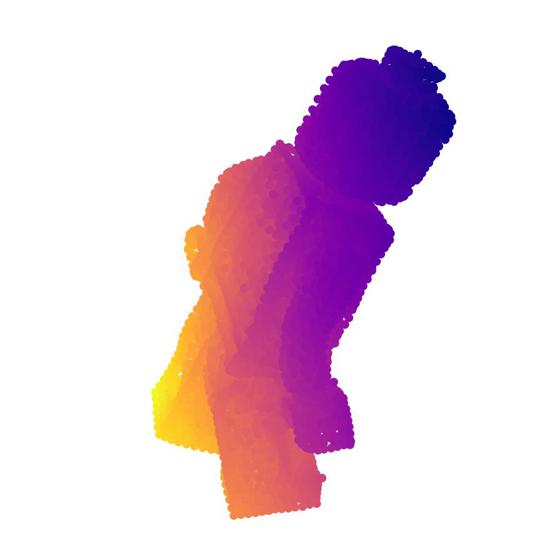}
    & \includegraphics[width=\hsize,valign=m]{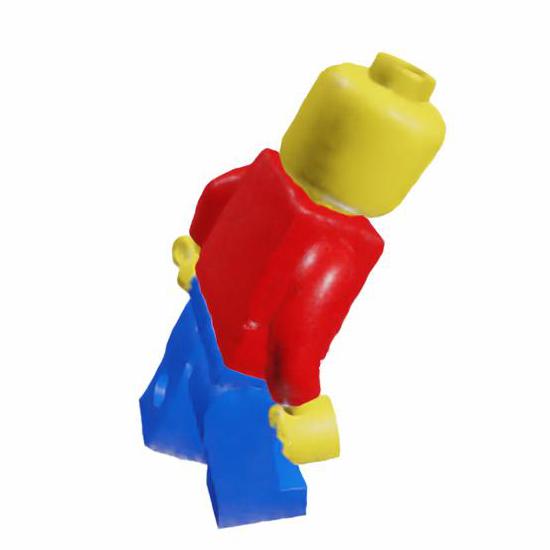}
    & \includegraphics[width=\hsize,valign=m]{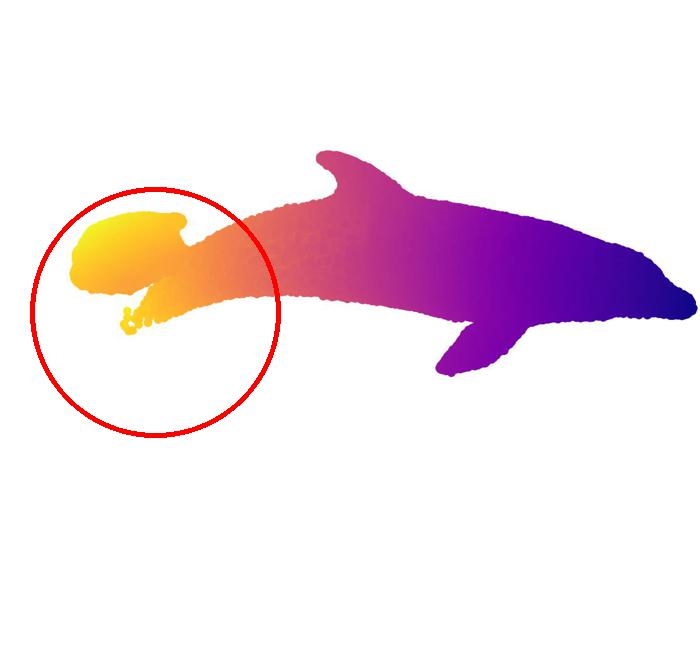}
    & \includegraphics[width=\hsize,valign=m]{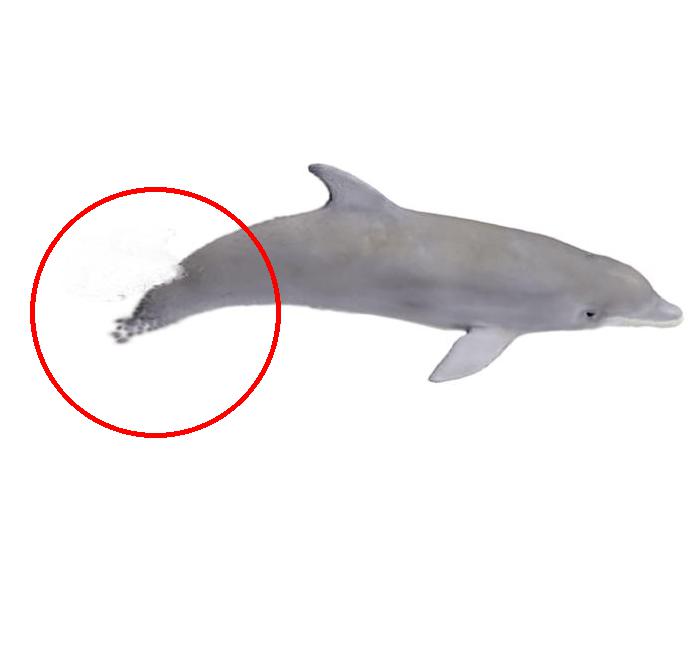}
    & \includegraphics[width=\hsize,valign=m]{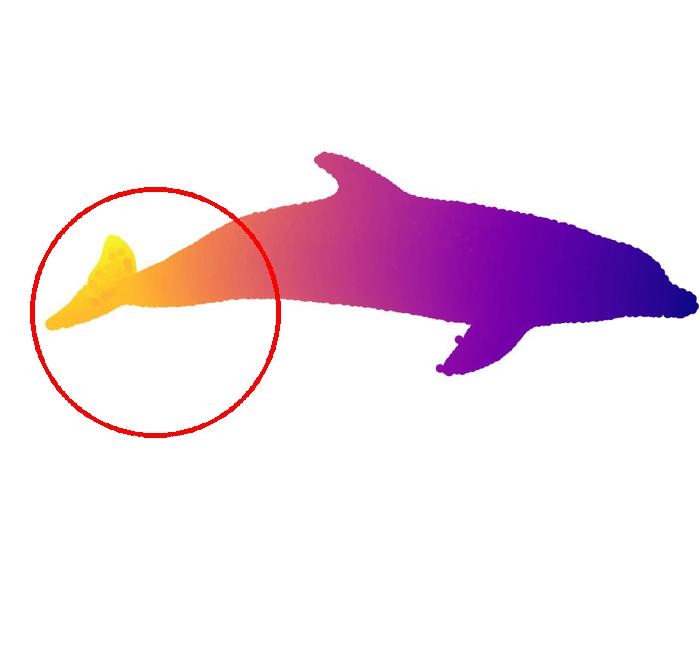}
    & \includegraphics[width=\hsize,valign=m]{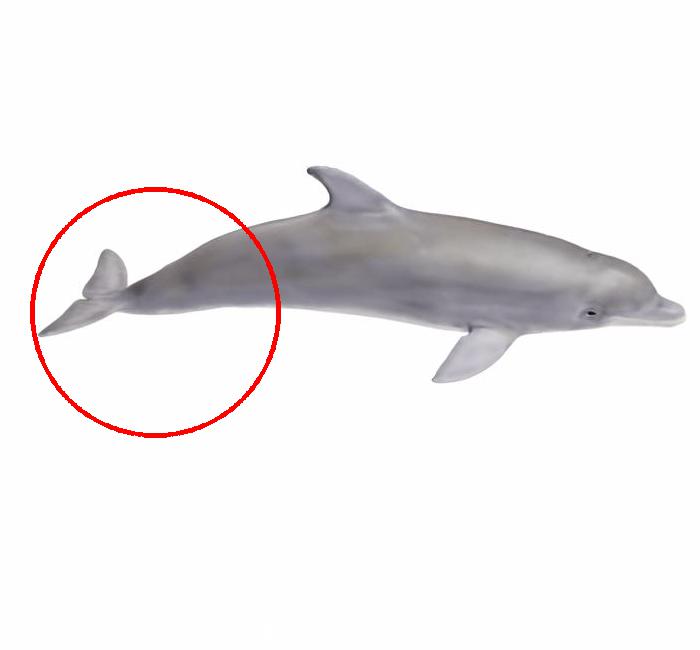}
    \\

    & \includegraphics[width=\hsize,valign=m]{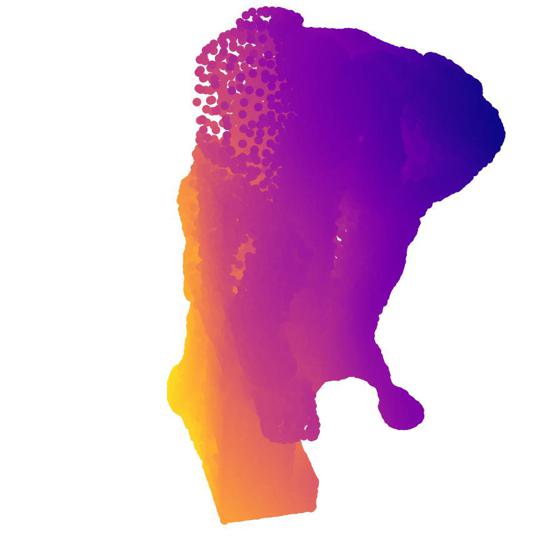}
    & \includegraphics[width=\hsize,valign=m]{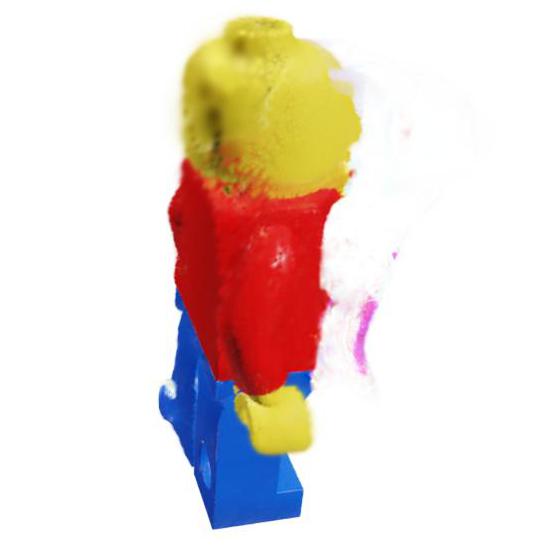}
    & \includegraphics[width=\hsize,valign=m]{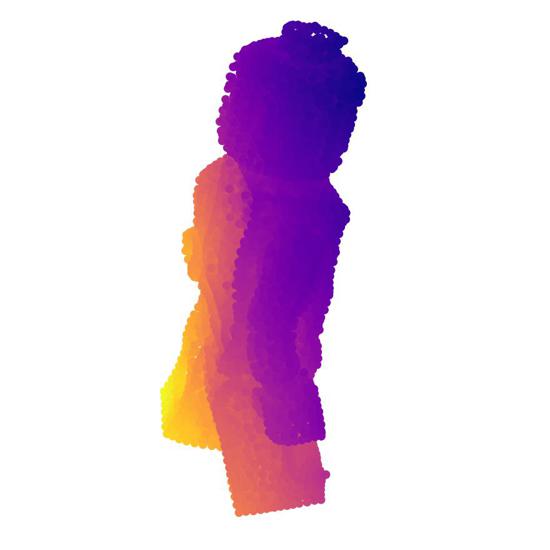}
    & \includegraphics[width=\hsize,valign=m]{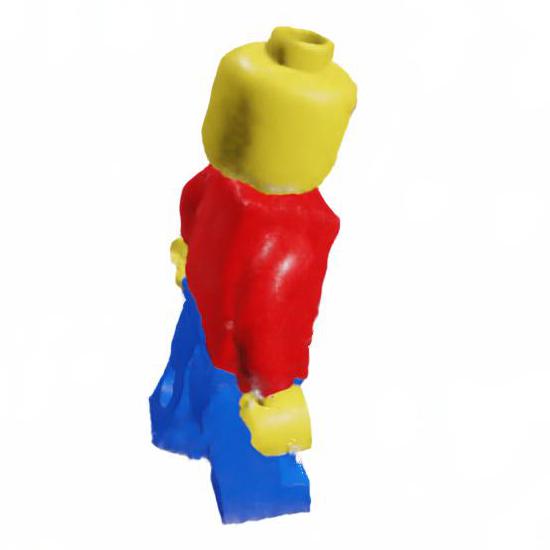}
    & \includegraphics[width=\hsize,valign=m]{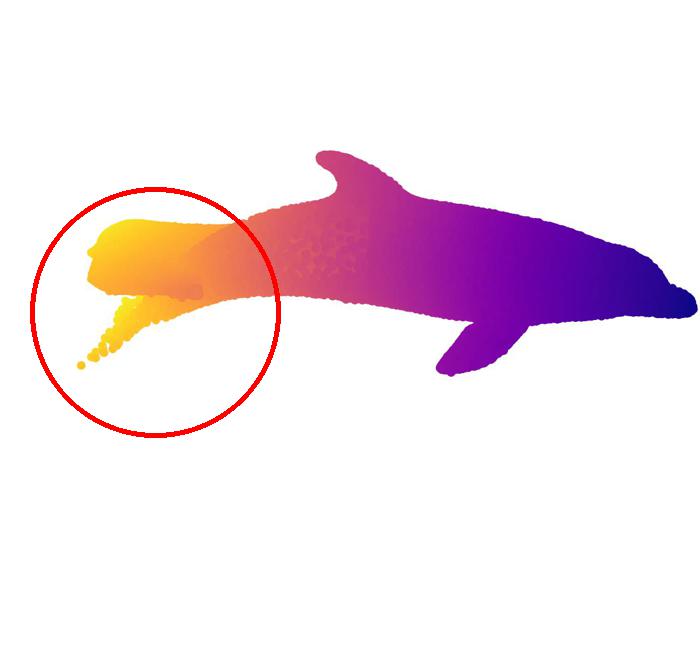}
    & \includegraphics[width=\hsize,valign=m]{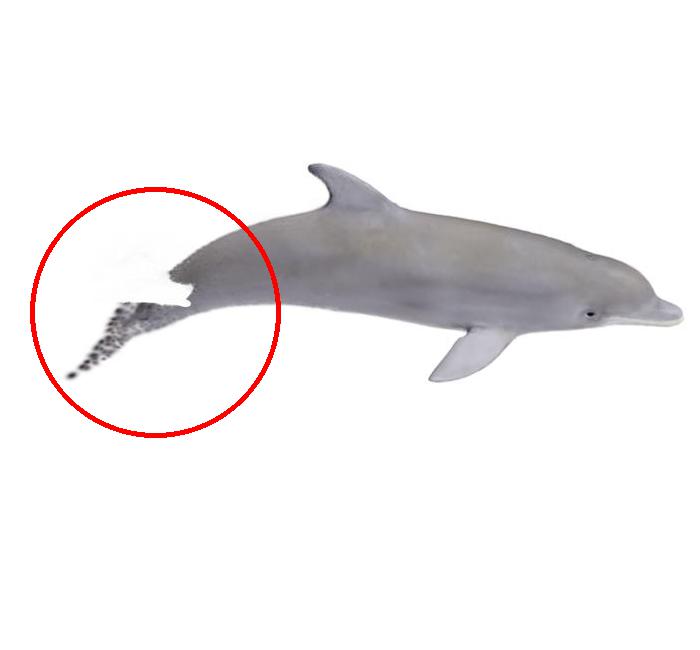}
    & \includegraphics[width=\hsize,valign=m]{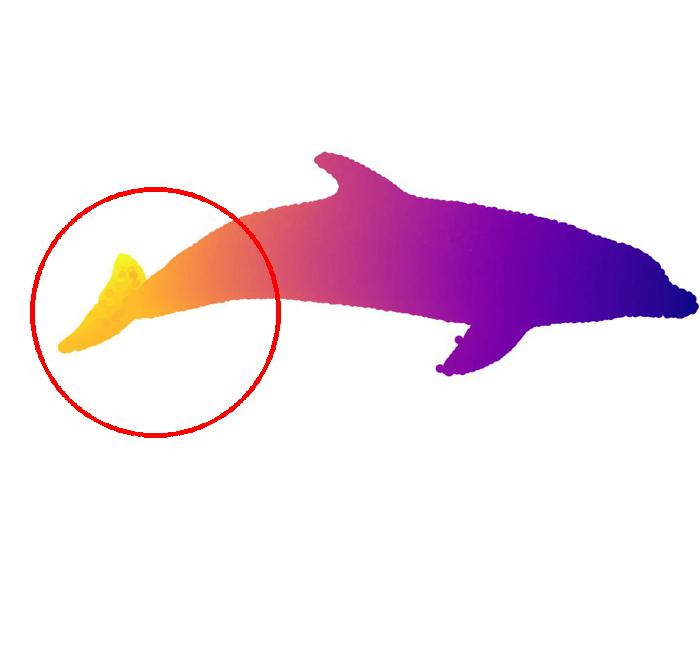}
    & \includegraphics[width=\hsize,valign=m]{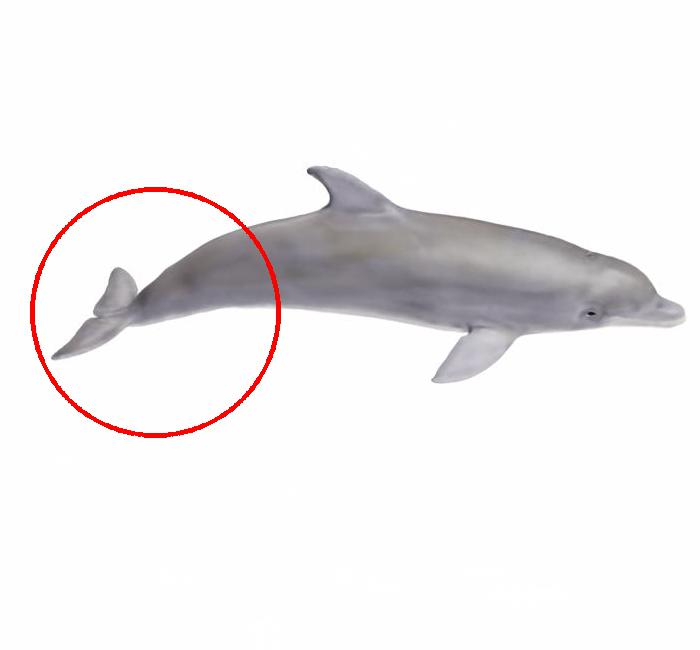}
    \\

    \rotatebox[origin=c]{90}{End}
    & \includegraphics[width=\hsize,valign=m]{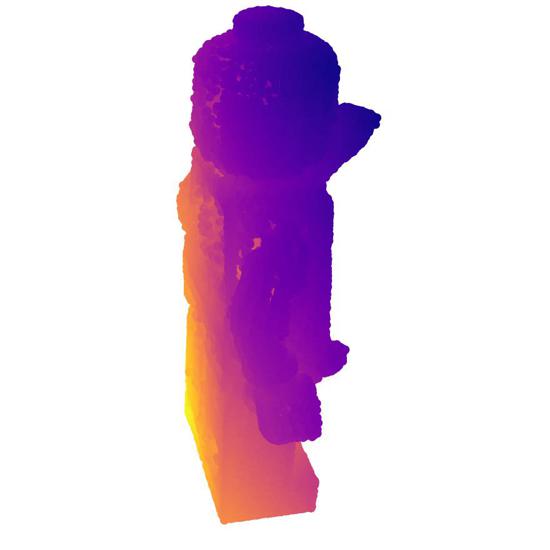}
    & \includegraphics[width=\hsize,valign=m]{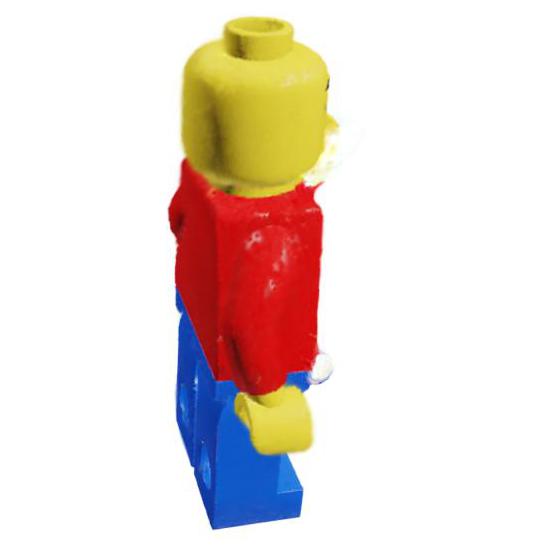}
    & \includegraphics[width=\hsize,valign=m]{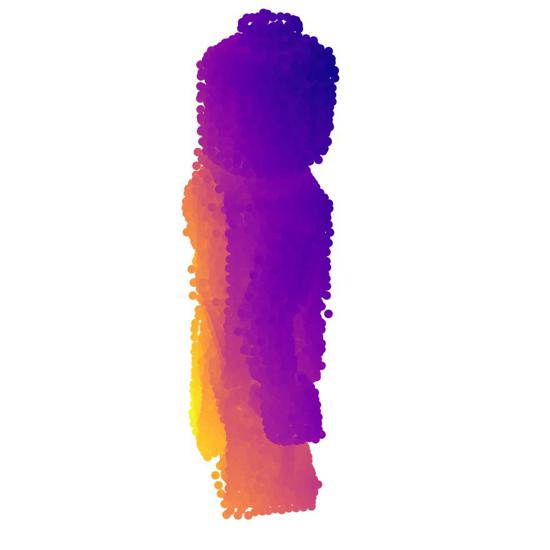}
    & \includegraphics[width=\hsize,valign=m]{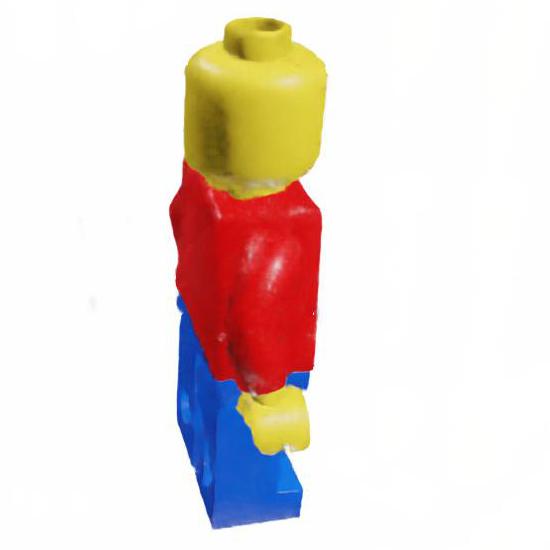}
    & \includegraphics[width=\hsize,valign=m]{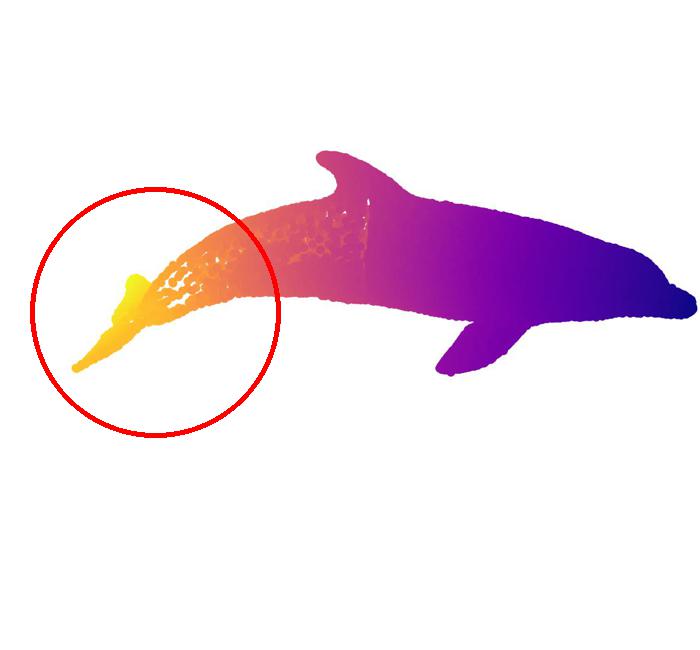}
    & \includegraphics[width=\hsize,valign=m]{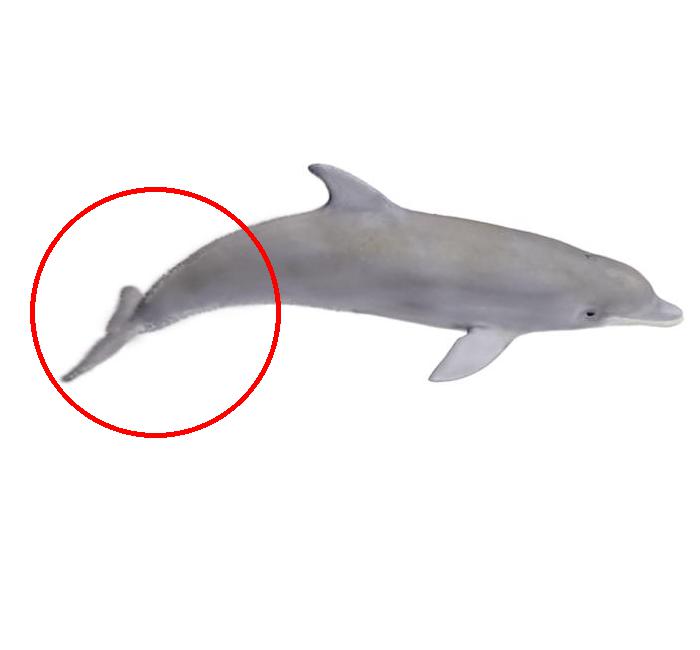}
    & \includegraphics[width=\hsize,valign=m]{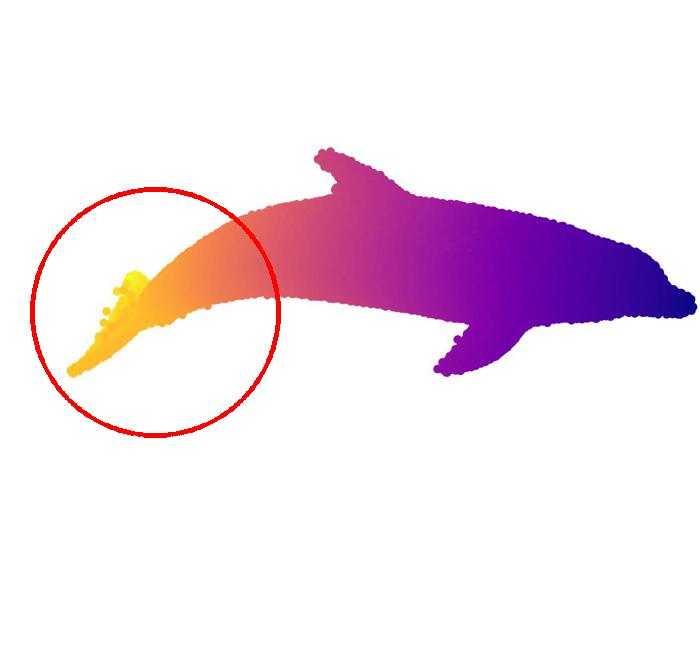}
    & \includegraphics[width=\hsize,valign=m]{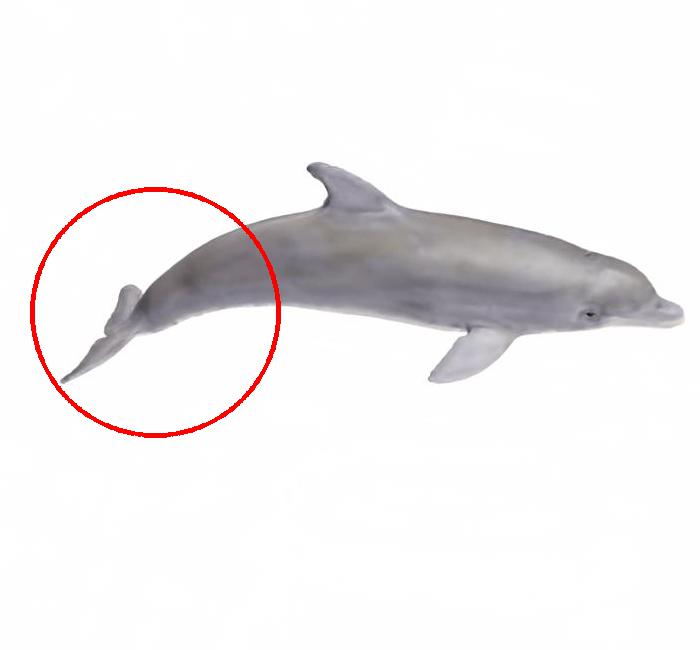}
    \\
    \rotatebox[origin=c]{90}{Trajectory}
    & \multicolumn{2}{c}{\includegraphics[width=0.12\linewidth,valign=m]{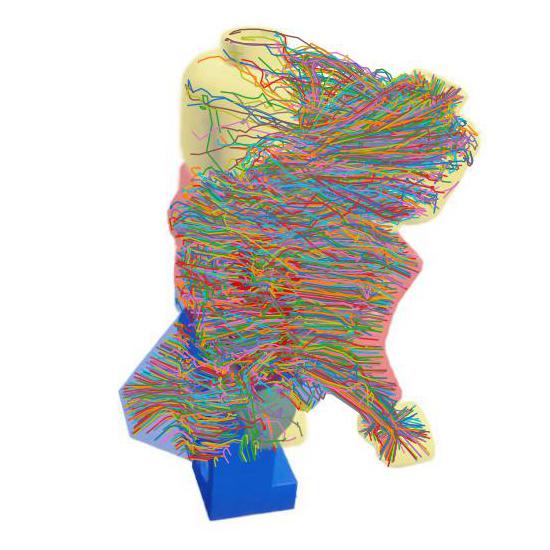}}
    & \multicolumn{2}{c}{\includegraphics[width=0.12\linewidth,valign=m]{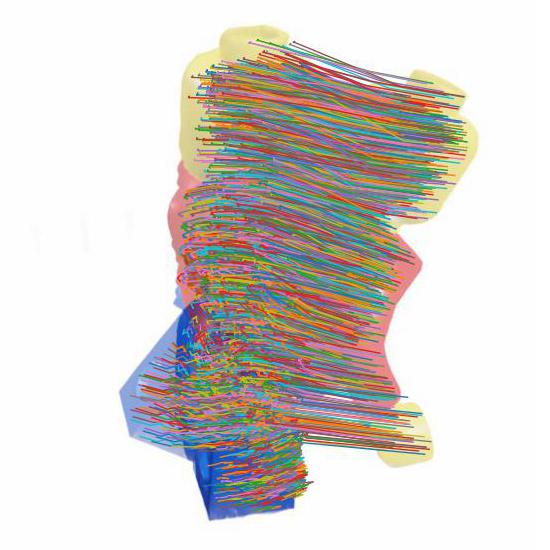}} 
    & \multicolumn{2}{c}{\includegraphics[width=0.18\linewidth,valign=m]{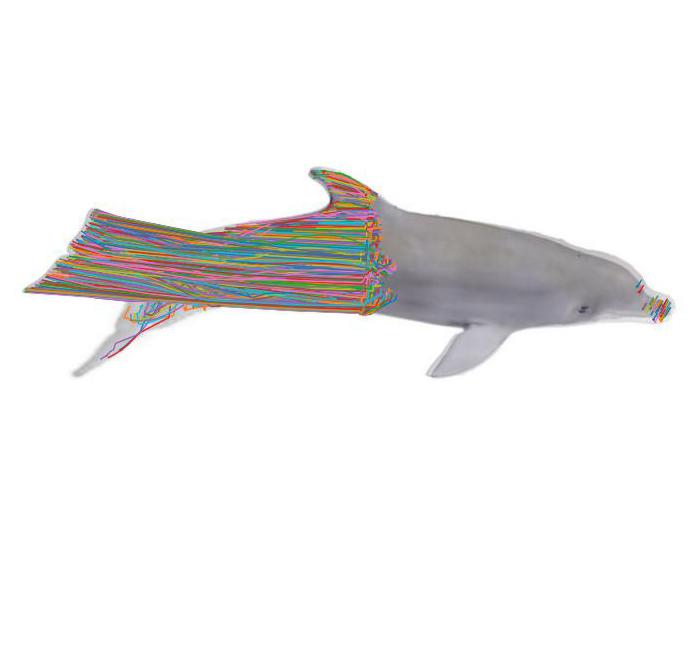}}
    & \multicolumn{2}{c}{\includegraphics[width=0.18\linewidth,valign=m]{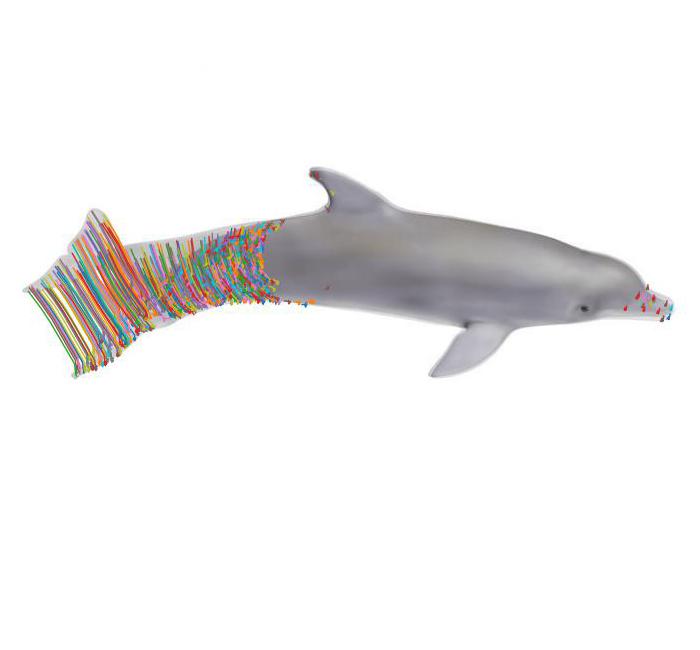}} 
\end{tabularx}
\caption{    
Qualitative comparison of 3D scene interpolation from start to end state using synthetic scenes. Both methods start by training a static model for the start state and subsequently finetune it towards the end state, all without any intermediate supervision. Dynamic Gaussian~\cite{Luiten2023Dynamic3G} struggles to maintain consistent scene geometry and appearance, resulting in unnatural motion, as shown by the point trajectories. On the other hand, PAPR in Motion creates smooth and plausible interpolations between states. It successfully handles a variety of transformations, from the rigid motion observed in the Lego man's standing pose to the non-rigid, fluid motion of the dolphin's tail.
}
\label{fig:qualitative}
\end{figure*}

\subsection{PAPR in Motion}
\label{sec:papr-in-motion}

We assume that the start and end states differ in geometry, possibly substantially, but are similar in appearance. However, because the ground truth geometries are not known at any intermediate state, we must estimate them. To do so, we will build an scene template that can support different geometries and fit the same template to both states by varying the geometry parameters. To build the scene template, we train a PAPR model of the scene at the start state. Given PAPR's proficiency in handling geometry deformations, we can use the PAPR model with all parameters fixed except for the point coordinates as an scene template. In this template, the point coordinates are the free parameters that represent geometry. On the other hand, the view-independent feature vectors associated with the points carry localized scene texture information, and represent the appearance. Each point can be viewed as a geometry-agnostic marker of a part that can move freely to match the observed state of the scene. 

We can then fit the scene template to the end state. This fitting process would gradually change the geometry from the start to the end state, resulting in a natural interpolation trajectory. More concretely, this entails finetuning the previously trained PAPR model using images from the end state, while keeping all parameters fixed except for the point coordinates. As the points' spatial positions change during finetuning, we track the trajectories they follow, which forms a Lagrangian representation of the scene's evolution.

To generate a point-level 3D scene interpolation, we use these trajectories to obtain the scene's geometry at intermediate states. As the appearance may vary between observed states due to geometry changes (e.g., cast shadows or specular reflections), we finetune the appearance parameters of our model after fitting the scene template's geometry to the end state. To determine the appearance at an intermediate state, we interpolate between the appearance parameters of the models at the start and end states. By combining these with the geometry parameters at that state, we can use PAPR to synthesize novel views of the scene at that particular state.

Below we delineate how we make interpolation trajectories physically plausible and handle appearance changes. 
\vspace{-1em}
\paragraph{Ensuring Plausible Interpolation Trajectories}

Na\"{i}vely finetuning the positions of all points independently can lead to implausible deformations to the part geometry or some points becoming trapped in a local optimum, especially in cases with substantial scene changes or in regions with near-constant texture. 

To avoid implausible deformations, we propose adding a regularization term, termed the \textbf{local distance preservation loss}, during finetuning, which encourages local rigidity by preserving distances between each point and its neighbours.

Specifically, we begin by finding the $k$-nearest neighbours $\text{NN}_{k}(i)$ for each point $i$ based on their positions at the start state $\mathbf{p}_i^{0}$, and keeping track of its squared $\ell_2$ distances $d_{i, j}^{0}$ to each neighbouring point $j$ in $\text{NN}_{k}(i)$:
\begin{align}
    d^0_{i, j} = ||\mathbf{p}_i^{0} - \mathbf{p}_j^{0}||_2^2
\end{align}
At any given time $t\in(0, T]$, we calculate the regularization loss based on the current squared $\ell_2$ distances $d_{i, j}^t$ from $\mathbf{p}_i^t$ to its neighbours from the start state: 
\begin{align}
    d_{i, j}^t &= ||\mathbf{p}_i^{t} - \mathbf{p}_j^{t}||_2^2\\
    \mathcal{L}_{rigid} &= \frac{1}{kN}\sum_{i}\sum_{j\in \text{NN}_{k}(i)}|d_{i, j}^0 - d_{i, j}^t|
\end{align}
where $N$ represents the total number of points in the scene. The updated loss function, incorporating this new regularization term, becomes the following:
\begin{align}
    \mathcal{L} = \mathcal{L}_{recon} + \lambda_\text{rigid}\cdot\mathcal{L}_{rigid}
\end{align}

While the integration of a local distance preservation loss helps regulate the movements of points and avoid excessive drift, we find that it's not always sufficient in cases involving substantial scene changes. In such scenarios, it may not fully ensure that points within a moving part move cohesively as a unit. This issue stems from the loss's inability to account for the directions of point motion. To address this and further smooth out point motion, we introduce a \textbf{local displacement averaging step}. 
This step is designed to encourage uniform motion of nearby points and its neighbours, which is crucial for maintaining the cohesiveness of moving parts and ensuring temporal visual coherence.

In the local displacement averaging step, we modify the position of each point $i$ by first recalculating its displacement from its initial position at the start state $\mathbf{h}_i^t = \mathbf{p}_i^t - \mathbf{p}_i^0$. Then we adjust it based on the average displacement of its $k$-nearest neighbours at the start state $\text{NN}_{k}(i)$ from their respective initial positions:
\begin{align}
    \mathbf{\hat{h}}_i^t&= \frac{1}{k}\sum_{j\in \text{NN}_{k}(i)}(\mathbf{p}_j^t - \mathbf{p}_j^0)\\
    \mathbf{\hat{p}}_i^t &= \mathbf{p}_i^0 + \mathbf{\hat{h}}_i^t
\end{align}
Here, $\mathbf{\hat{p}}_i^t$ denotes the adjusted position for point $i$. This averaging regularization is applied every $m$ iterations to gently guide the points towards a more coherent trajectory.

Once the geometry has nearly converged, we disable local displacement averaging to allow the point cloud to more closely match the end state geometry. 

The point positions generated throughout this optimization process serve as interpolations for the scene geometry. We then apply temporal smoothing to the point positions by taking the arithmetic mean over a sliding window of size 7.

\paragraph{Handling Appearance Changes}

In the final phase, with the scene geometry aligned to the geometry of the end state, we shift our focus to refining the appearance parameters of the template. At this stage, the positions of the points are fixed, and we finetune two key elements. First, the feature vectors associated with each point are optimized. This finetuning specifically targets adapting view-independent aspects of scene appearance, for example changes in colour due to variations in shadows cast by moving parts. Second, we refine the ray-dependent attention mechanism. This adjustment is crucial for adapting view-dependent effects, including reflections. The combination of adjusting both the feature vectors and the attention mechanism enables a more precise capture of the nuances in appearance between the start and end states.

To render the appearance for an intermediate state $t\in (0, T)$ with point positions $\{\mathbf{p}_i^t\}_{i=1}^N$,  we interpolate both the feature vectors and the model weights associated with the attention mechanism between the original model and the model after end-state finetuning. This interpolation enables gradual and smooth changes in appearance, such as shadows or specular reflections, between the start and end states. The interpolation weight is given by its progress $\alpha_t$ in the interpolation process within the range of $[0,1]$ as follows:
\begin{align}
    \alpha_t = \frac{\sum_{i=1}^N{||\mathbf{p}_i^t - \mathbf{p}_i^0||_2}}{\sum_{i=1}^N{||\mathbf{p}_i^T - \mathbf{p}_i^0||_2}}
    \label{eqn:progress}
\end{align}
Here, $N$ denotes the total number of points in the point cloud, and $||\cdot||_2$ represents the $\ell_2$ distance. Essentially, $\alpha_t$ measures the progress of interpolation using the total distance travelled by the points from their initial positions. The farther they have travelled relative to their final distances, the later the checkpoint is on the interpolation timeline. 

\section{Experiments}
\label{sec:exp}

\begin{figure*}[t]
\footnotesize
\begin{tabularx}{\linewidth}{lYYYY|YYYY}
& \multicolumn{2}{c}{Dynamic Gaussian~\cite{Luiten2023Dynamic3G}} & \multicolumn{2}{c}{PAPR in Motion (Ours)} & \multicolumn{2}{c}{Dynamic Gaussian~\cite{Luiten2023Dynamic3G}} & \multicolumn{2}{c}{PAPR in Motion (Ours)} \\
& Point Cloud & Rendering & Point Cloud & Rendering & Point Cloud & Rendering & Point Cloud & Rendering \\
    \rotatebox[origin=c]{90}{Start} 
    & \includegraphics[width=\hsize,valign=m]{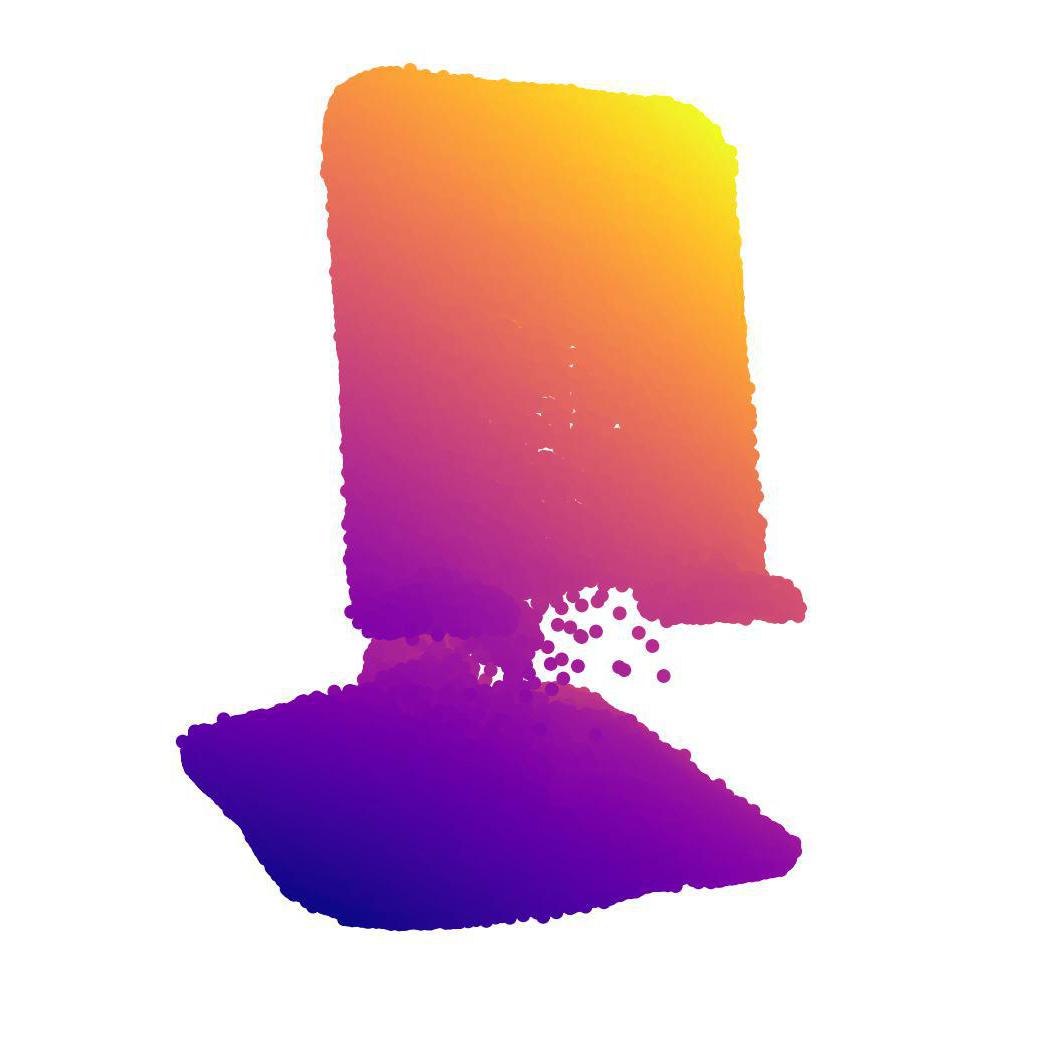}
    & \includegraphics[width=\hsize,valign=m]{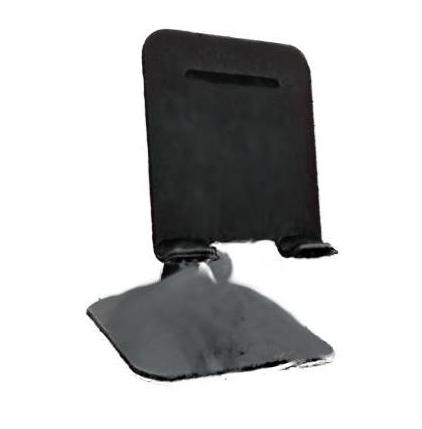}
    & \includegraphics[width=\hsize,valign=m]{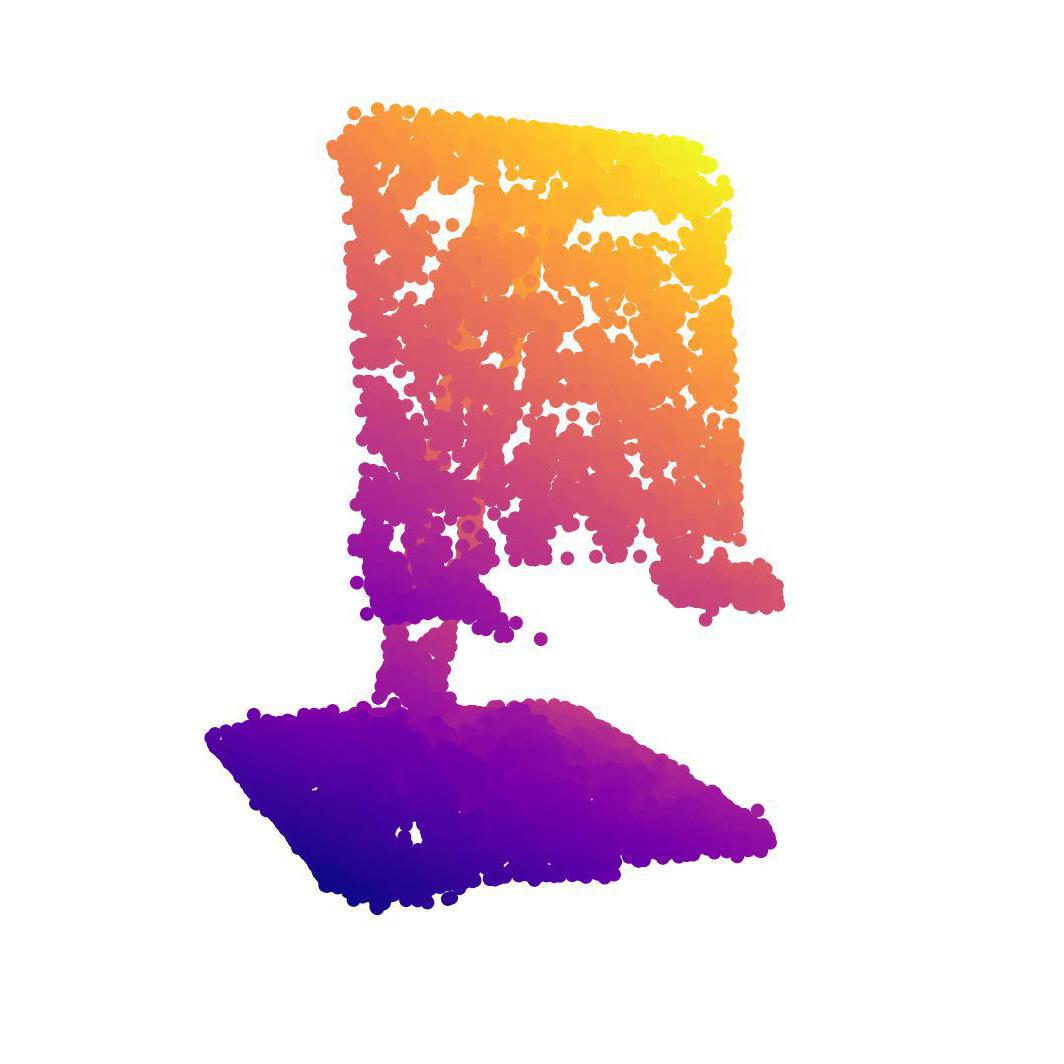}
    & \includegraphics[width=\hsize,valign=m]{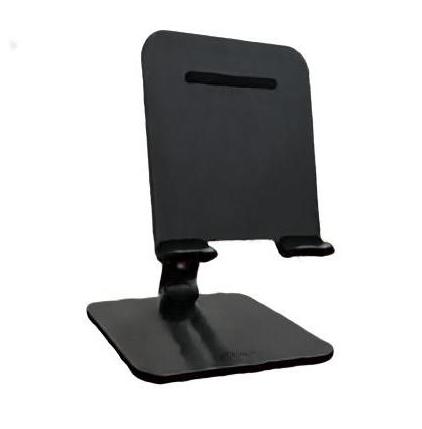}
    & \includegraphics[width=\hsize,valign=m]{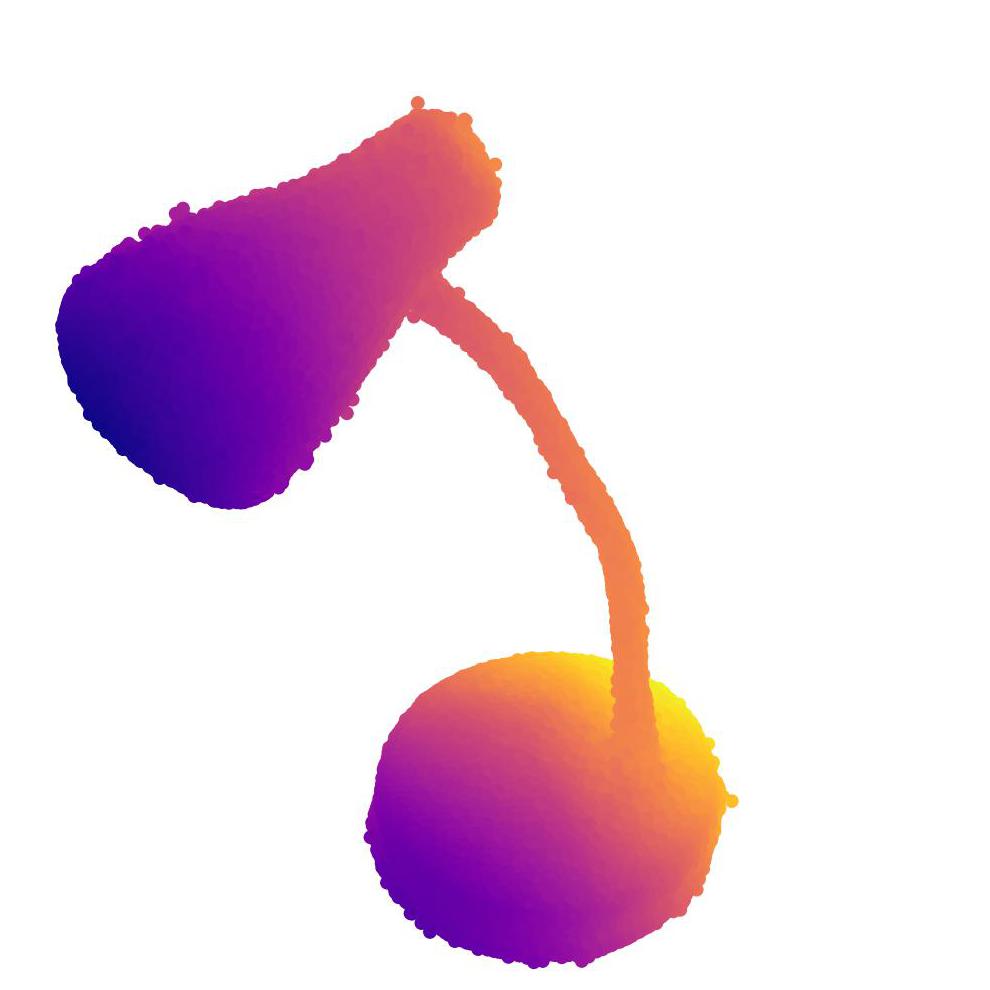}
    & \includegraphics[width=\hsize,valign=m]{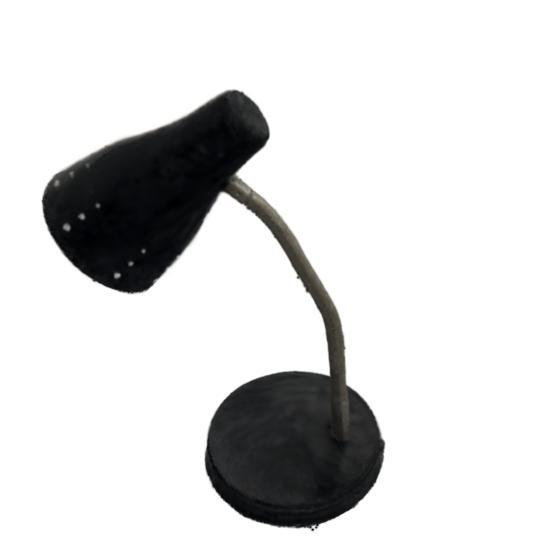}
    & \includegraphics[width=\hsize,valign=m]{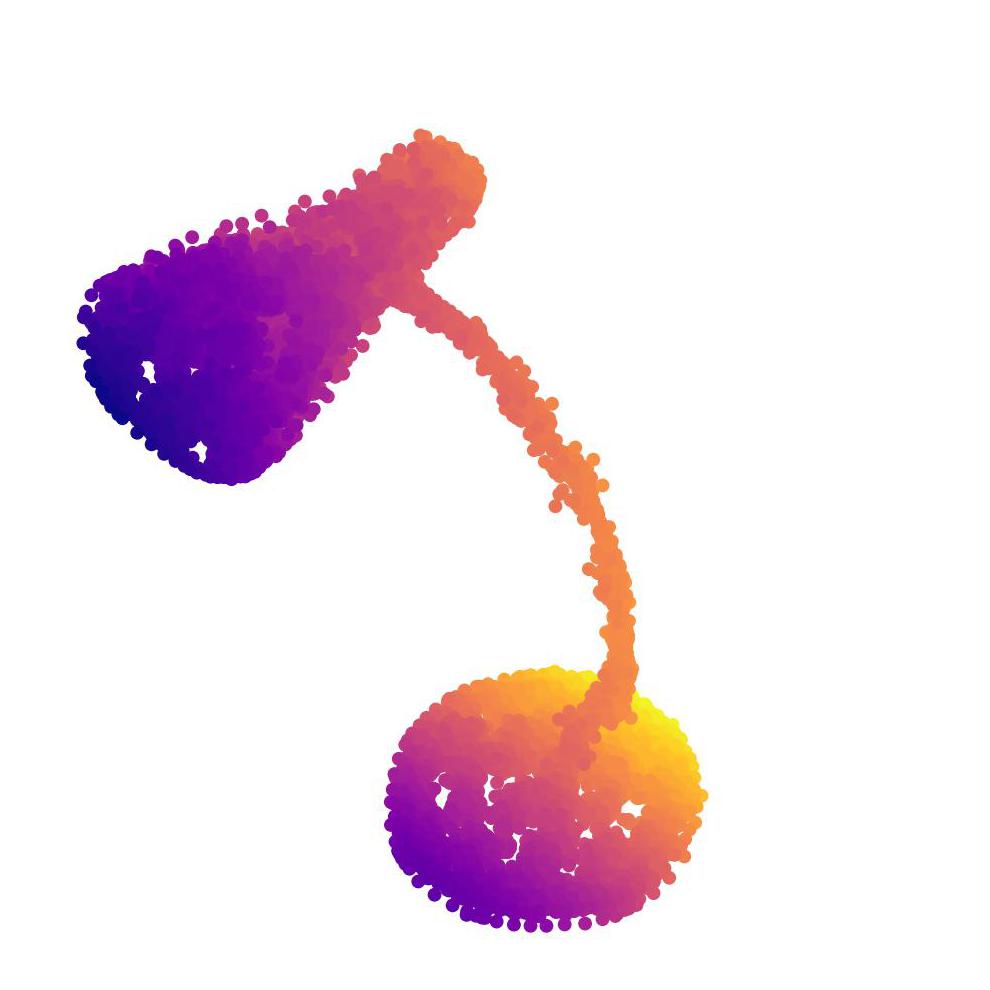}
    & \includegraphics[width=\hsize,valign=m]{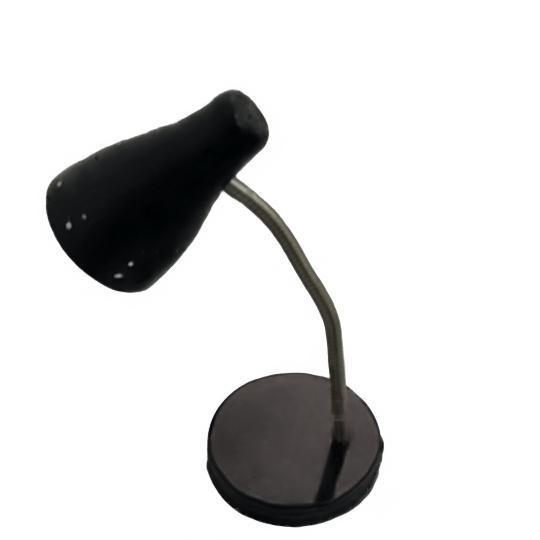}
    \\
    \multirow{2}{*}{\rotatebox[origin=c]{90}{Intermediate}}
    & \includegraphics[width=\hsize,valign=m]{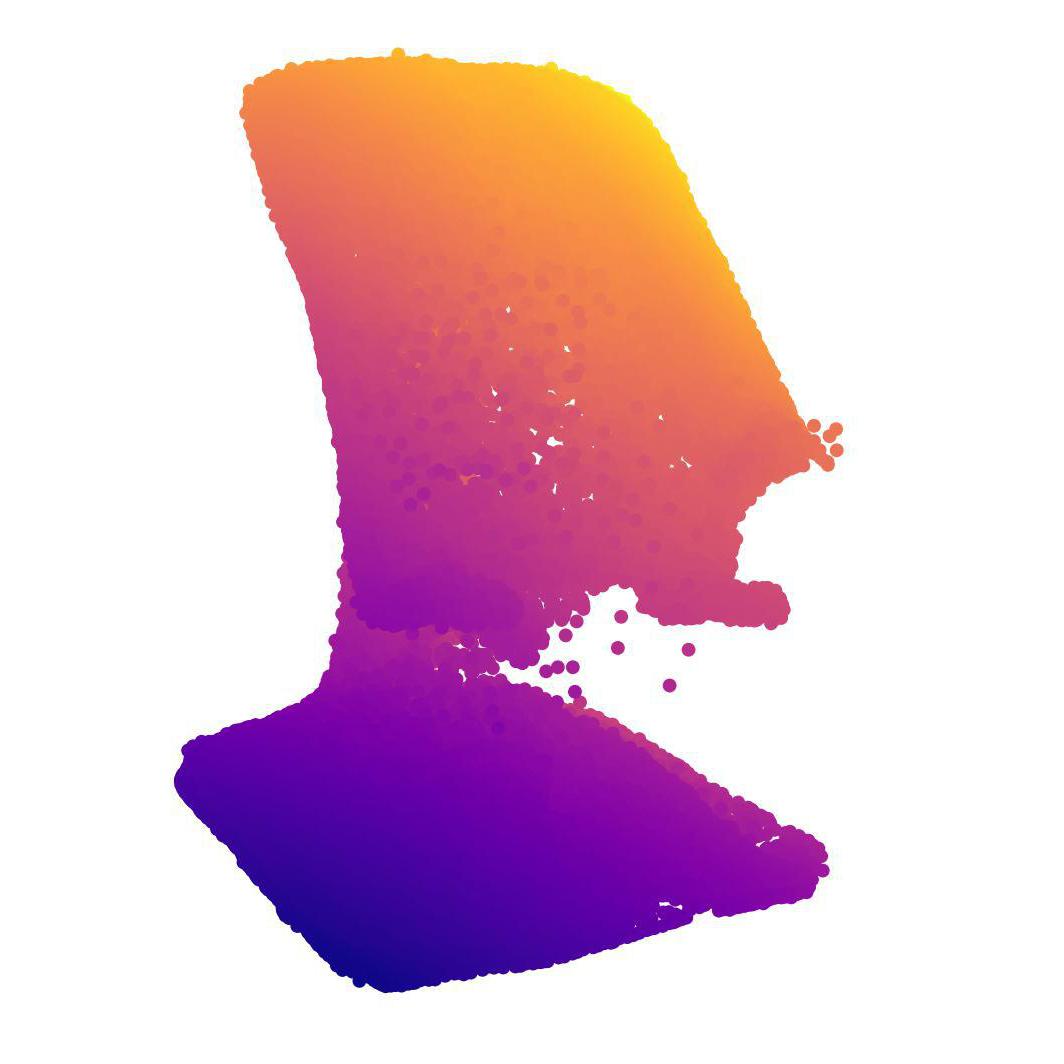}
    & \includegraphics[width=\hsize,valign=m]{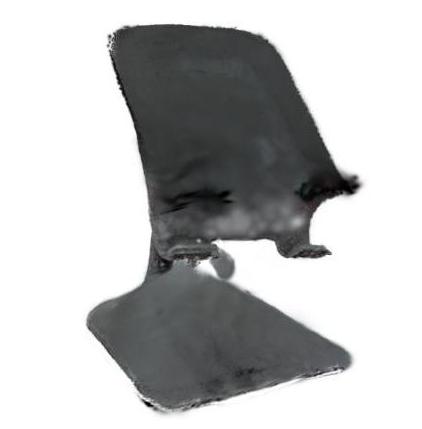}
    & \includegraphics[width=\hsize,valign=m]{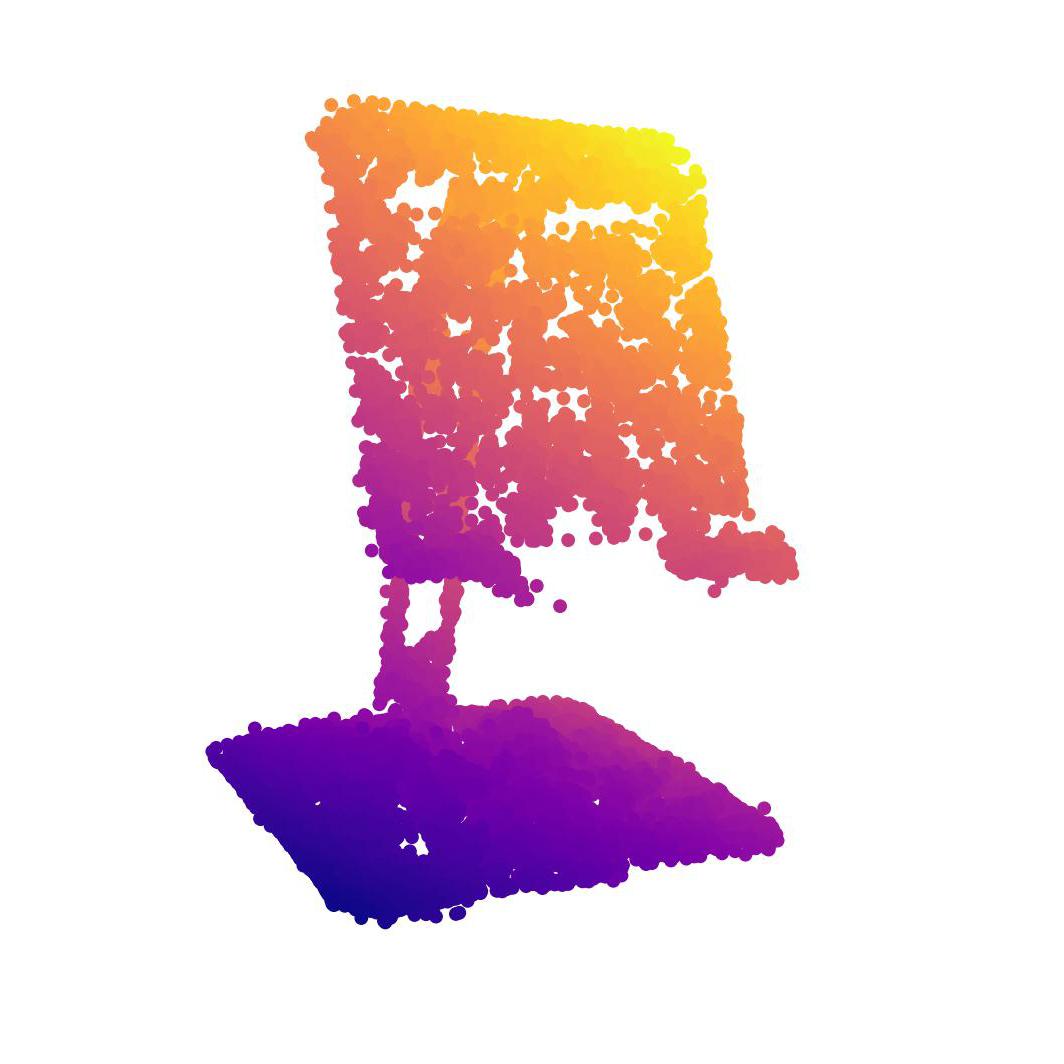}
    & \includegraphics[width=\hsize,valign=m]{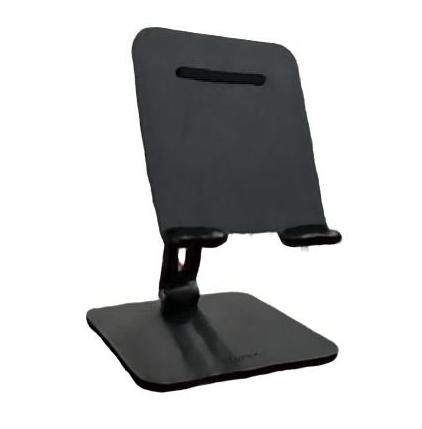}
    & \includegraphics[width=\hsize,valign=m]{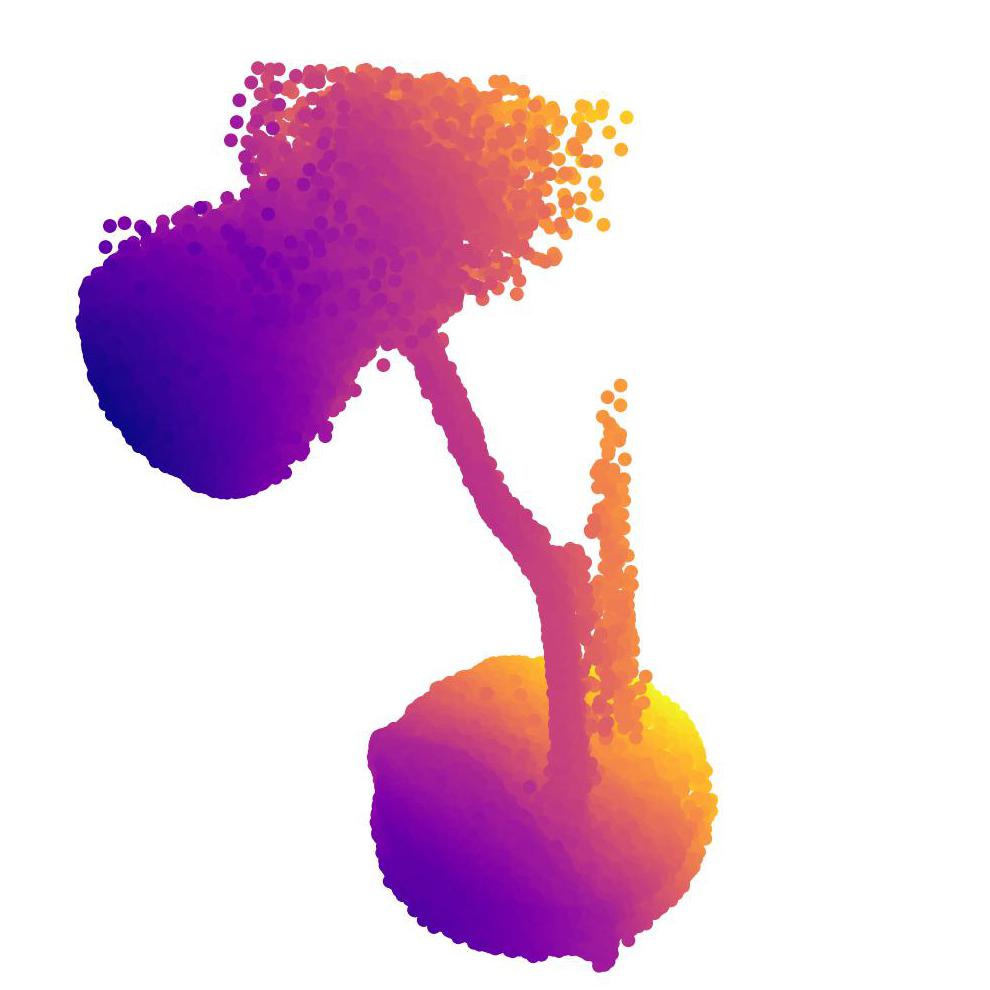}
    & \includegraphics[width=\hsize,valign=m]{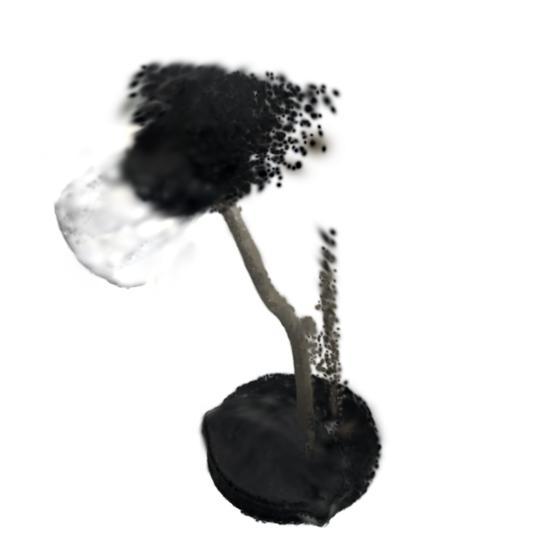}
    & \includegraphics[width=\hsize,valign=m]{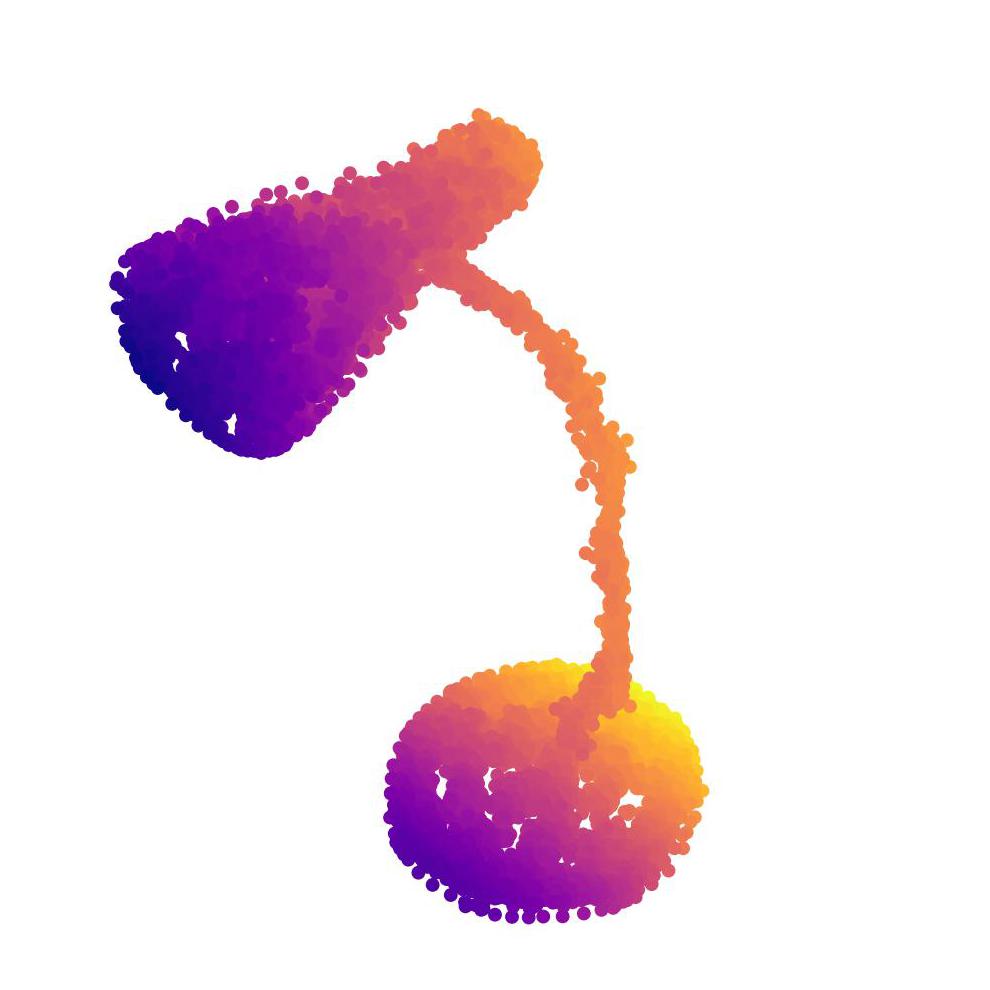}
    & \includegraphics[width=\hsize,valign=m]{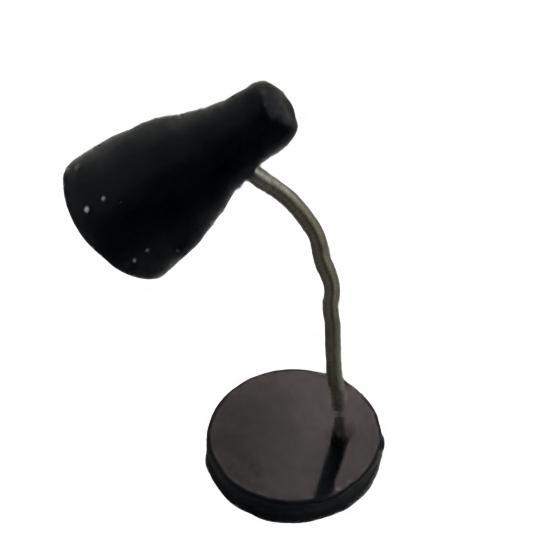}
    \\
    & \includegraphics[width=\hsize,valign=m]{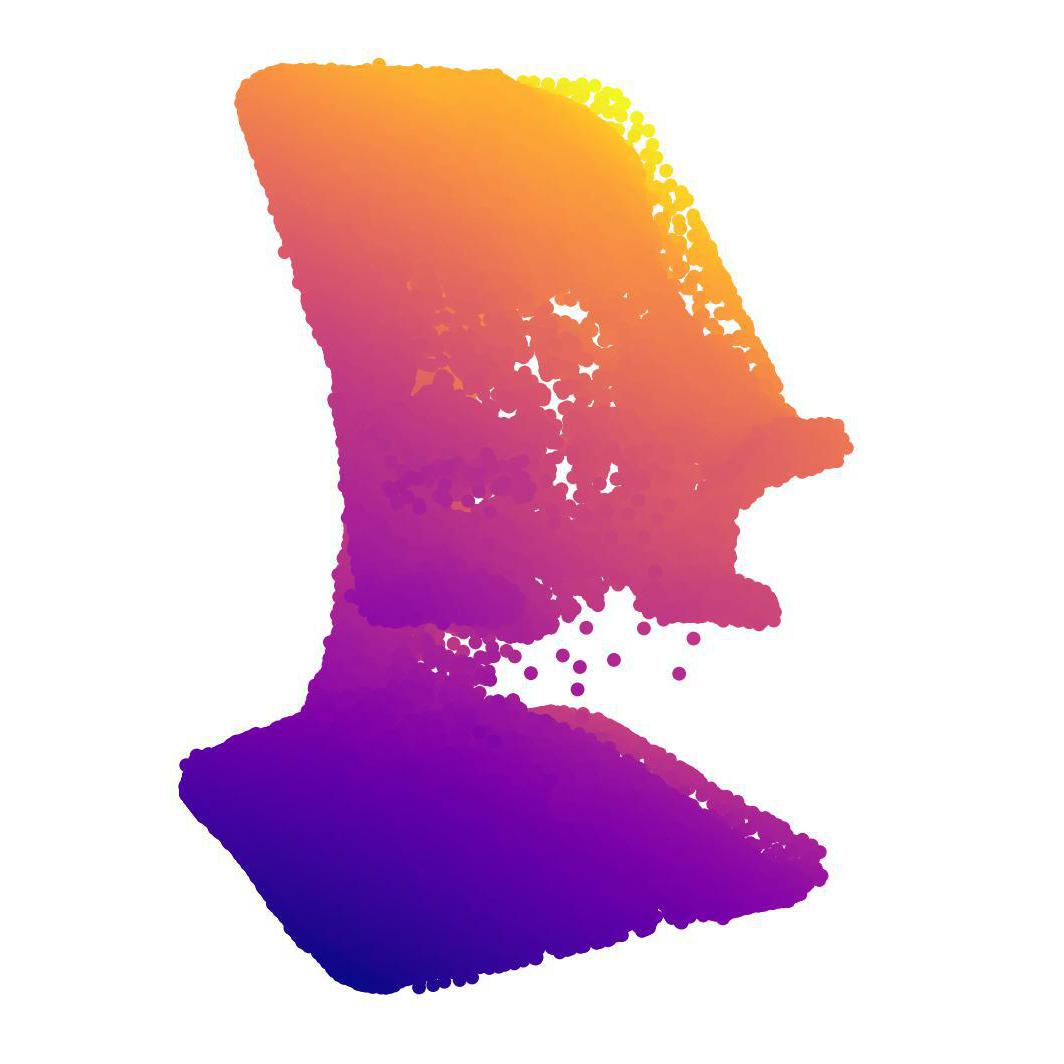}
    & \includegraphics[width=\hsize,valign=m]{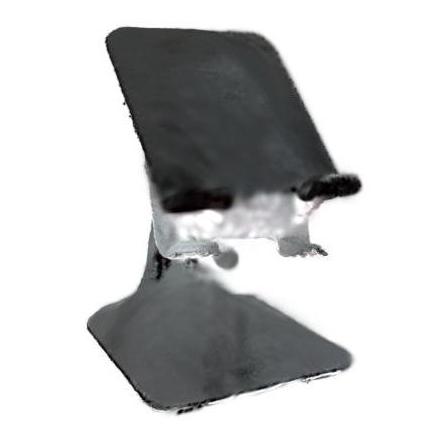}
    & \includegraphics[width=\hsize,valign=m]{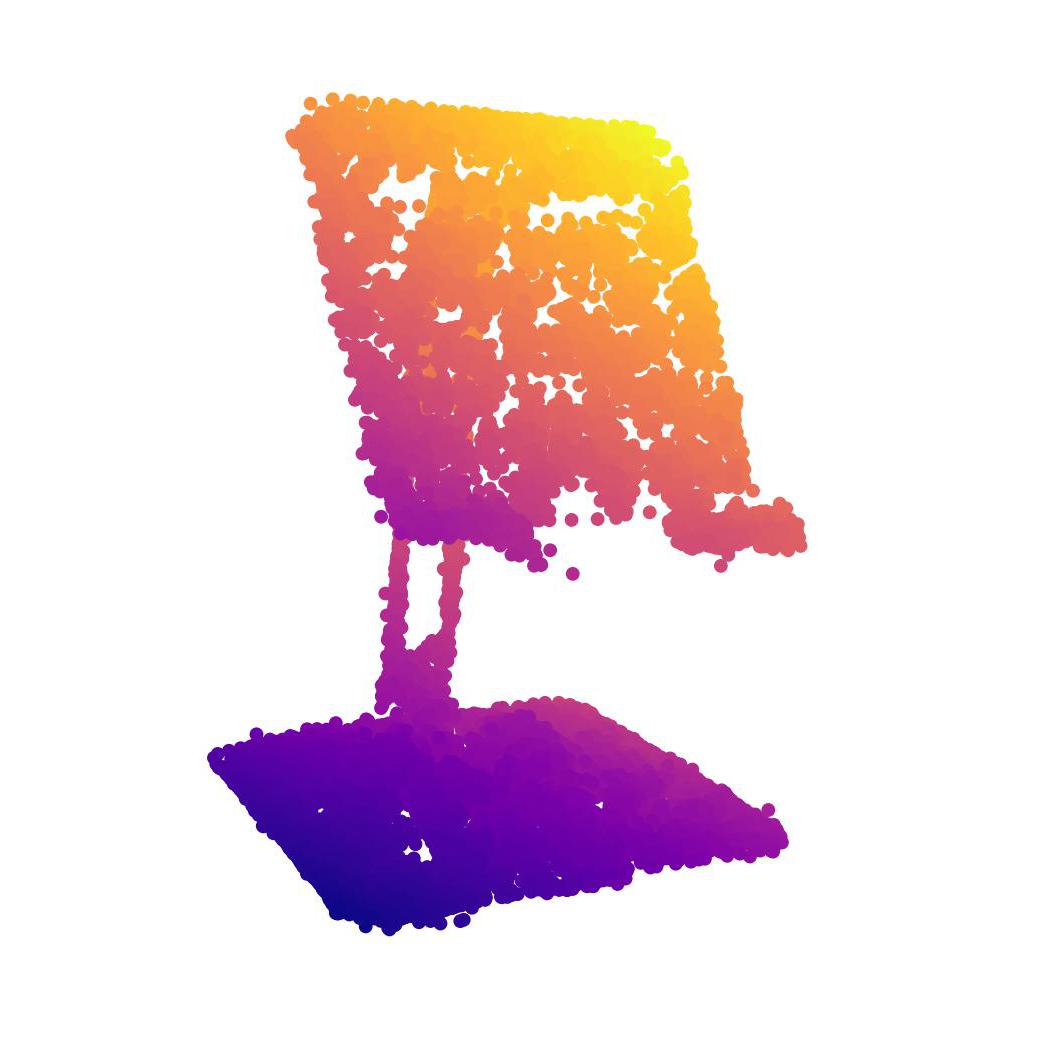}
    & \includegraphics[width=\hsize,valign=m]{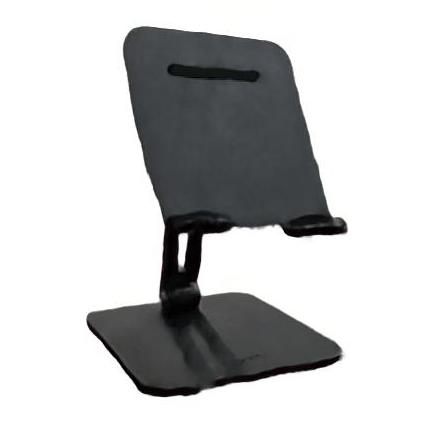}
    & \includegraphics[width=\hsize,valign=m]{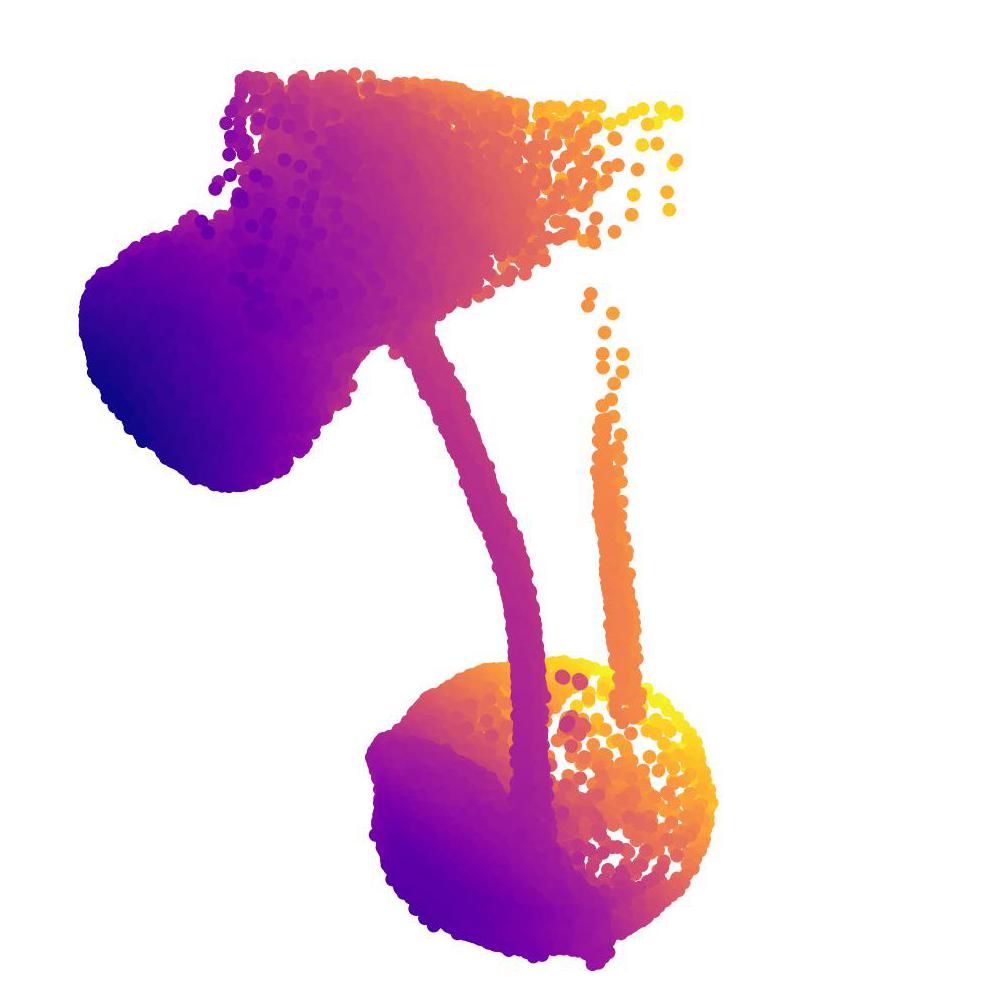}
    & \includegraphics[width=\hsize,valign=m]{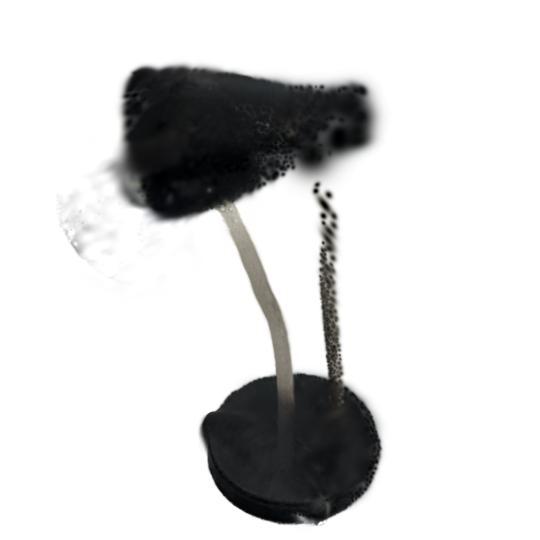}
    & \includegraphics[width=\hsize,valign=m]{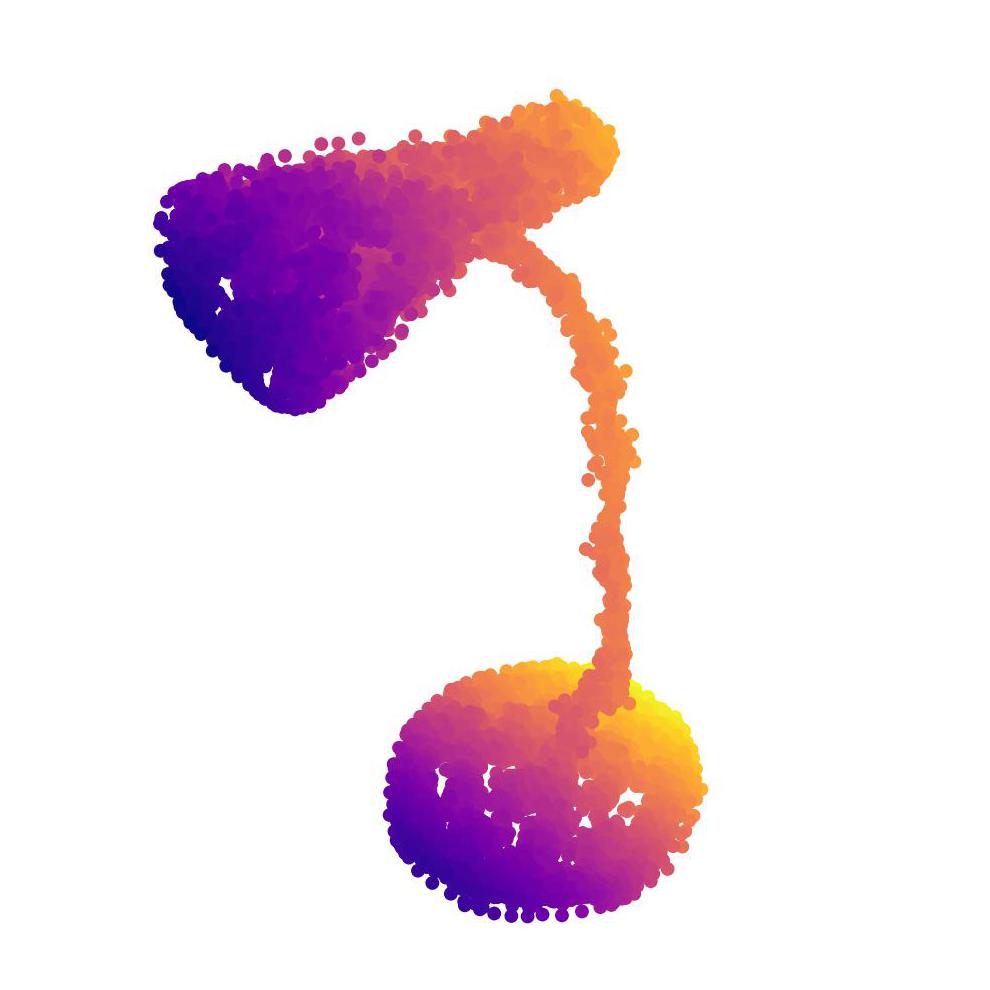}
    & \includegraphics[width=\hsize,valign=m]{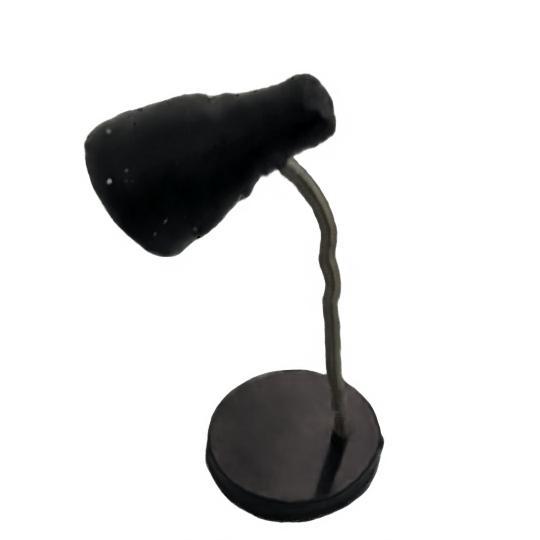}
    \\

    \rotatebox[origin=c]{90}{End}
    & \includegraphics[width=\hsize,valign=m]{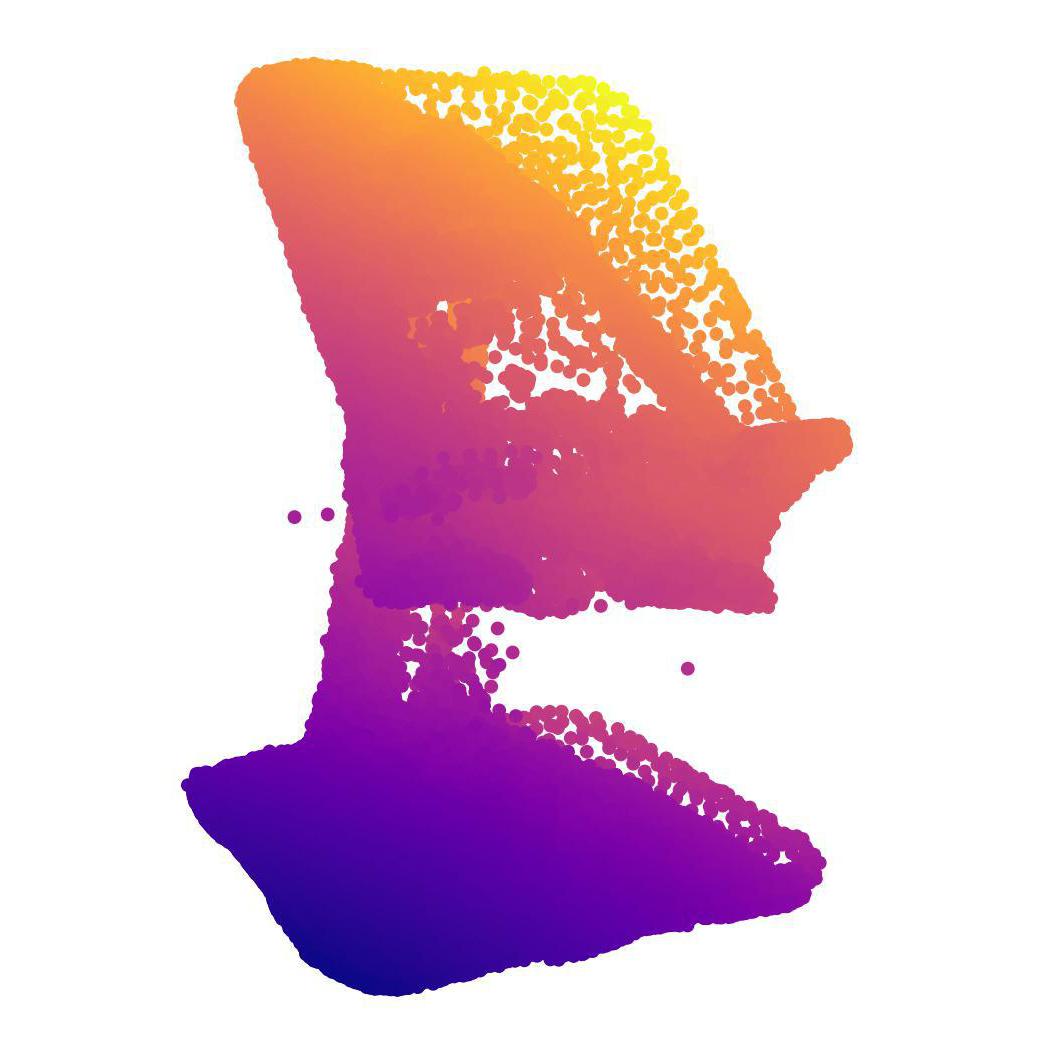}
    & \includegraphics[width=\hsize,valign=m]{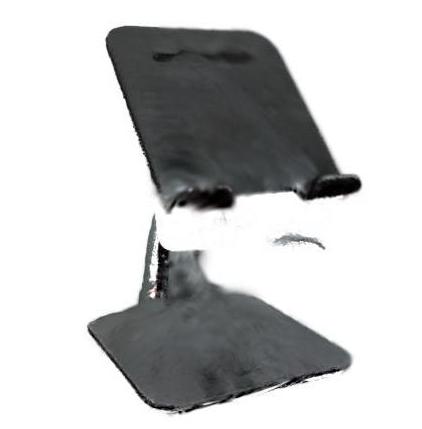}
    & \includegraphics[width=\hsize,valign=m]{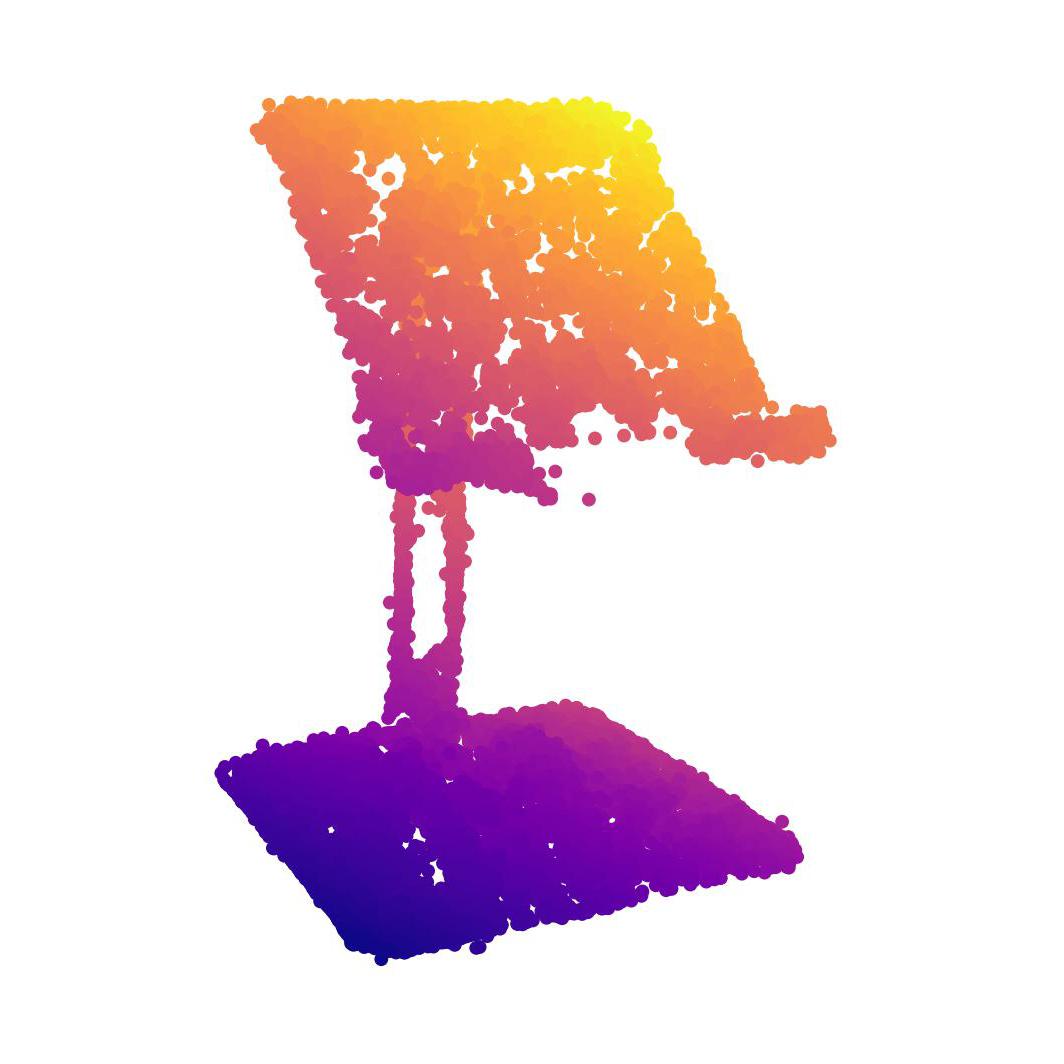}
    & \includegraphics[width=\hsize,valign=m]{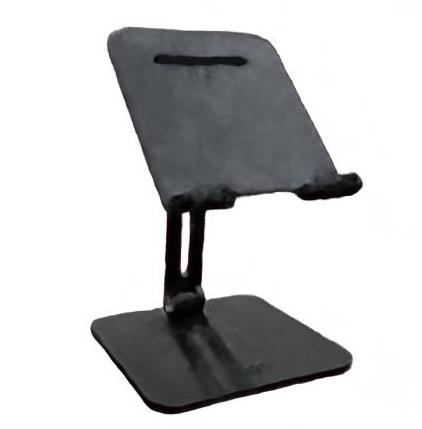}
    & \includegraphics[width=\hsize,valign=m]{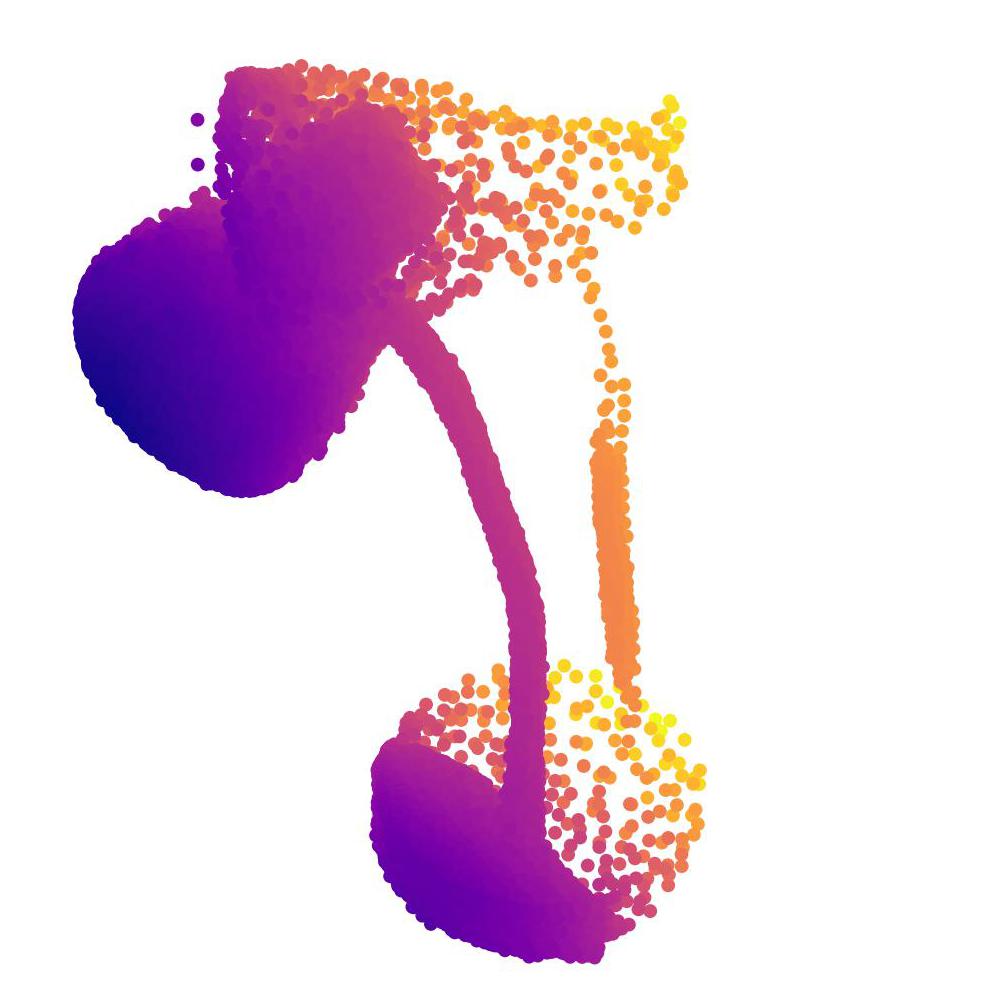}
    & \includegraphics[width=\hsize,valign=m]{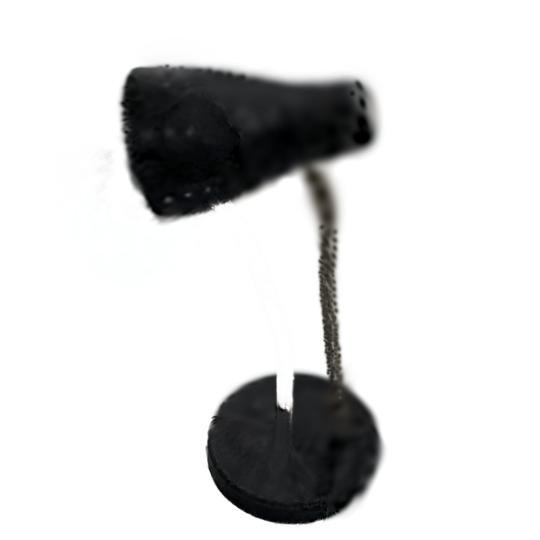}
    & \includegraphics[width=\hsize,valign=m]{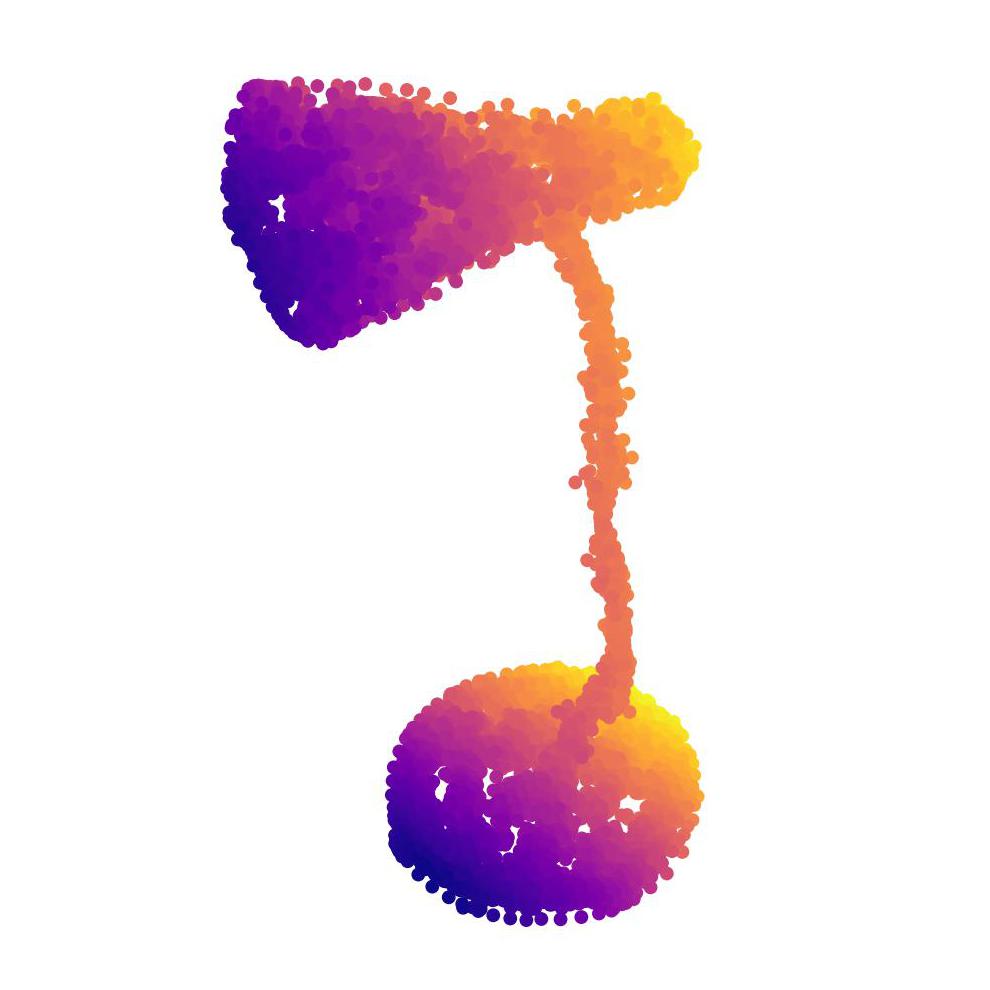}
    & \includegraphics[width=\hsize,valign=m]{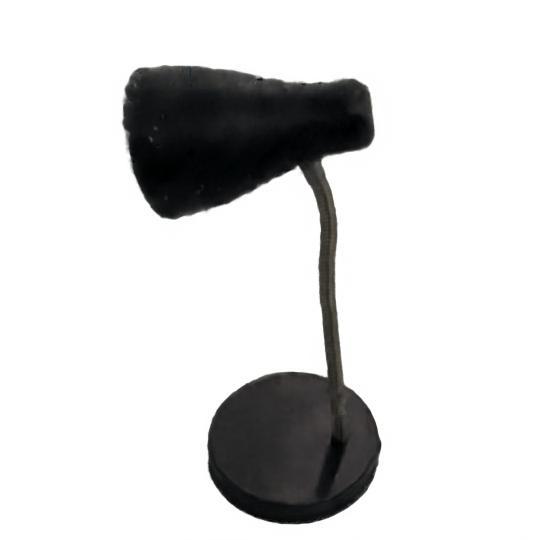}
    \\
    \rotatebox[origin=c]{90}{Trajectory}
    & \multicolumn{2}{c}{\includegraphics[width=0.12\linewidth,valign=m]{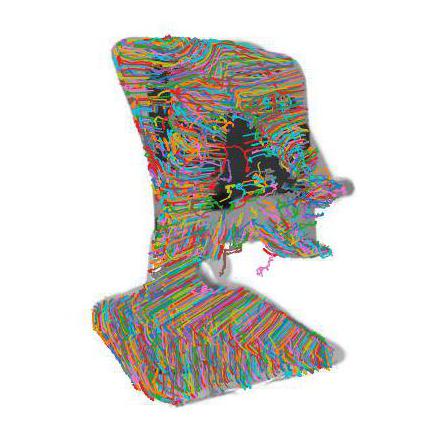}}
    & \multicolumn{2}{c}{\includegraphics[width=0.12\linewidth,valign=m]{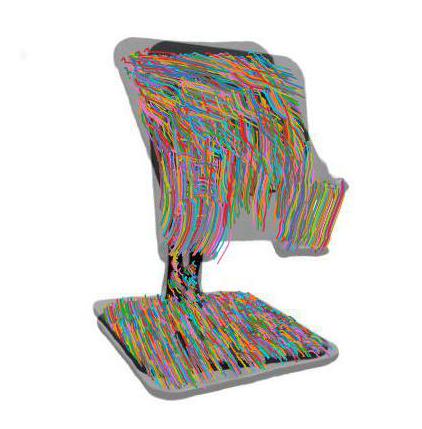}} 
    & \multicolumn{2}{c}{\includegraphics[width=0.12\linewidth,valign=m]{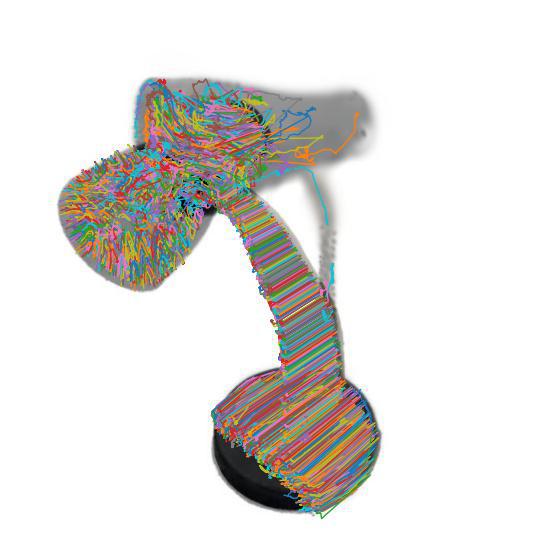}}
    & \multicolumn{2}{c}{\includegraphics[width=0.12\linewidth,valign=m]{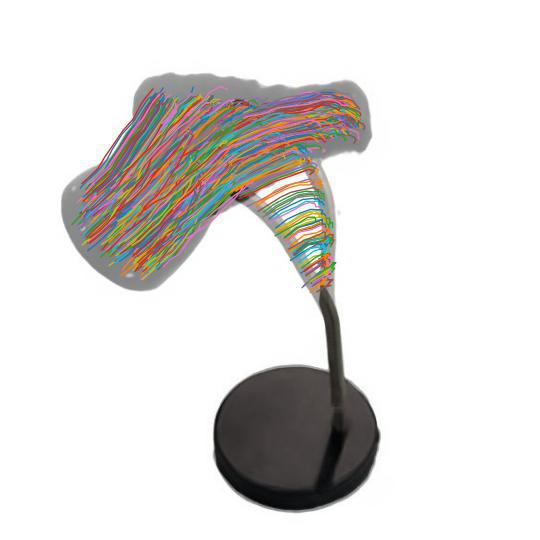}} 
    
\end{tabularx}
\caption{    
Qualitative comparison of 3D scene interpolation from start to end state using real-world scenes. Both methods start by training a static model for the start state and subsequently finetune it towards the end state, all without any intermediate supervision. Dynamic Gaussian~\cite{Luiten2023Dynamic3G} struggles with significant scene changes, as shown by both scenes where only a portion of the points move to the right place. In contrast, PAPR in Motion successfully handles these challenging scenarios, producing smooth and realistic interpolations between states.
}
\label{fig:qualitative_1}
\end{figure*}

To evaluate the effectiveness of our method on point-level 3D scene interpolation, we experiment on both synthetic and real-world scenes. The synthetic scenes include six scenes from a combination of the NeRF Synthetic Dataset~\cite{Mildenhall2020NeRFRS} and the Objaverse dataset~\cite{Deitke2022ObjaverseAU}. In addition, we manually captured two real-world scenes.
The scenes we use represent a wide range of complex scene changes, encompassing rigid and non-rigid transformations, large displacements, and scenarios with multiple parts in motion.

Given the novelty of this task and the absence of existing methods for the task, we compare our method to the leading point-based method for dynamic scenes, Dynamic Gaussian~\cite{Luiten2023Dynamic3G}.
To pre-train Dynamic Gaussian for the initial state, we begin with a point cloud constructed by COLMAP, containing between 100,000 to 500,000 points for each scene. 
In contrast, the pre-training of PAPR for each scene uses a significantly sparser point cloud of just 30,000 points, which are initialized randomly. This approach is feasible due to PAPR's ability in learning point clouds from scratch and its capability to learn parsimonious yet effective representations.
Regarding our method's parameters, we set $\lambda = 0.01$, $\lambda_\text{rigid}=5$, $m=100$ and a $k$ that ranges from 70 to 300, depending on the specific scene. 

We evaluate each method qualitatively on both the geometry and the renderings of the scenes. This evaluation spans the entire duration of the interpolation, from the start state to the end state.
Additionally, we visualize and compare the interpolation trajectories of points within the scene, providing more insights into the quality of the interpolated motion generated by each method. For the quantitative evaluation, due to the absence of an existing metric for evaluating scene interpolation quality, we introduce a novel metric termed the \textit{scene interpolation metric}.

\paragraph{Scene Interpolation Metric}
This metric is designed to assess the overall quality of interpolation by evaluating the quality at different intermediate steps and aggregating these individual quality assessments to arrive at an overall score.

Specifically, the evaluation is performed sequentially at each model checkpoint throughout the interpolation process. For the $t^{th}$ checkpoint, we compute its progress $\alpha_t$ using Equation~\ref{eqn:progress}. In our evaluations, we select 24 evenly spaced checkpoints that span the interpolation process.

Given the lack of ground truth geometries and appearances for intermediate steps, at each step $t$, we assess the quality $f_t$ of interpolation $P_t$ by comparing it separately to the ground truth at the start and end states, $P_{start}$ and $P_{end}$:
\[
    f_t = (1 - \alpha_t)d(P_{start}, P_t) + \alpha_t d(P_{end}, P_t)
\]
Here, $P$ represents either the geometry (as a point cloud) or the appearance (as a set of scene renderings using camera poses $\mathcal{C}$). For the ground truth point clouds at the start and end states, we use those derived from training the corresponding static 3D scene reconstruction method, namely PAPR~\cite{zhang2023papr} for our method and Gaussian Splatting~\cite{kerbl20233d} for Dynamic Gaussian~\cite{Luiten2023Dynamic3G}. For appearance evaluation, we choose $\mathcal{C}$ to consist of 128 viewpoints encircling the scene in 3D. The distance function $d(\cdot, \cdot)$ denotes either the Chamfer distance (CD) or the Earth Mover's distance (EMD) for geometry evaluation and the Fréchet inception distance (FID)~\cite{Heusel2017GANsTB} for appearance evaluation.

The overall scene interpolation metric is derived by aggregating the measurements across all steps:
\[
    \sum_{t=1}^{T} (\alpha_{t} - \alpha_{t-1})\cdot\frac{(f_{t} + f_{t-1})}{2}
\]
We use Scene Interpolation CD and EMD (SI-CD, SI-EMD) for geometry evaluation and Scene Interpolation FID (SI-FID) for appearance evaluation.

\subsection{Quantitative Results}

We include the scene interpolation metric scores in Table~\ref{tab:main}. As shown, our method significantly outperforms the baseline across all metrics in both synthetic and real-world scenes. These results demonstrate our method's capability to interpolate scene appearance and geometry effectively between the start and end states. 

\begin{table*}[ht]
    \centering
    \footnotesize
    \resizebox{\linewidth}{!}{
    \begin{tabular}{lcccccccc|ccc}
    \toprule
    \multirow{2}{*}{Metric} & \multirow{2}{*}{Method} & \multicolumn{7}{c}{Synthetic Scenes} & \multicolumn{3}{c}{Real-world Scenes} \\
    & & Butterfly & Crab & Dolphin & Giraffe & Lego Bulldozer & Lego Man & Avg & Stand & Lamp & Avg \\
    \midrule
    \multirow{2}{*}{SI-FID $\downarrow$} & Dynamic Gaussian~\cite{Luiten2023Dynamic3G} & $352.43$ & $106.82$ & $134.01$ & $199.76$& $173.07$ & $201.74$ & $194.64$ & $293.60$ & $266.87$ & $280.24$ \\
    & PAPR in Motion (Ours) & \boldsymbol{$99.06$} & \boldsymbol{$76.95$} & \boldsymbol{$120.27$} & \boldsymbol{$173.89$} & \boldsymbol{$110.78$} & \boldsymbol{$140.84$} & \boldsymbol{$120.30$} & \boldsymbol{$182.74$} & \boldsymbol{$216.74$} & \boldsymbol{$199.74$} \\
    \midrule
    \multirow{2}{*}{SI-CD $\downarrow$} & Dynamic Gaussian~\cite{Luiten2023Dynamic3G} & $45.63$ & $1.66$ & $0.66$ & $1.06$& $6.31$ & $2.74$ & $9.68$ & $20.27$ & $53.99$ & $37.13$ \\
    & PAPR in Motion (Ours) & \boldsymbol{$17.98$} & \boldsymbol{$1.32$} & \boldsymbol{$0.25$} & \boldsymbol{$0.27$} & \boldsymbol{$1.99$} & \boldsymbol{$1.82$} & \boldsymbol{$3.94$} & \boldsymbol{$8.91$} & \boldsymbol{$9.19$} & \boldsymbol{$9.05$} \\
    \midrule
    \multirow{2}{*}{SI-EMD $\downarrow$} & Dynamic Gaussian~\cite{Luiten2023Dynamic3G} & $56.74$ & $17.33$ & $22.29$ & $18.04$& $55.62$ & $18.95$ & $31.49$ & $70.30$ & $120.32$ & $95.31$ \\
    & PAPR in Motion (Ours) & \boldsymbol{$35.85$} & \boldsymbol{$9.45$} & \boldsymbol{$2.67$} & \boldsymbol{$5.06$} & \boldsymbol{$8.78$} & \boldsymbol{$13.89$} & \boldsymbol{$12.62$} & \boldsymbol{$44.52$} & \boldsymbol{$61.52$} & \boldsymbol{$53.02$} \\
    \bottomrule
  \end{tabular}
  }
  \vspace{2pt}
  \caption{Comparison of scene interpolation appearance rendering quality, measured by Scene Interpoation FID (SI-FID), and geometry quality, measured by Scene Interpolation Chamfer distance (SI-CD) and Scene Interpolation Earth Mover’s distance (SI-EMD). Lower scores for all three metrics indicate better quality. The SI-CD and SI-EMD results are reported on a scale of $10^{-3}$. 
  Our method consistently outperforms the baseline on both synthetic and real-world scenes.
  }
    \label{tab:main}
\end{table*}

\subsection{Qualitative Results}

Figures~\ref{fig:qualitative} and \ref{fig:qualitative_1} present a qualitative comparison between our method, PAPR in Motion, and the baseline Dynamic Gaussian~\cite{Luiten2023Dynamic3G} on various synthetic and real-world scenes. PAPR in Motion demonstrates superior capabilities in generating plausible and seamless scene interpolations. It effectively handles a diverse range of motion types, including rigid transformations seen in the Lego man and non-rigid bending motions observed in the dolphin in the synthetic scenes (Fig.\ref{fig:qualitative}). Additionally, our method proves effective in scenes with significant changes, such as raising the tablet stand and the lamp in the real-world scenes (Fig.\ref{fig:qualitative_1}).
In contrast, the baseline method fails to maintain consistent scene geometry and appearance, resulting in unnatural interpolations. For more qualitative comparisons, please refer to the supplementary materials.

\subsection{Ablation Study}

\begin{figure}[ht]
\footnotesize
\begin{tabularx}{\linewidth}{lYYYY}
  & Start & \multicolumn{2}{c}{Intermediate} & End  \\
\multirow{2}{*}{\rotatebox[origin=c]{90}{Full Model}} 
        & \includegraphics[width=\hsize,valign=m]{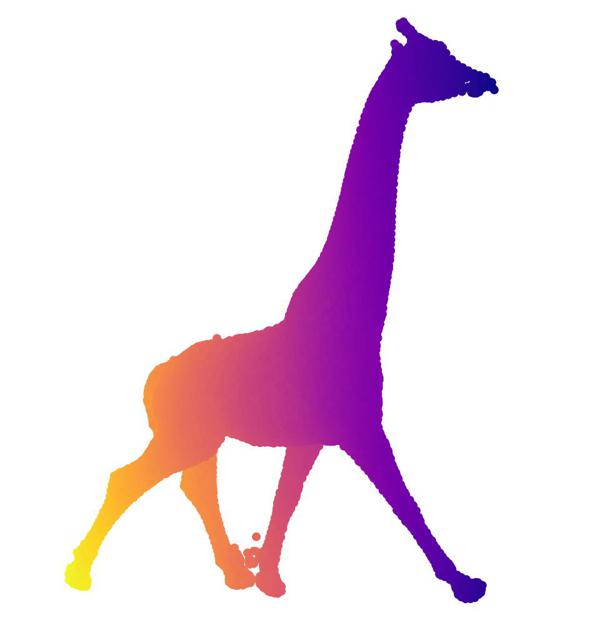}
        & \includegraphics[width=\hsize,valign=m]{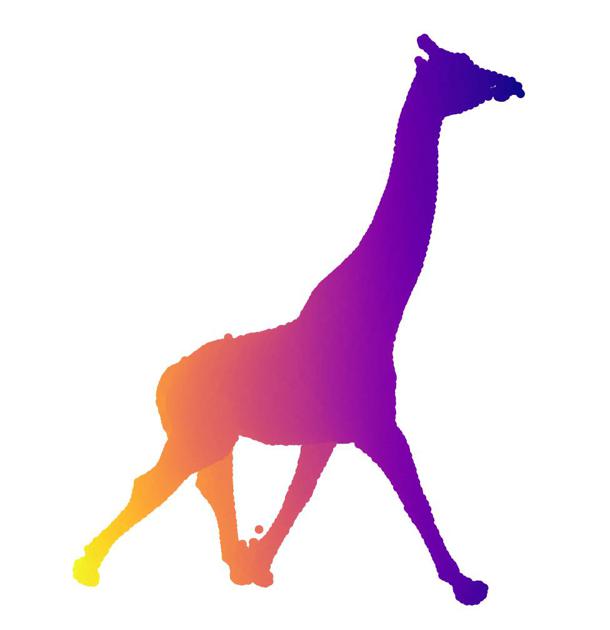}
        & \includegraphics[width=\hsize,valign=m]{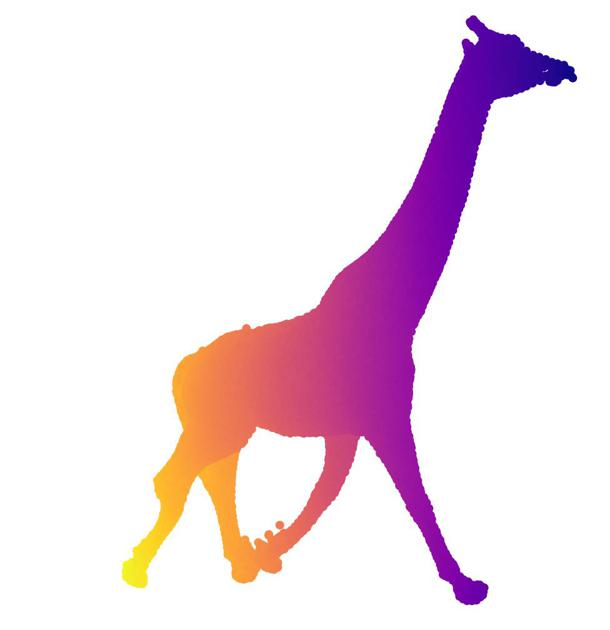}
        & \includegraphics[width=\hsize,valign=m]{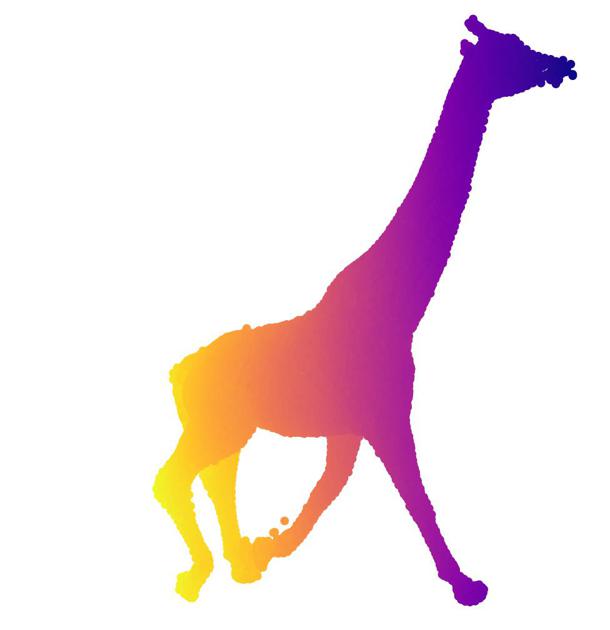}
         \\
        & \includegraphics[width=\hsize,valign=m]{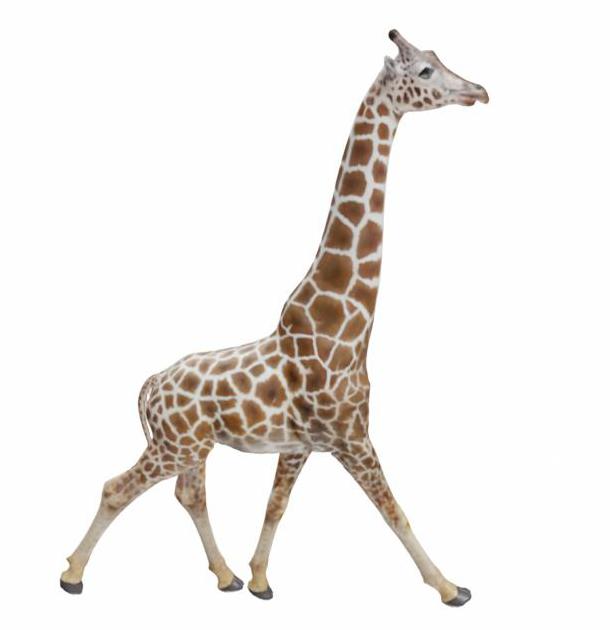}
        & \includegraphics[width=\hsize,valign=m]{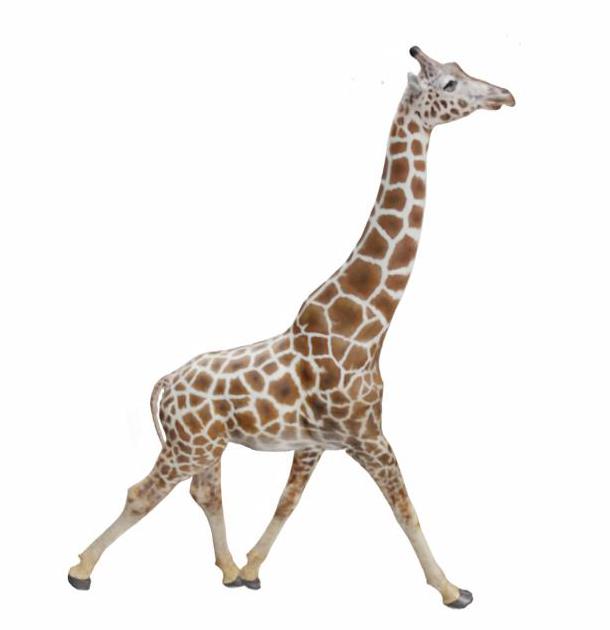}
        & \includegraphics[width=\hsize,valign=m]{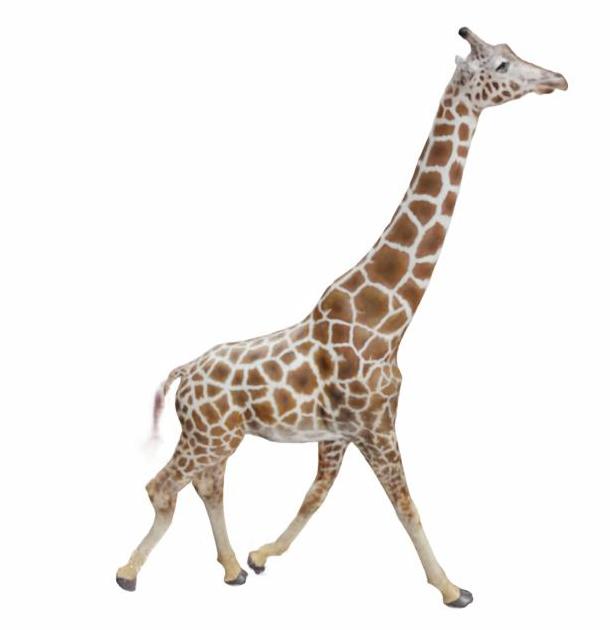}
        & \includegraphics[width=\hsize,valign=m]{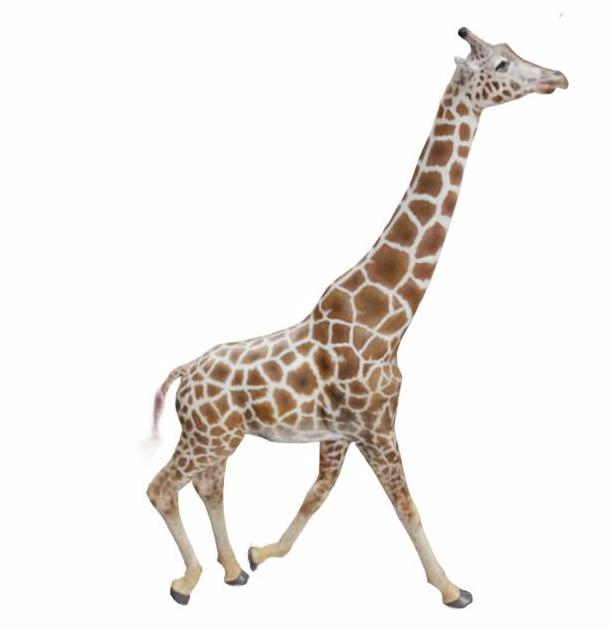}
         \\
         \midrule
\multirow{2}{*}{\rotatebox[origin=c]{90}{No LDAS}} 
        & \includegraphics[width=\hsize,valign=m]{figs/ab_giraffe_pc_start.jpg}
        & \includegraphics[width=\hsize,valign=m]{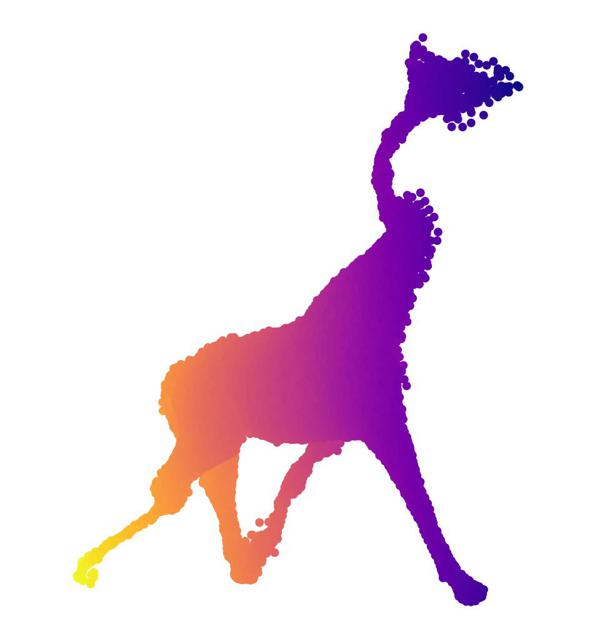}
        & \includegraphics[width=\hsize,valign=m]{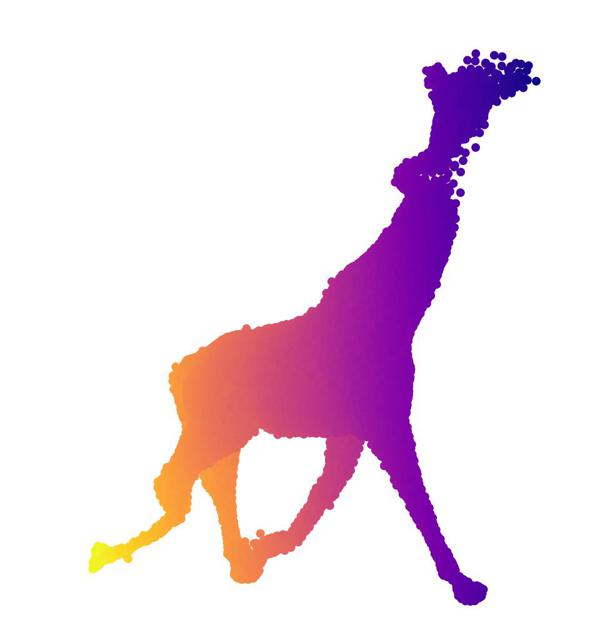}
        & \includegraphics[width=\hsize,valign=m]{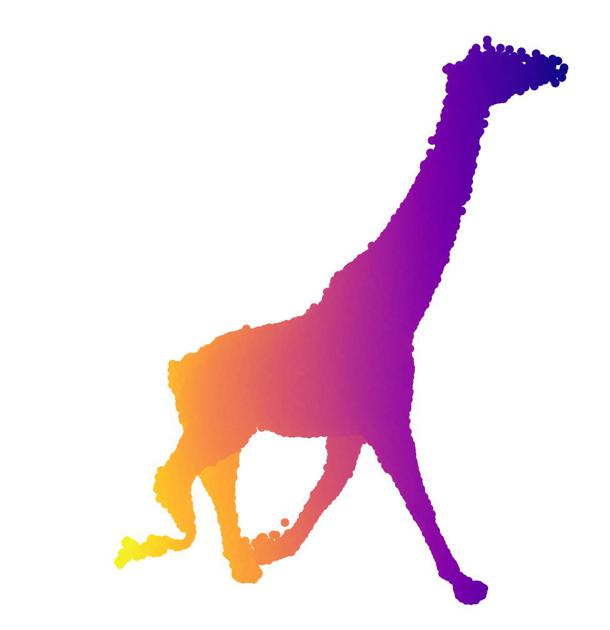}
        \\
        & \includegraphics[width=\hsize,valign=m]{figs/ab_giraffe_rgb_start.jpg}
        & \includegraphics[width=\hsize,valign=m]{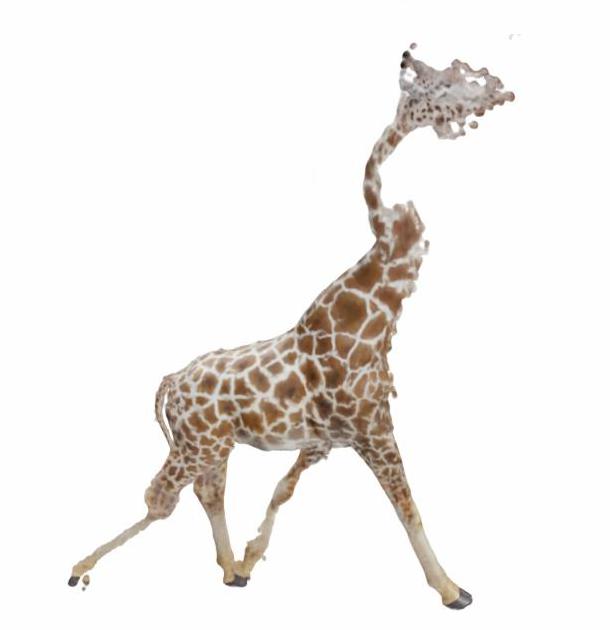}
        & \includegraphics[width=\hsize,valign=m]{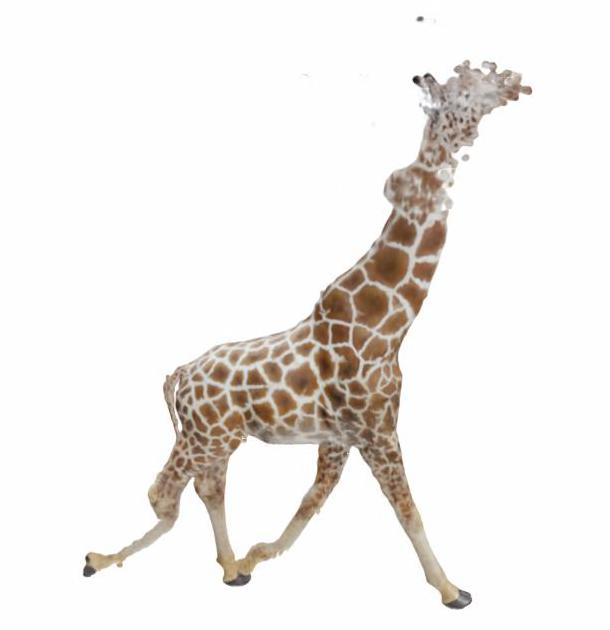}
        & \includegraphics[width=\hsize,valign=m]{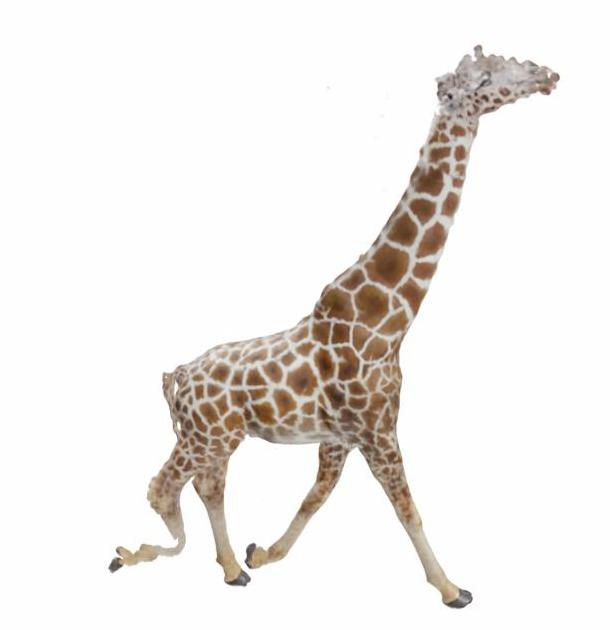}
        \\\midrule
\multirow{2}{*}{\rotatebox[origin=c]{90}{No LDAS, $\mathcal{L}_{rigid}$}} 
        & \includegraphics[width=\hsize,valign=m]{figs/ab_giraffe_pc_start.jpg}
        & \includegraphics[width=\hsize,valign=m]{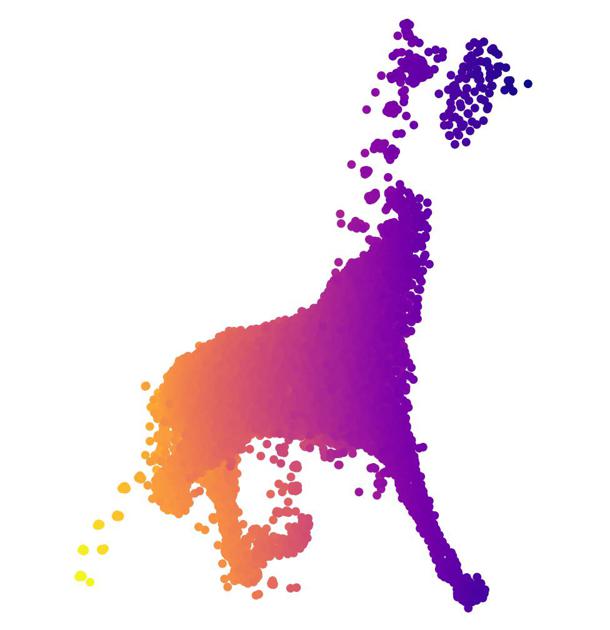}
        & \includegraphics[width=\hsize,valign=m]{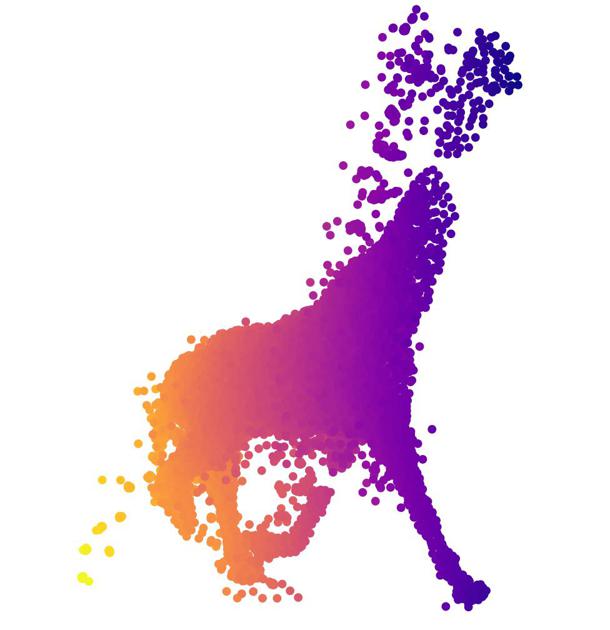}
        & \includegraphics[width=\hsize,valign=m]{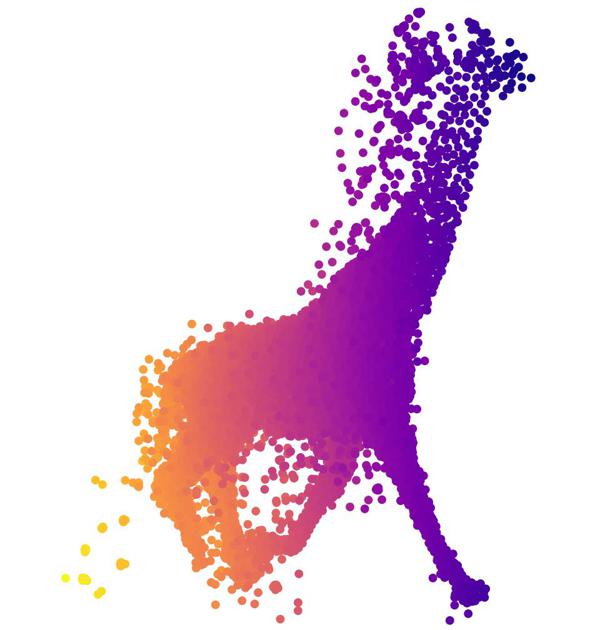}
        \\
        & \includegraphics[width=\hsize,valign=m]{figs/ab_giraffe_rgb_start.jpg}
        & \includegraphics[width=\hsize,valign=m]{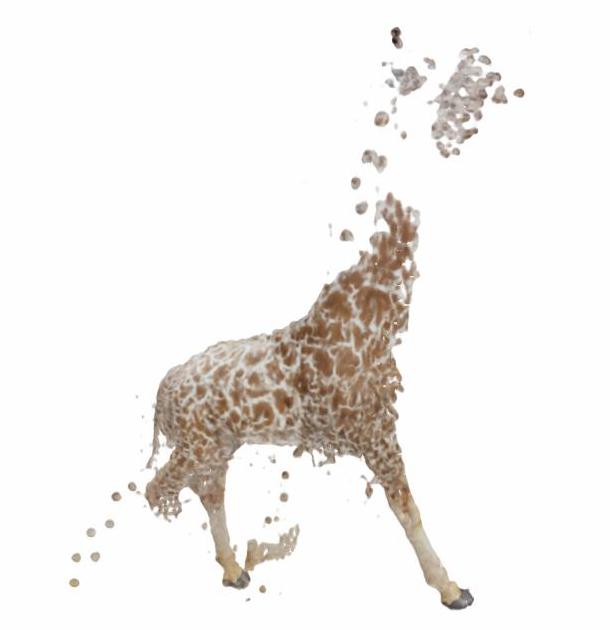}
        & \includegraphics[width=\hsize,valign=m]{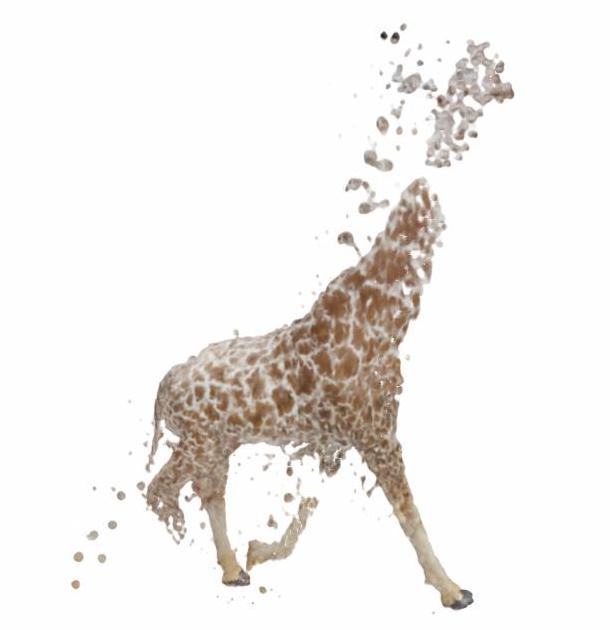}
        & \includegraphics[width=\hsize,valign=m]{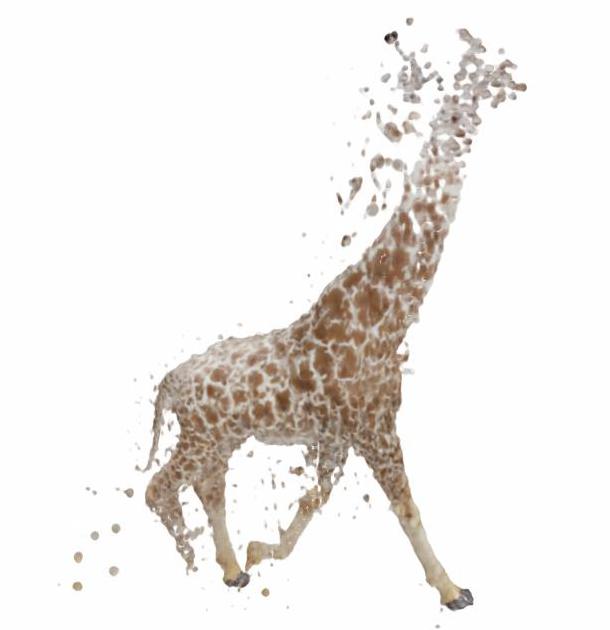}
        
\end{tabularx}
\caption{
    Ablation study on the impact of proposed regularization terms, where we incrementally remove the local displacement averaging step (LDAS) and the local distance preserving loss ($\mathcal{L}_{rigid}$). The results show that removing LDAS leads to unwanted deformations in the part geometry. When $\mathcal{L}_{rigid}$ is also removed, there is a further detrimental effect, with points drifting away from the object surface. These findings validates the critical role of both regularization techniques in maintaining the integrity and the quality of our model's interpolations.
}
\label{fig:ablation}
\end{figure}

Figure~\ref{fig:ablation} presents the results of our ablation study, where we incrementally remove the local displacement averaging step (LDAS) and the local distance preserving loss $\mathcal{L}_{rigid}$. The study reveals that upon the removal of the local averaging step, the points, while maintaining cohesiveness, still exhibit undesirable deformations in the part geometry, such as in the neck and rear leg of the giraffe. Further, the elimination of the loss $\mathcal{L}_{rigid}$ leads to excessive drift of the points, resulting in a failure to preserve the structural integrity of the object's shape. These results underscore the vital role of our regularization techniques in maintaining scene geometry and demonstrate their necessity for facilitating smooth and plausible scene interpolations.

\section{Discussion and Conclusion}

\paragraph{Limitations}

While our method showcases promising results, it does have certain limitations, which open avenues for future research. Specifically, since our model is pretrained on the start state, it might have challenges rendering areas not visible in the initial state but are present in the end state, which can happen when new objects enter the scene. Additionally, since we do not assume any prior knowledge on the scene object, such as its semantics, the generated motion may not precisely replicate real-world motion. These challenges, however, present exciting opportunities for further exploration and expansion of our method.

\paragraph{Conclusion}
In this paper, we introduce the novel task of point-level 3D scene interpolation between two distinct scene states. Addressing this novel and challenging task, we leverage the capabilities of the Proximity Attention Point Rendering (PAPR) technique to develop a method that can create plausible and smooth interpolations. Notably, our approach proves to be effective even in cases involving substantial non-rigid scene changes. Our work establishes a baseline on this challenging task and paves the way for further exploration and advancement.

\paragraph{Acknowledgements}
This research was enabled in part by support provided by NSERC, the BC DRI Group and the Digital Research Alliance of Canada.
\clearpage
\newpage
{
    \small
    \bibliographystyle{ieeenat_fullname}
    \bibliography{main}
}
\clearpage
\newpage

\setcounter{page}{1}
\maketitlesupplementary
\appendix

\section{Video Results}
Please visit our \href{https://niopeng.github.io/PAPR-in-Motion/}{project website} for the video results, where we include animations of our method and comparisons to the baseline.

\section{Sensitivity Analysis}

\begin{figure}[ht]
\footnotesize
\begin{tabularx}{\linewidth}{lYYYY}
  & Start & \multicolumn{2}{c}{Intermediate} & End  \\
\multirow{2}{*}[1ex]{\rotatebox[origin=c]{90}{$k=50$}}
        & \includegraphics[width=\hsize,valign=m]{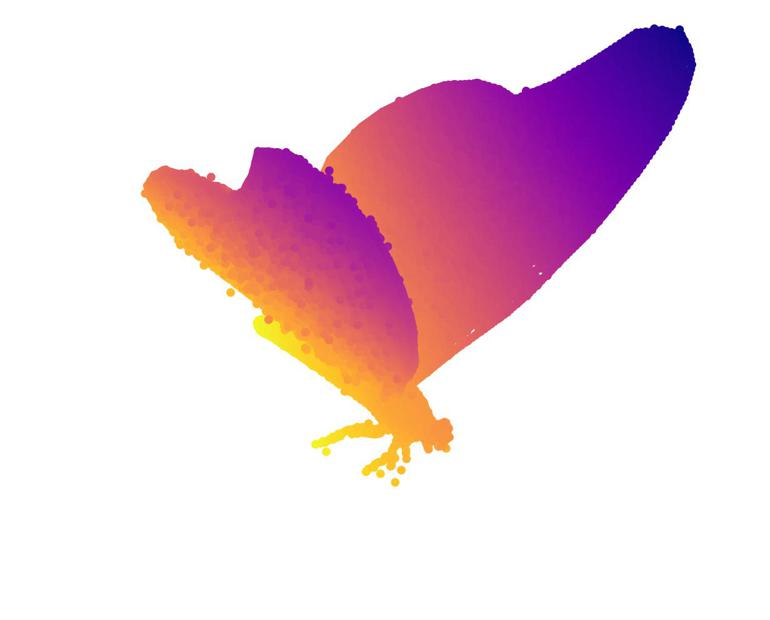}
        & \includegraphics[width=\hsize,valign=m]{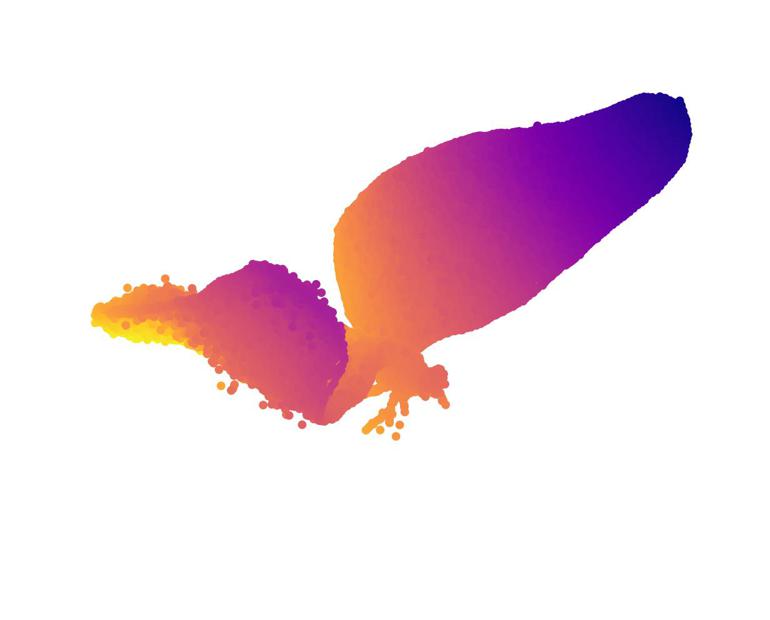}
        & \includegraphics[width=\hsize,valign=m]{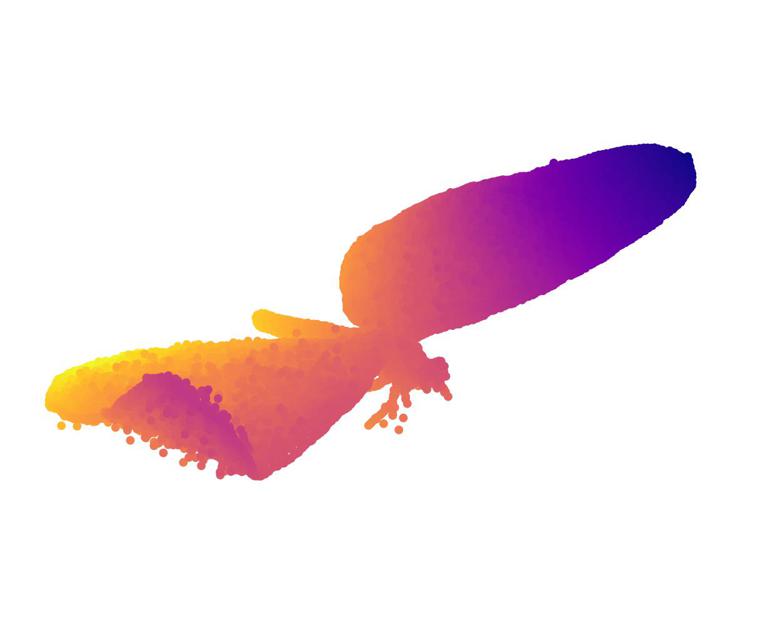}
        & \includegraphics[width=\hsize,valign=m]{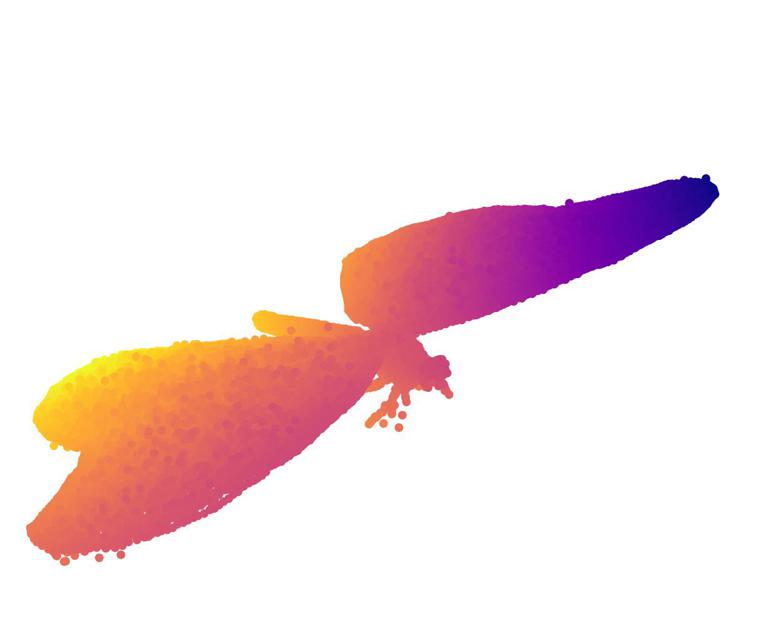}
         \\
        & \includegraphics[width=\hsize,valign=m]{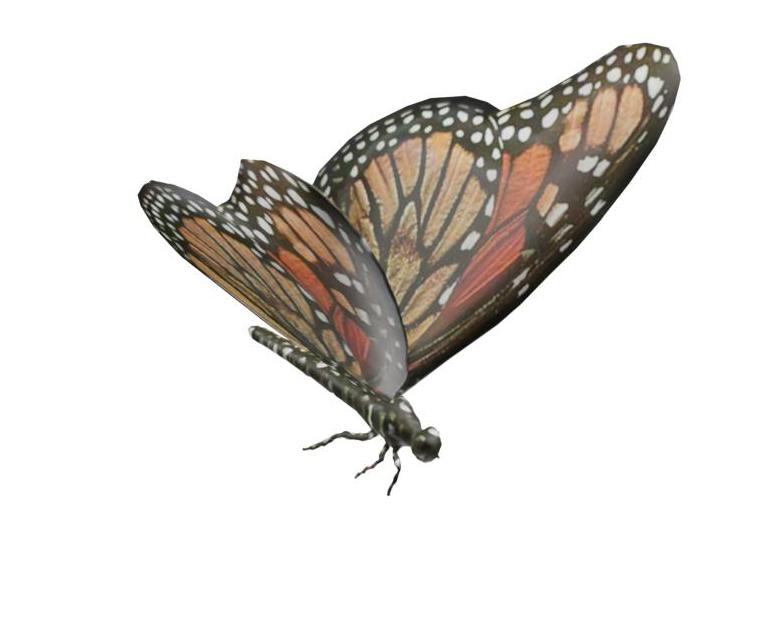}
        & \includegraphics[width=\hsize,valign=m]{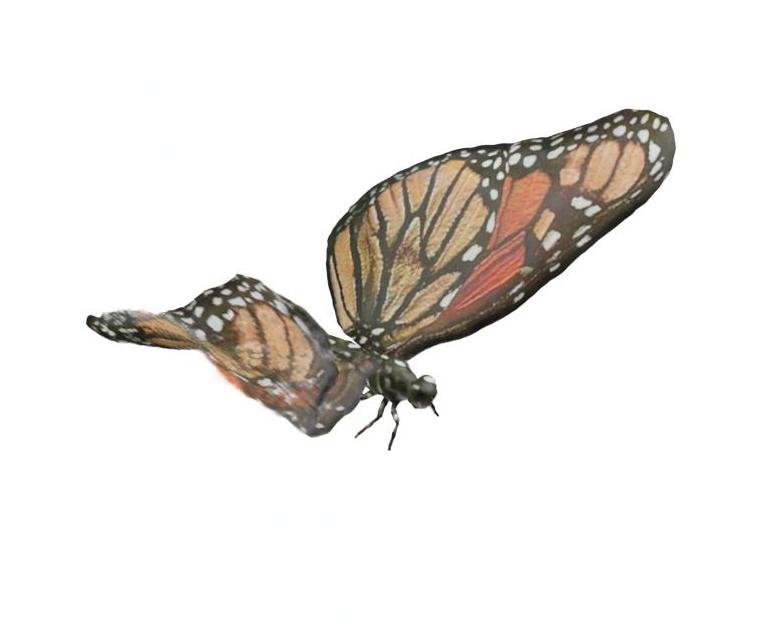}
        & \includegraphics[width=\hsize,valign=m]{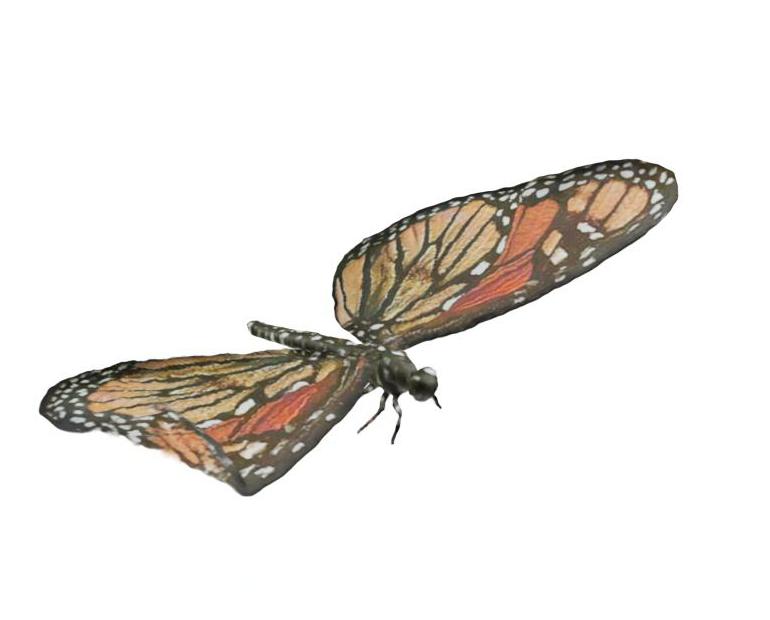}
        & \includegraphics[width=\hsize,valign=m]{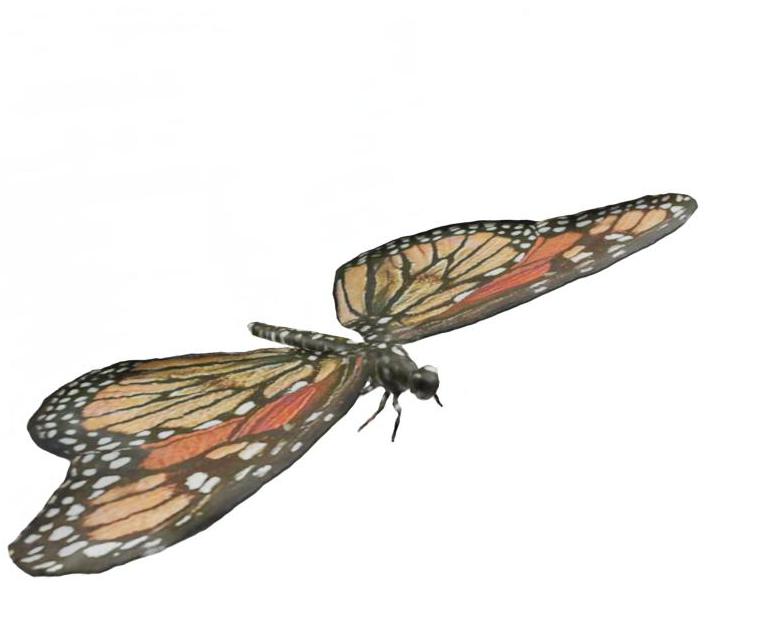}
         \\
         \midrule
\multirow{2}{*}[1ex]{\rotatebox[origin=c]{90}{$k=150$}}
        & \includegraphics[width=\hsize,valign=m]{figs/ab_but_pc_start.jpg}
        & \includegraphics[width=\hsize,valign=m]{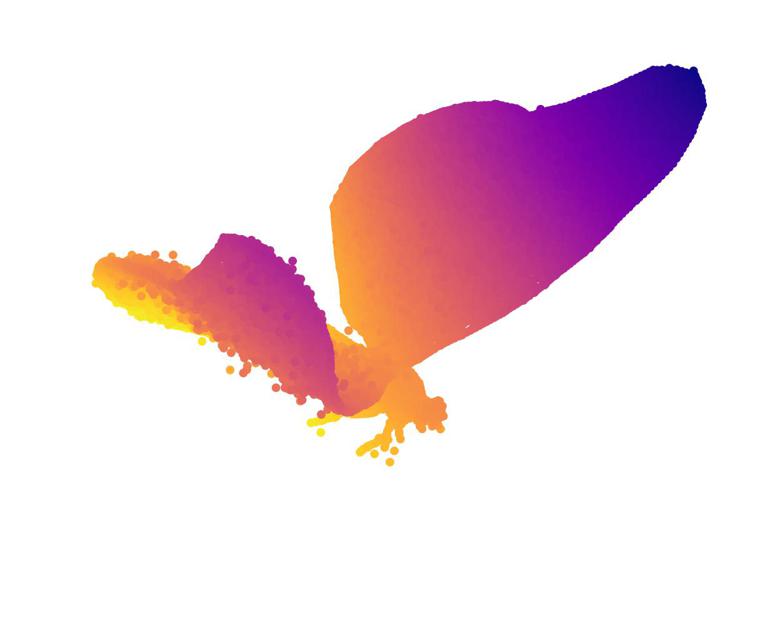}
        & \includegraphics[width=\hsize,valign=m]{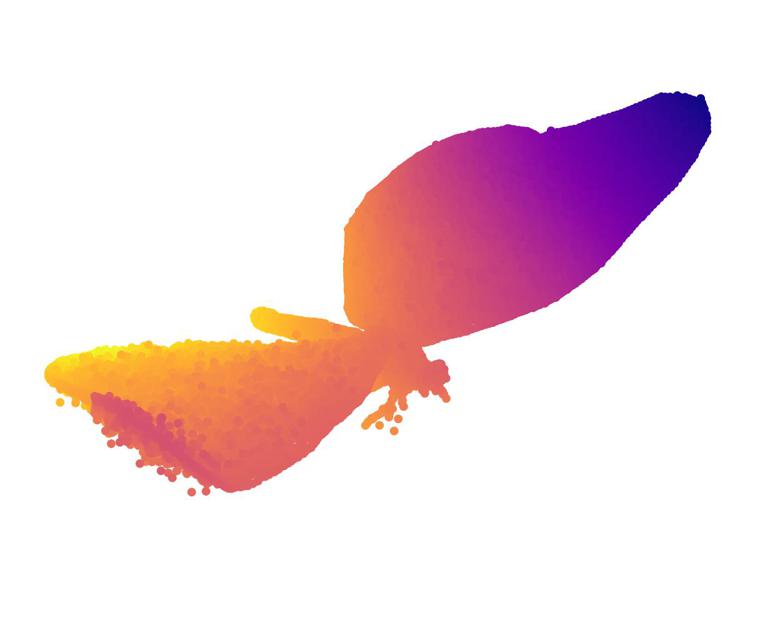}
        & \includegraphics[width=\hsize,valign=m]{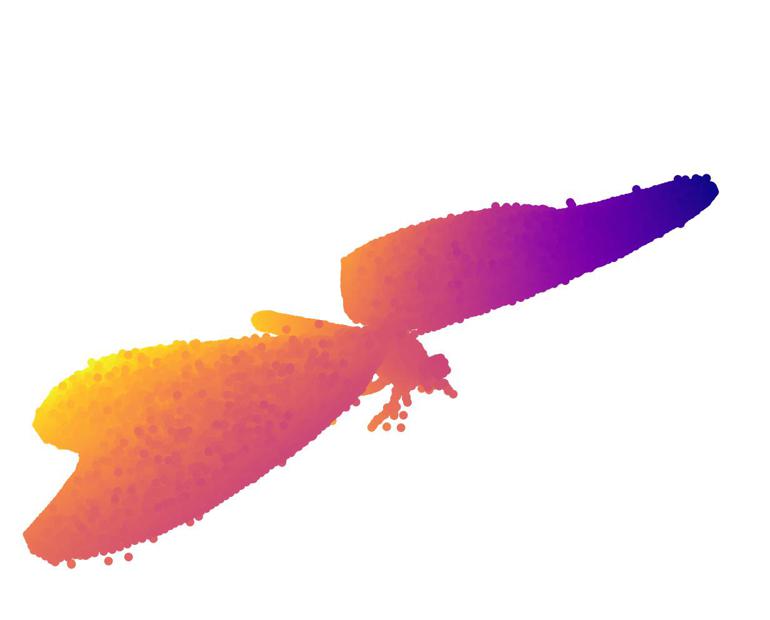}
        \\
        & \includegraphics[width=\hsize,valign=m]{figs/ab_but_rgb_start.jpg}
        & \includegraphics[width=\hsize,valign=m]{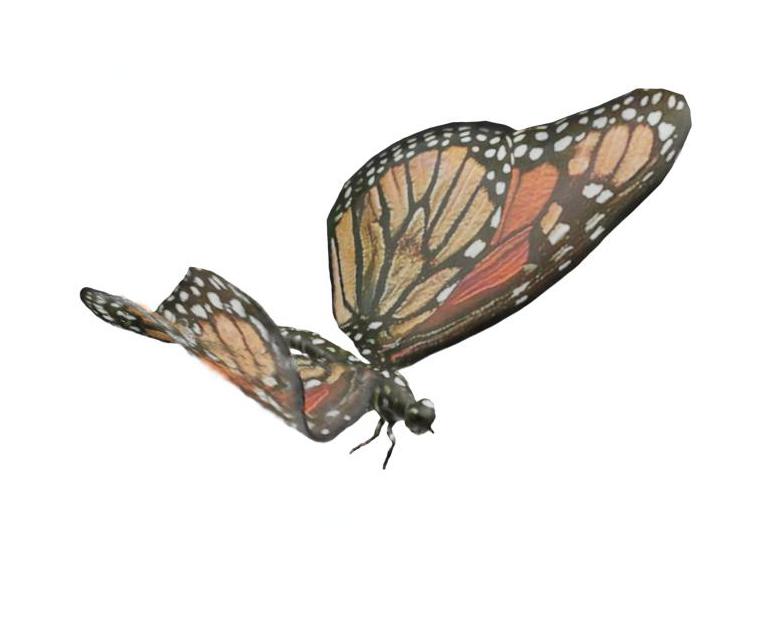}
        & \includegraphics[width=\hsize,valign=m]{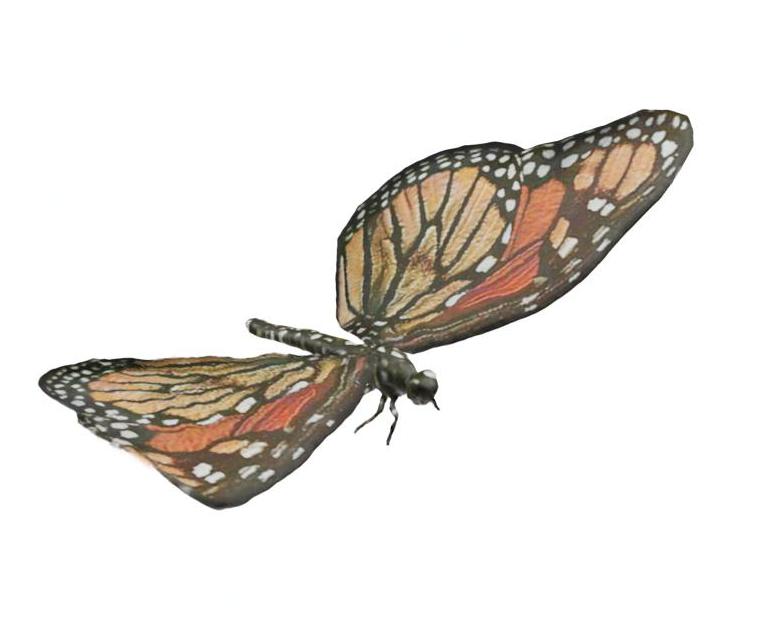}
        & \includegraphics[width=\hsize,valign=m]{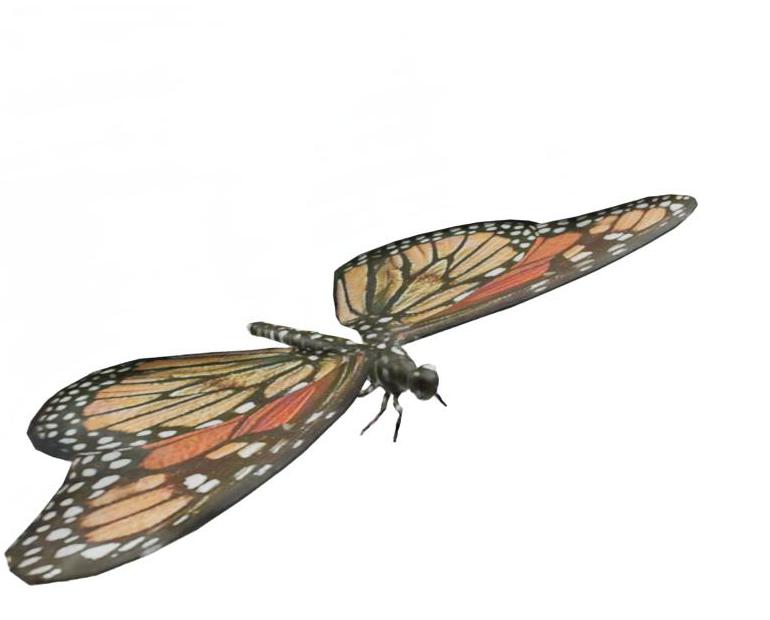}
        \\\midrule
\multirow{2}{*}[1ex]{\rotatebox[origin=c]{90}{$k=300$}}
        & \includegraphics[width=\hsize,valign=m]{figs/ab_but_pc_start.jpg}
        & \includegraphics[width=\hsize,valign=m]{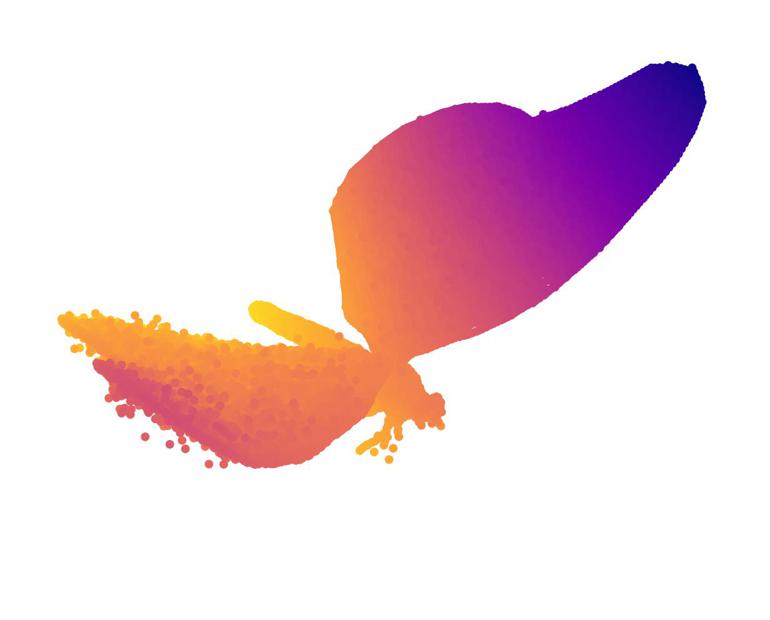}
        & \includegraphics[width=\hsize,valign=m]{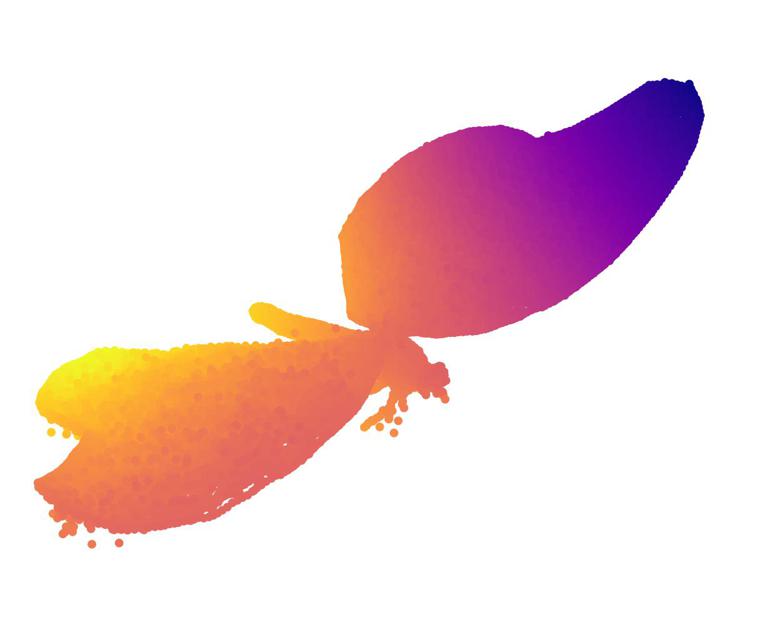}
        & \includegraphics[width=\hsize,valign=m]{figs/but-pc-end.jpg}
        \\
        & \includegraphics[width=\hsize,valign=m]{figs/ab_but_rgb_start.jpg}
        & \includegraphics[width=\hsize,valign=m]{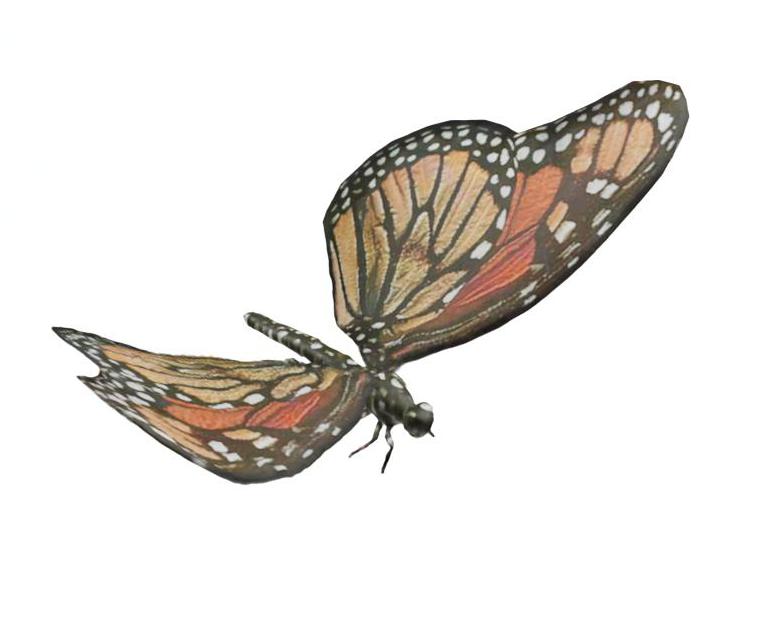}
        & \includegraphics[width=\hsize,valign=m]{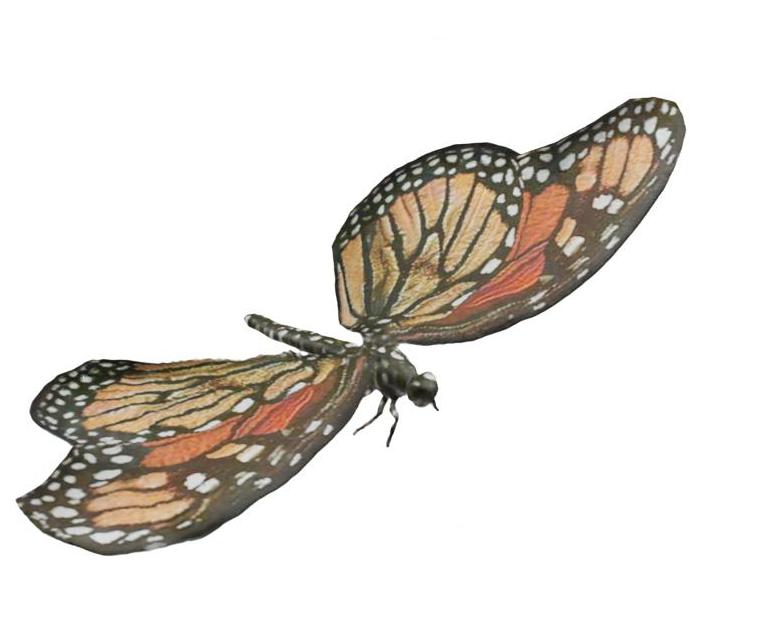}
        & \includegraphics[width=\hsize,valign=m]{figs/but-rgb-end.jpeg}
        
\end{tabularx}
\caption{
    Sensitivity analysis on the effect of $k$, the number of nearest neighbour used in regularization calculations. The results show that larger values of $k$ tend to increase the rigidity of moving parts while smaller values of $k$ result in more flexible part movements. In this example, $k=300$ exhibits the best surface continuity and smoothness throughout the interpolation process.
}
\label{fig:sensitivity}
\end{figure}

We perform a sensitivity analysis to evaluate our model's performance in relation to its hyperparameters. Specifically, we focused on examining the impact of varying the number of nearest neighbours, denoted as $k$, used in the regularization techniques. Figure~\ref{fig:sensitivity} illustrates the findings from our analysis of $k$'s influence. As shown, higher values of $k$ generally lead to increased rigidity in the moving parts of objects. This heightened rigidity contributes to more accurate interpolations, as observed in the improved preservation of surface smoothness and structural integrity, particularly noticeable in the butterfly's wing during its motion.

However, it is important to note that as $k$ increases, it imposes more constraints on object movements due to the larger neighbourhood size considered. This can be a limiting factor, particularly in scenes with intricate geometries, where an excessively high $k$ value might overly restrict movement and hinder the parts from interpolating correctly. In practice, we choose a value of $k$ that strikes a balance between maintaining structural rigidity and allowing sufficient flexibility for part movement.

\begin{figure}[ht]
\footnotesize
\begin{tabularx}{\linewidth}{lYYYY}
  & Epoch 0 & Epoch 2 & Epoch 4 & Epoch 8  \\
\multirow{2}{*}[1ex]{\rotatebox[origin=c]{90}{$m=10$}}
        & \includegraphics[width=\hsize,valign=m]{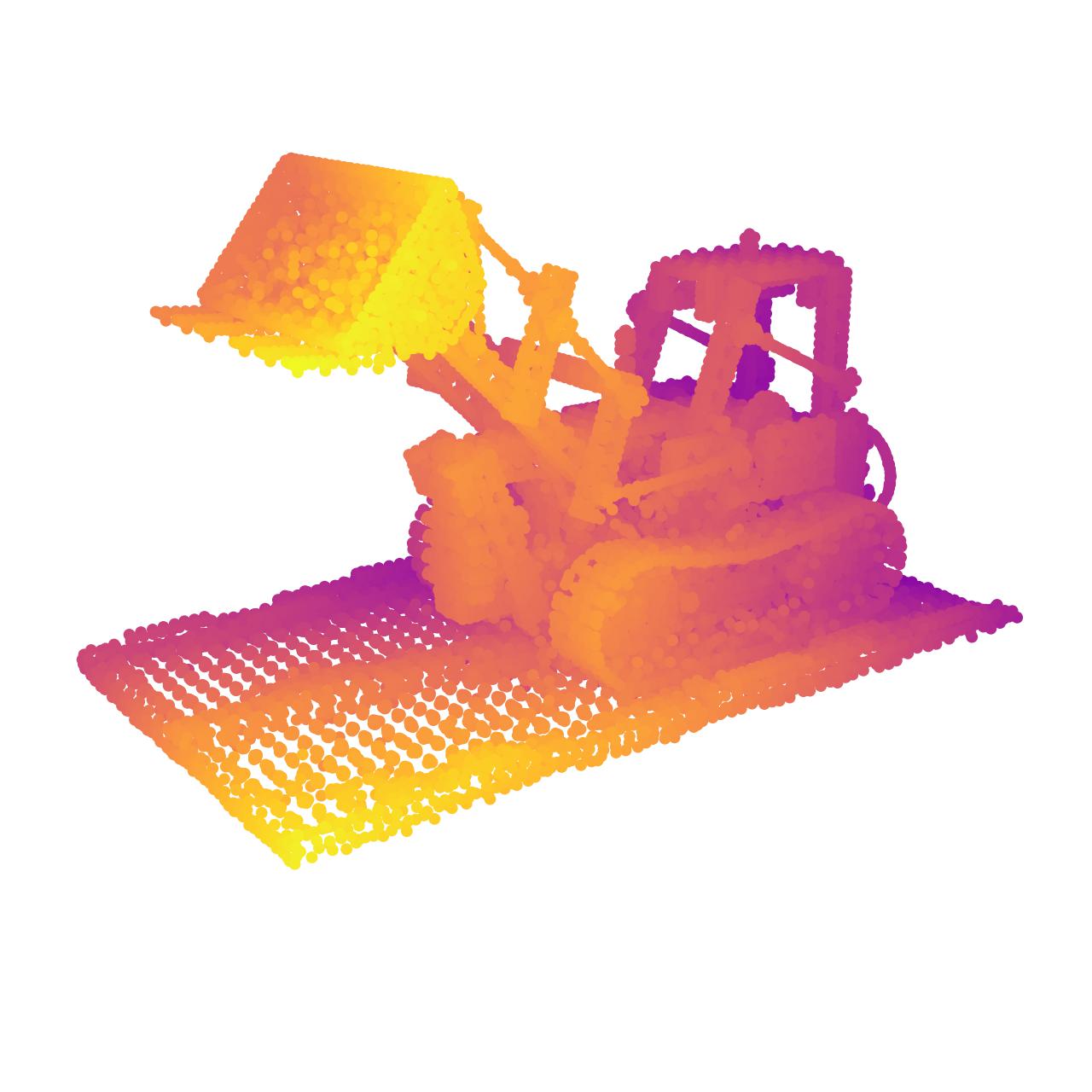}
        & \includegraphics[width=\hsize,valign=m]{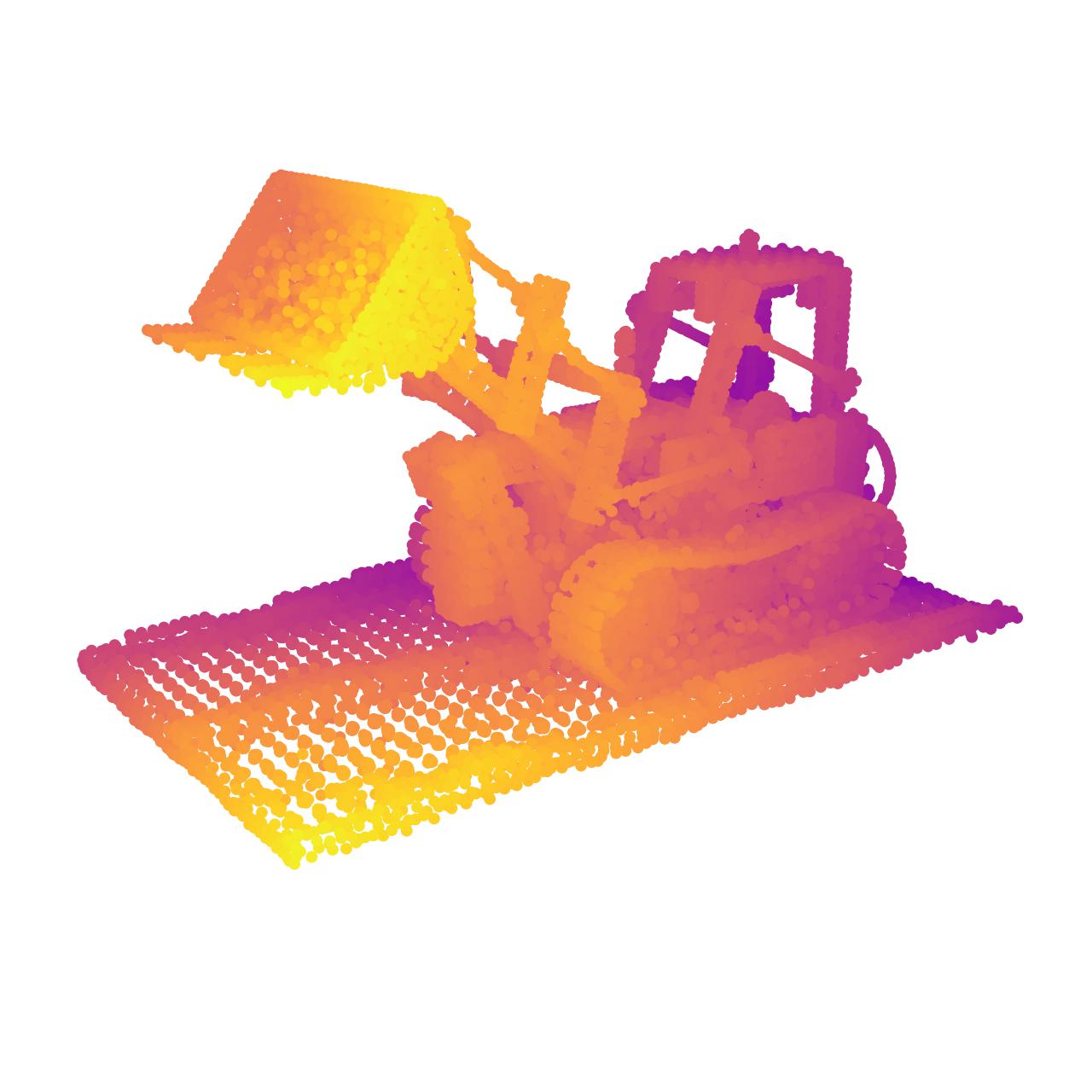}
        & \includegraphics[width=\hsize,valign=m]{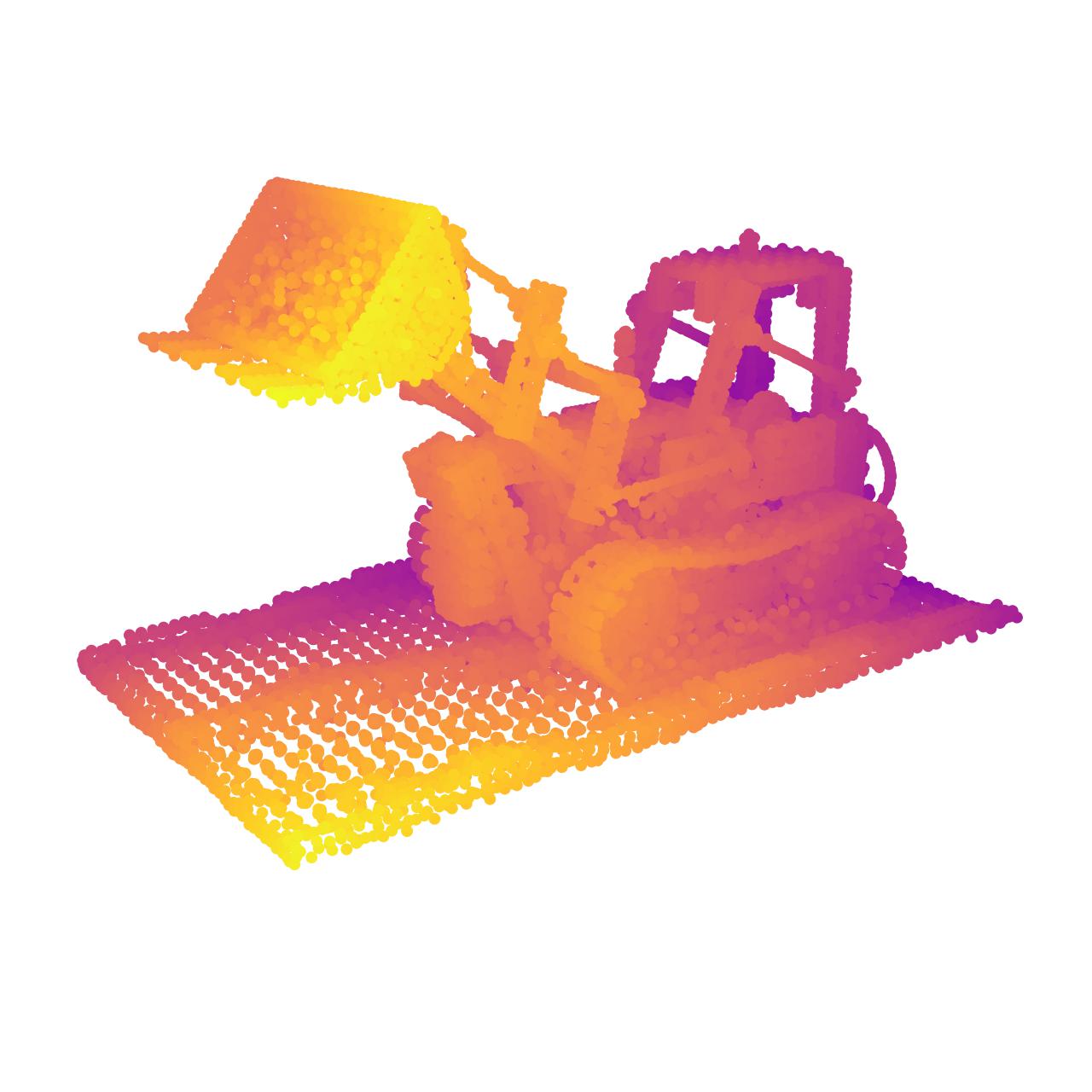}
        & \includegraphics[width=\hsize,valign=m]{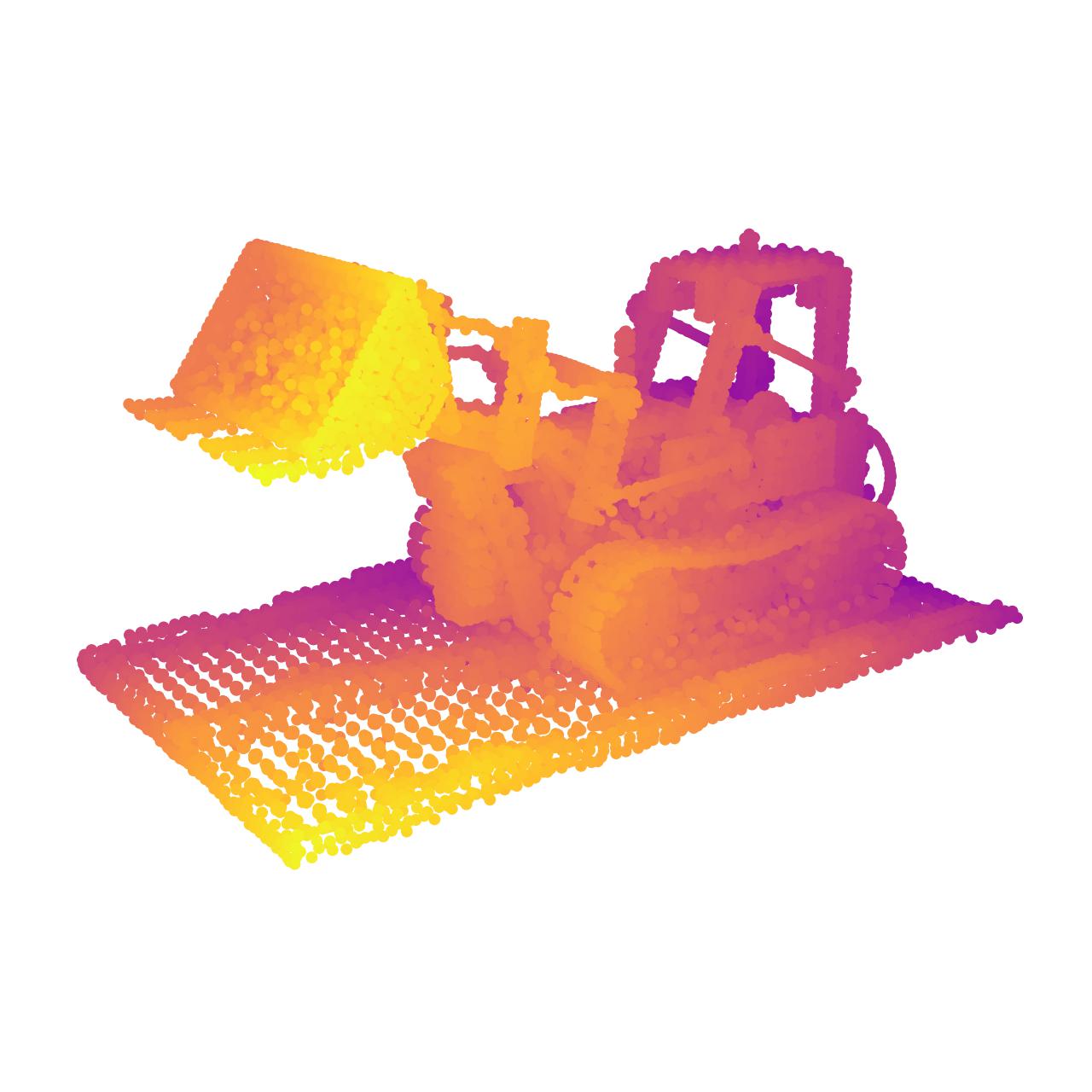}
         \\
        & \includegraphics[width=\hsize,valign=m]{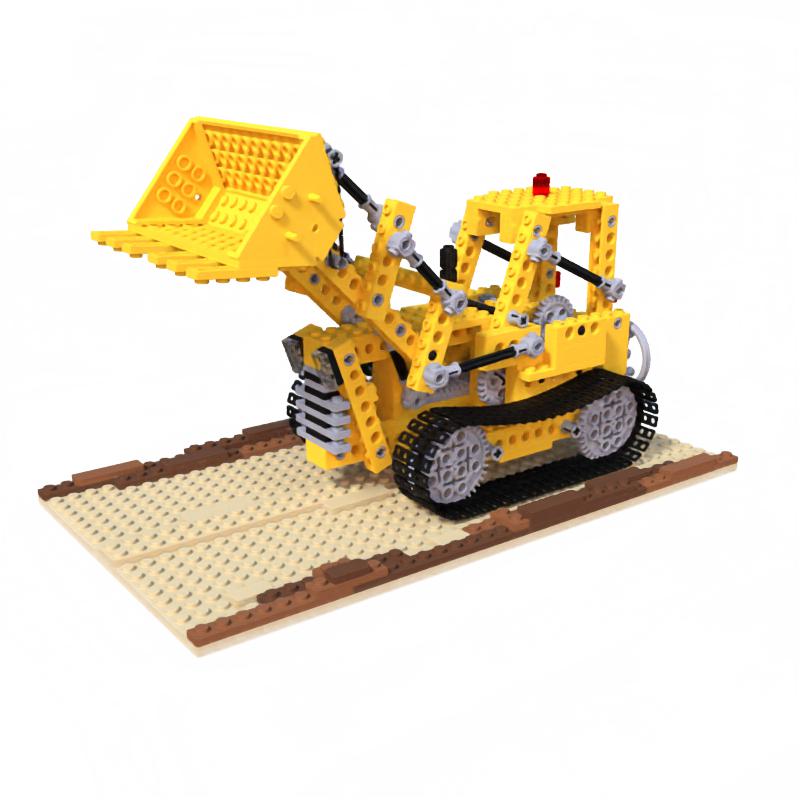}
        & \includegraphics[width=\hsize,valign=m]{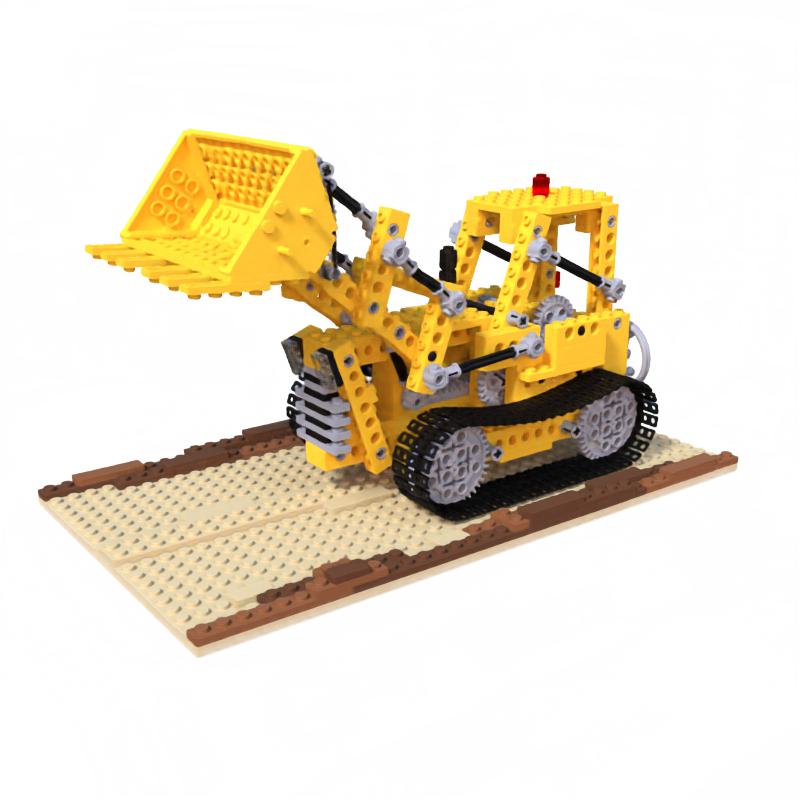}
        & \includegraphics[width=\hsize,valign=m]{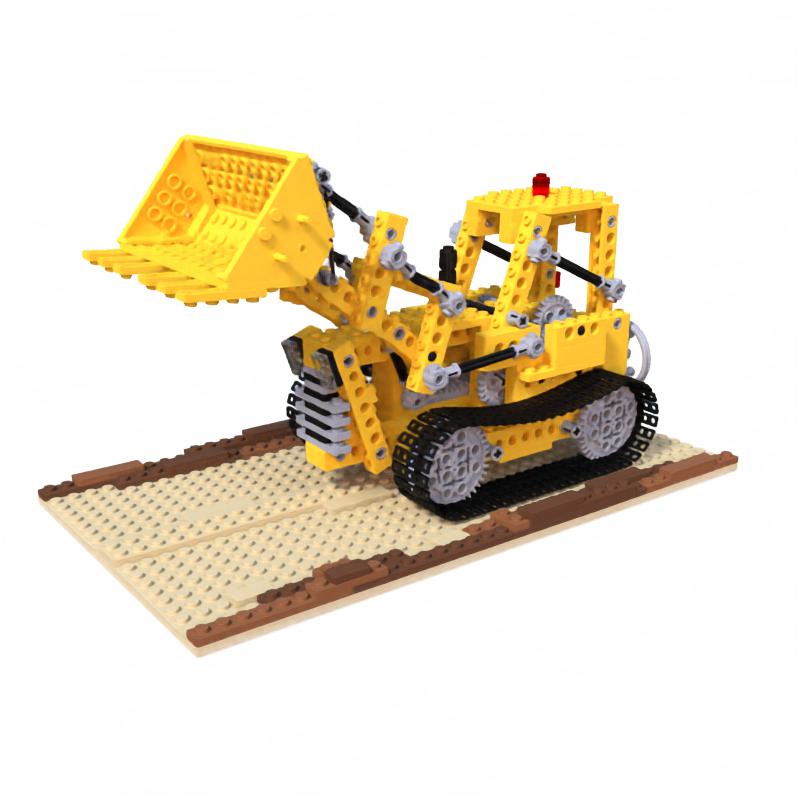}
        & \includegraphics[width=\hsize,valign=m]{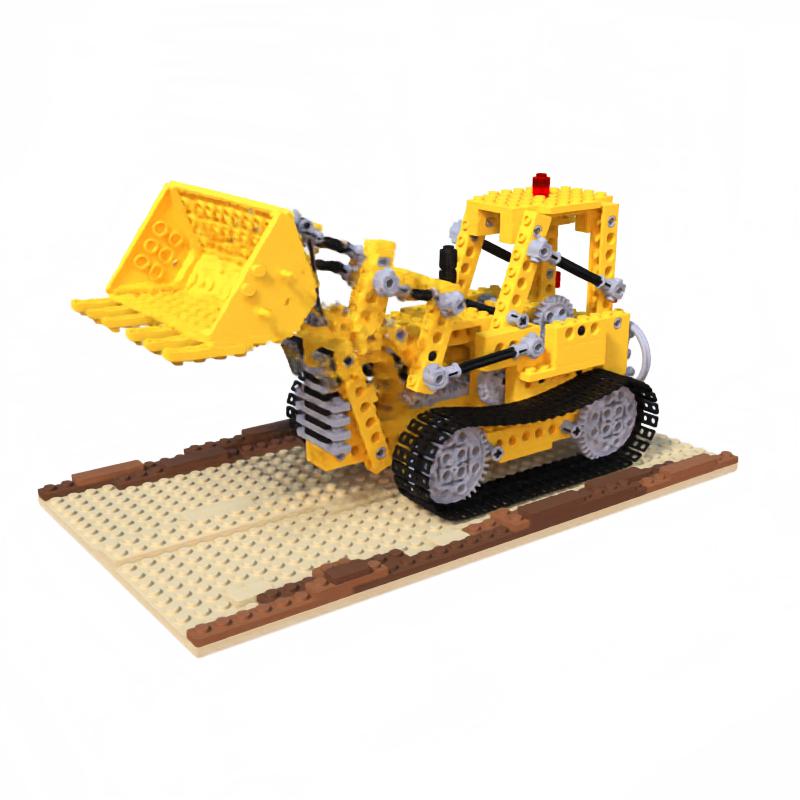}
         \\
         \midrule
\multirow{2}{*}[1ex]{\rotatebox[origin=c]{90}{$m=50$}}
        & \includegraphics[width=\hsize,valign=m]{figs/lego-pc-start.jpg}
        & \includegraphics[width=\hsize,valign=m]{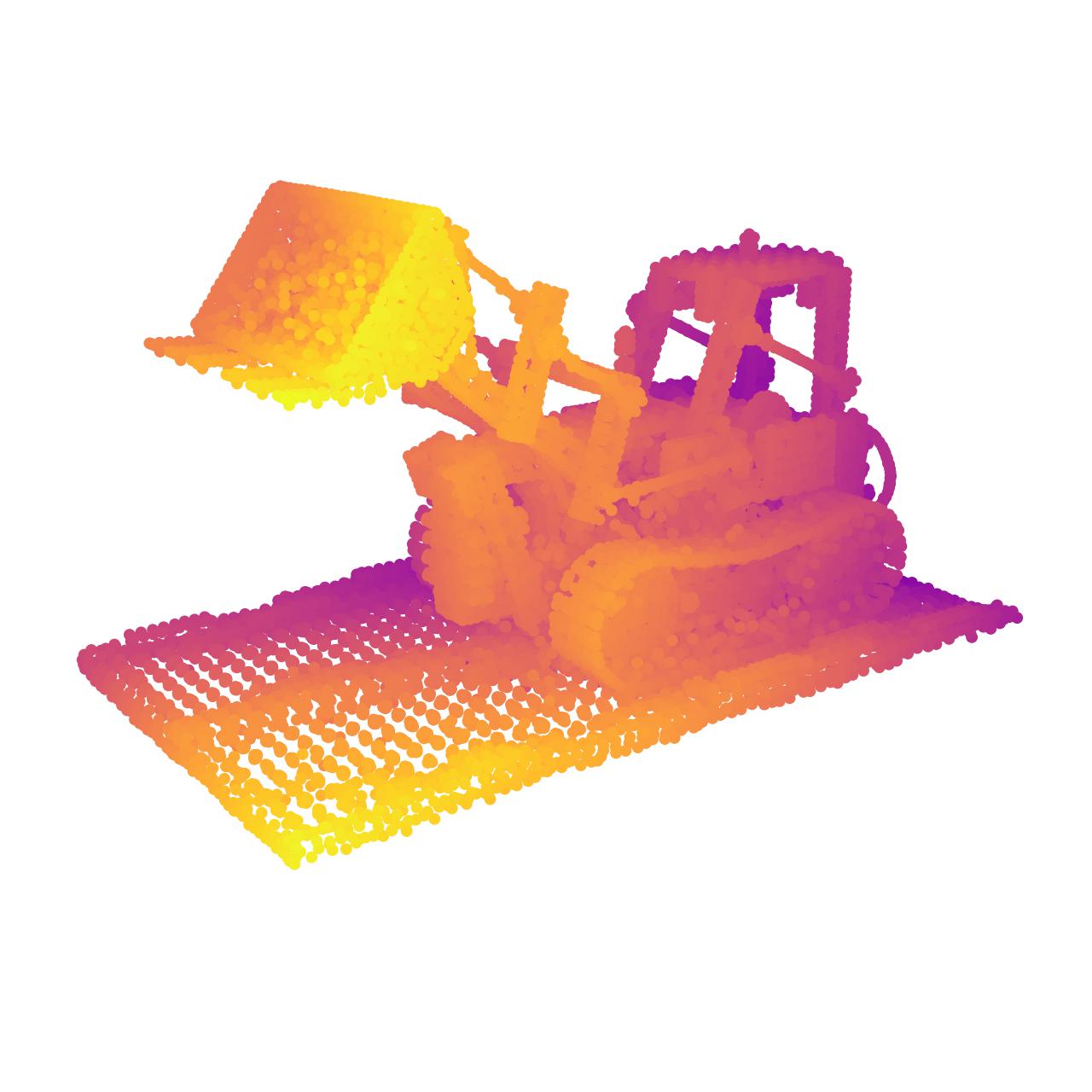}
        & \includegraphics[width=\hsize,valign=m]{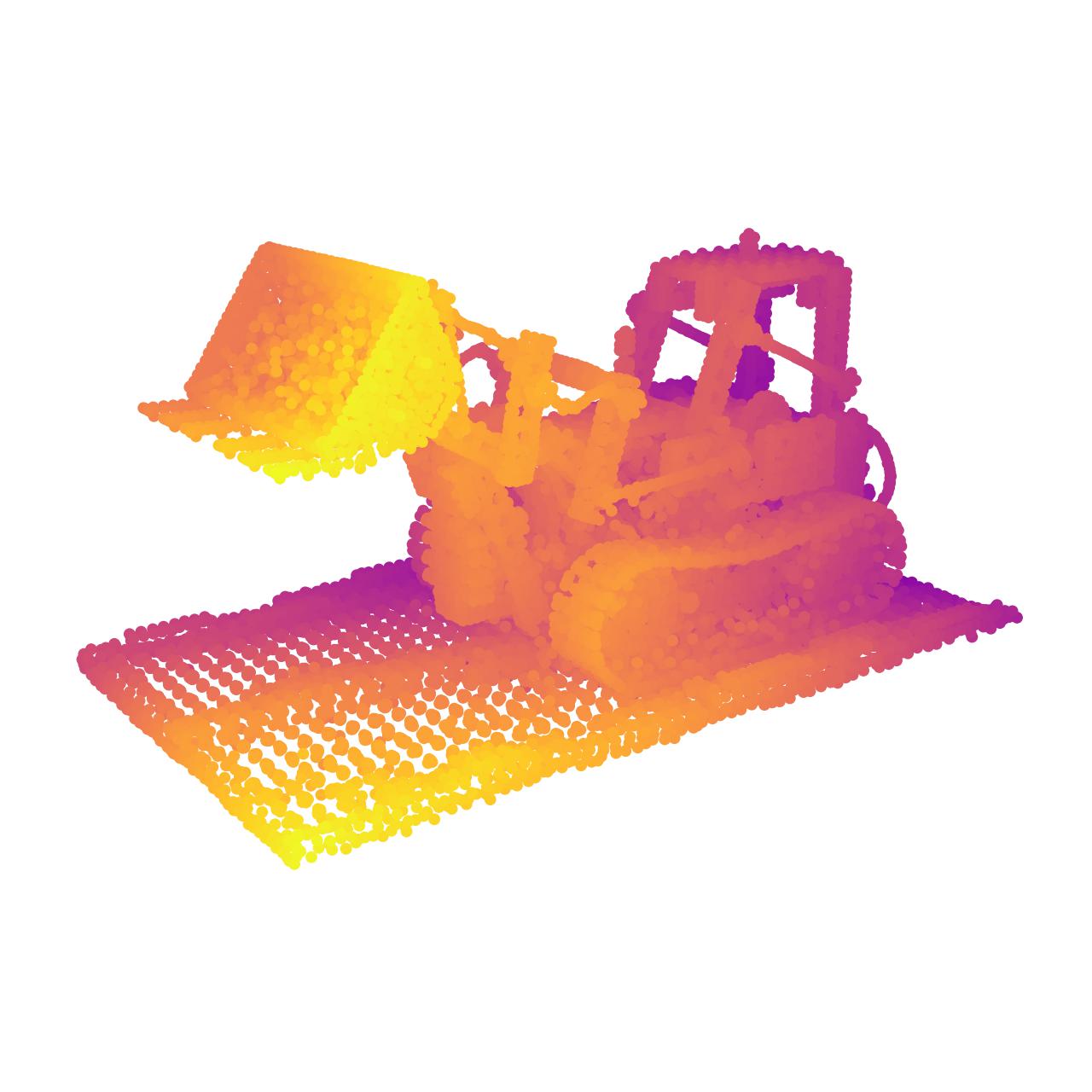}
        & \includegraphics[width=\hsize,valign=m]{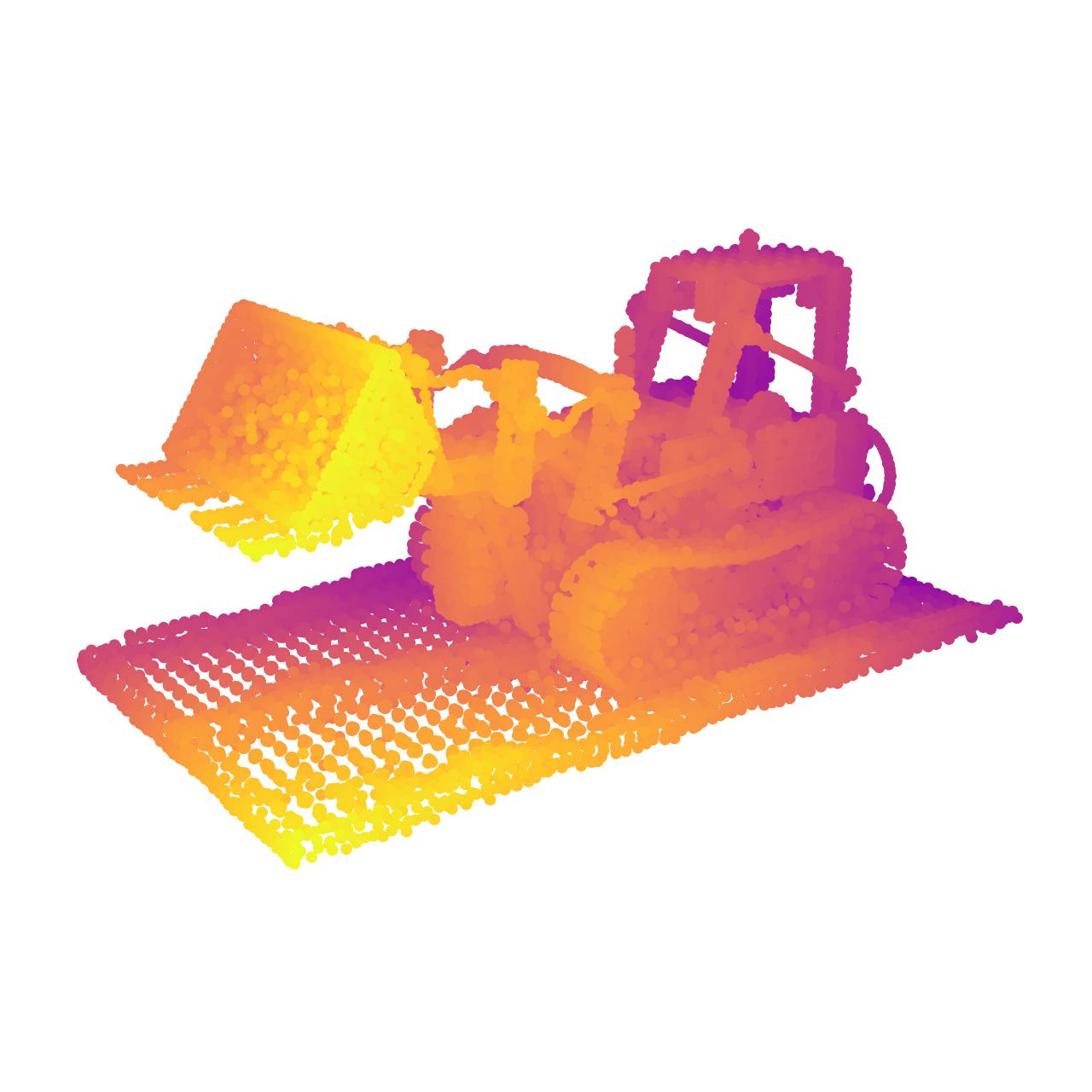}
        \\
        & \includegraphics[width=\hsize,valign=m]{figs/lego-rgb-start.jpg}
        & \includegraphics[width=\hsize,valign=m]{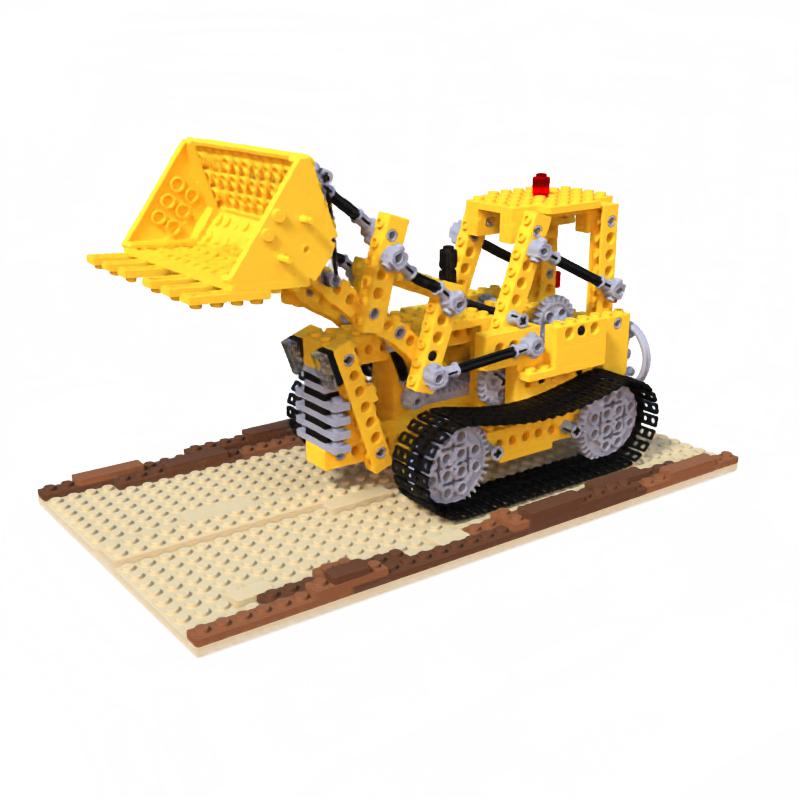}
        & \includegraphics[width=\hsize,valign=m]{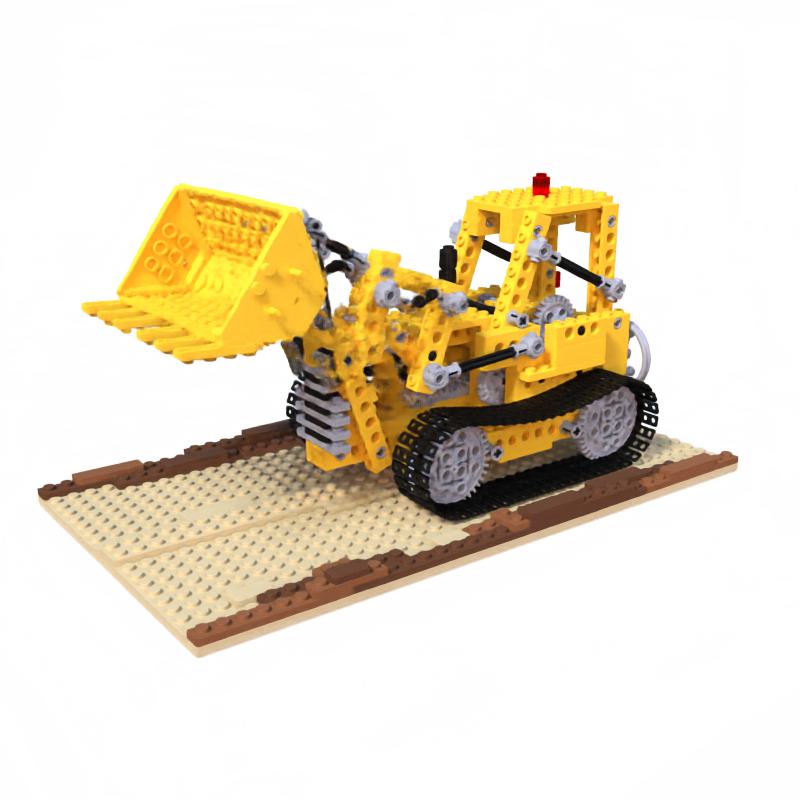}
        & \includegraphics[width=\hsize,valign=m]{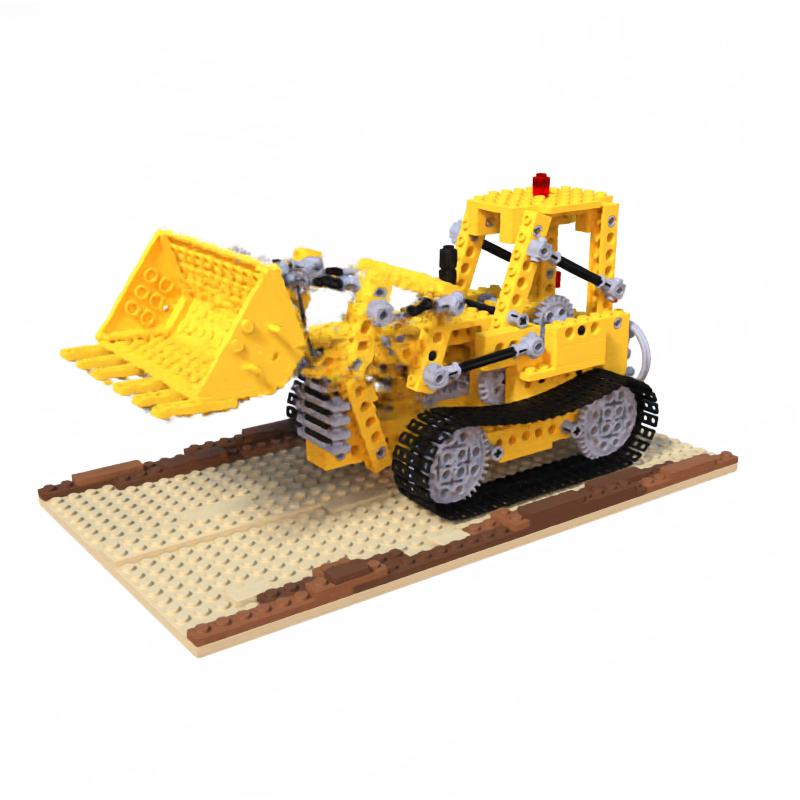}
        \\
         \midrule
\multirow{2}{*}[1ex]{\rotatebox[origin=c]{90}{$m=100$}}
        & \includegraphics[width=\hsize,valign=m]{figs/lego-pc-start.jpg}
        & \includegraphics[width=\hsize,valign=m]{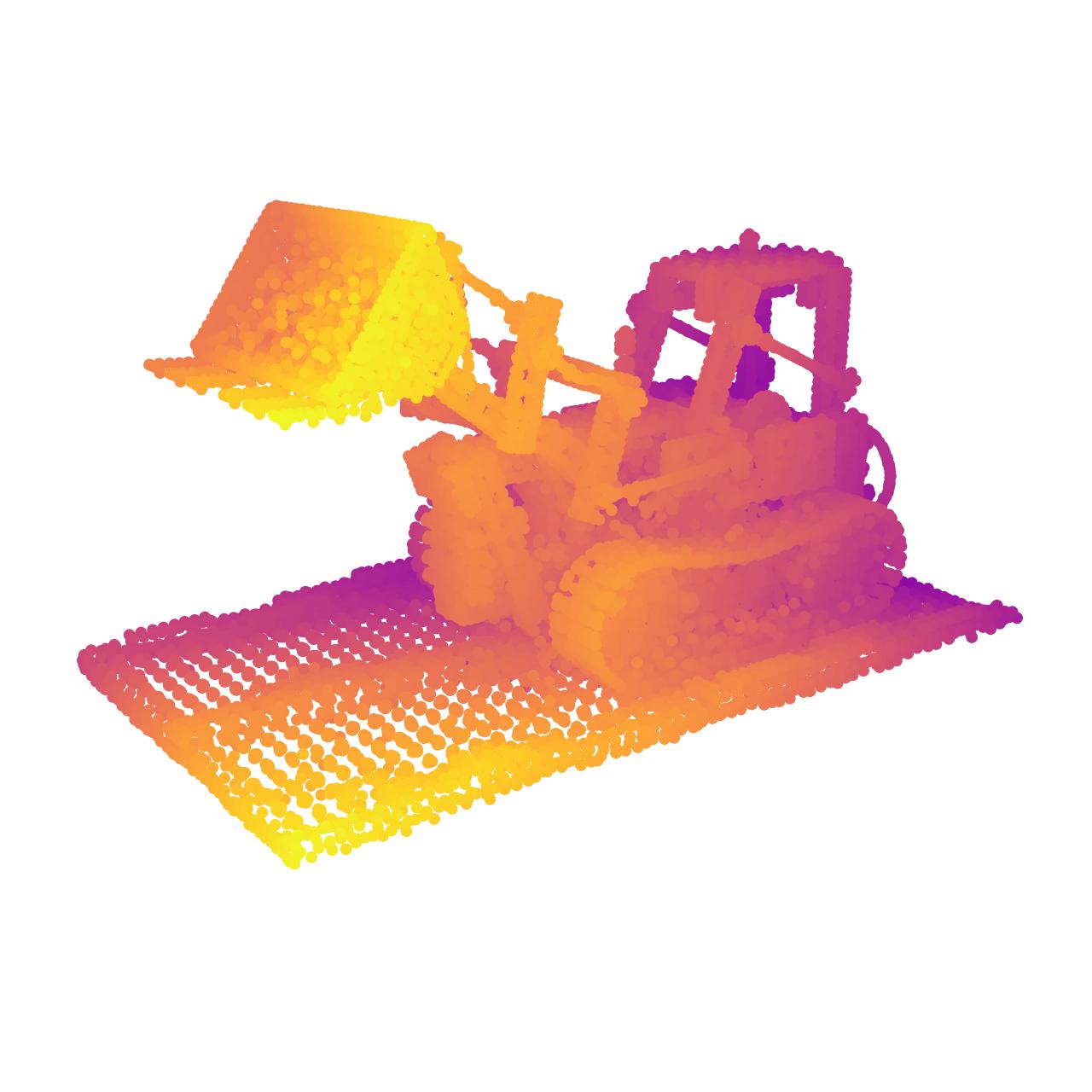}
        & \includegraphics[width=\hsize,valign=m]{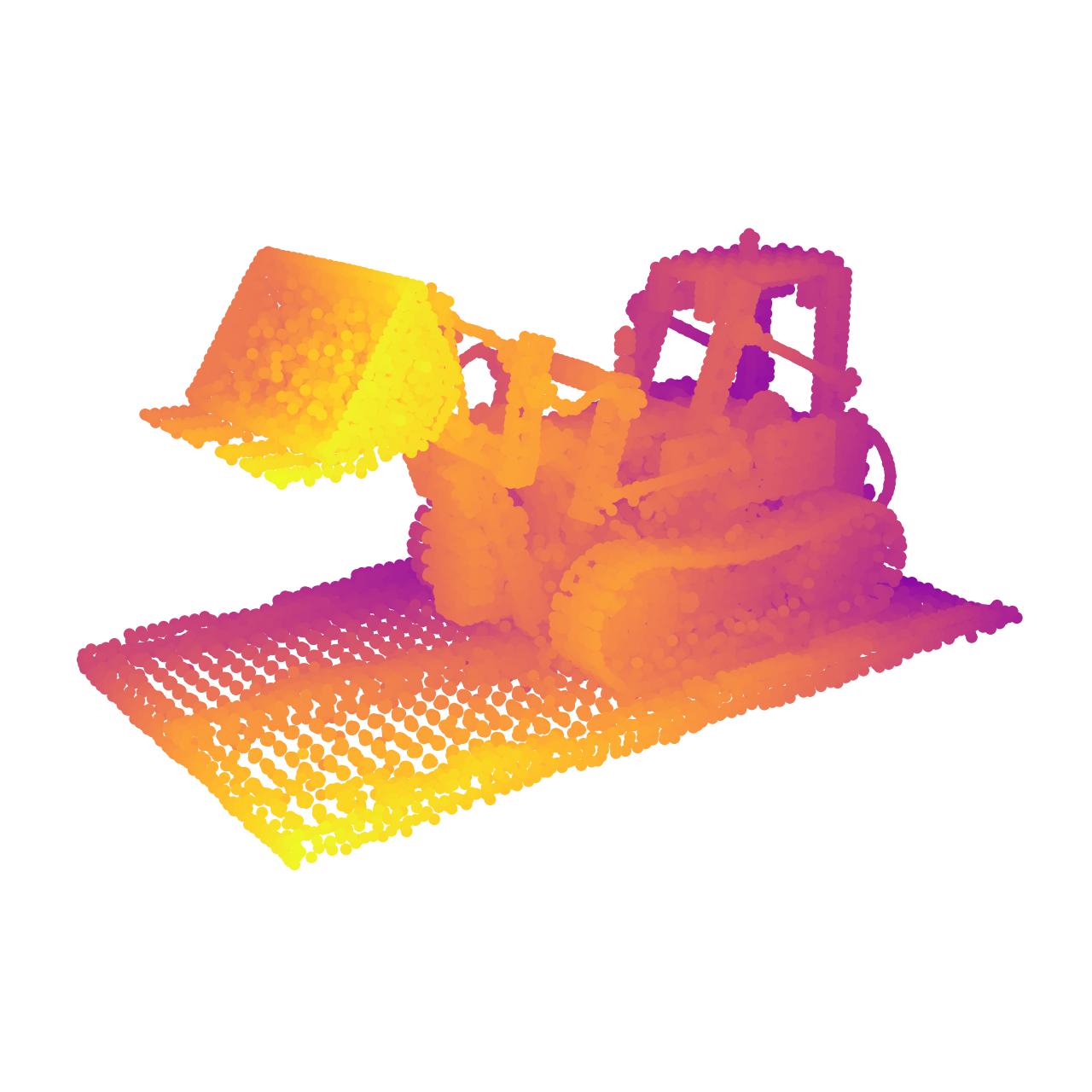}
        & \includegraphics[width=\hsize,valign=m]{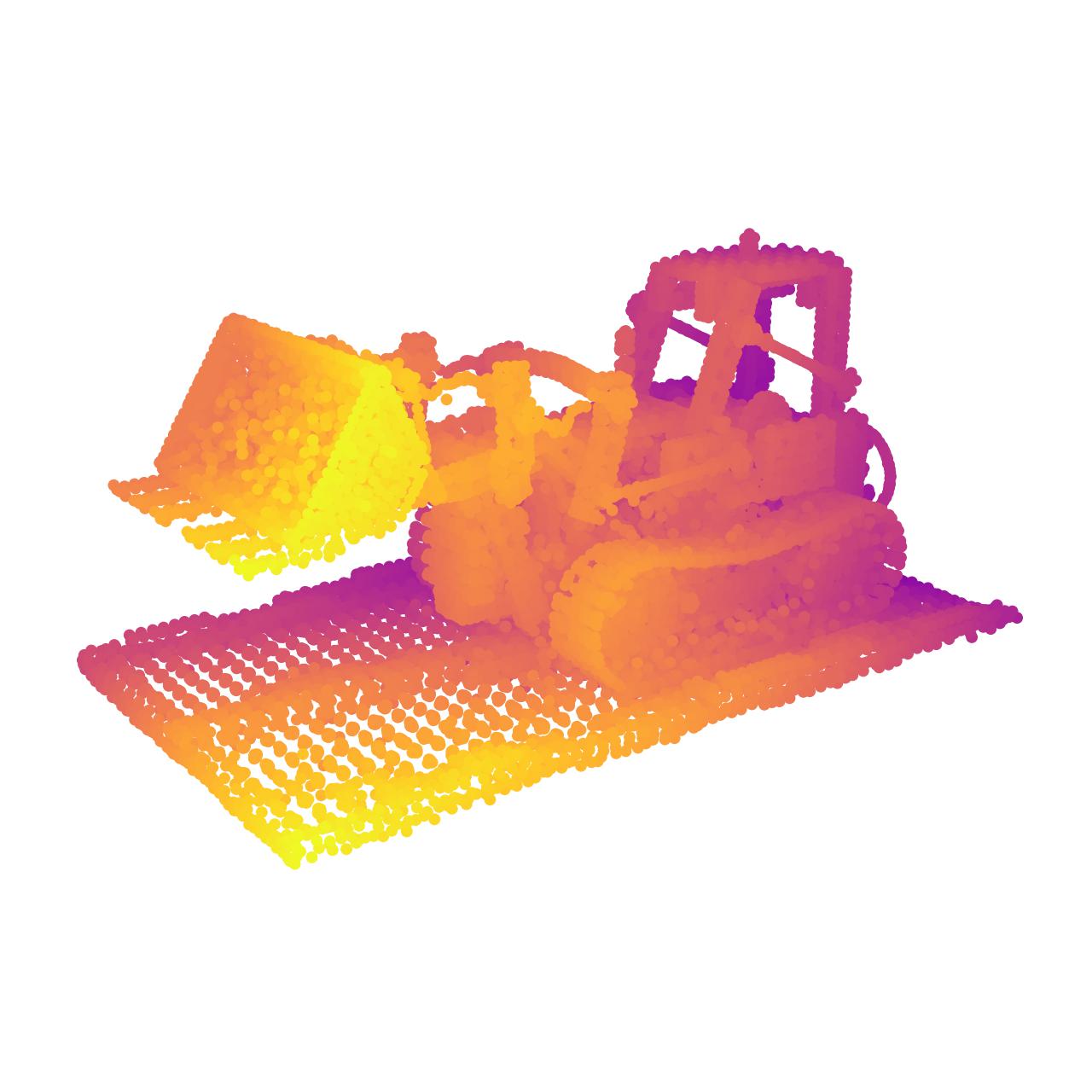}
        \\
        & \includegraphics[width=\hsize,valign=m]{figs/lego-rgb-start.jpg}
        & \includegraphics[width=\hsize,valign=m]{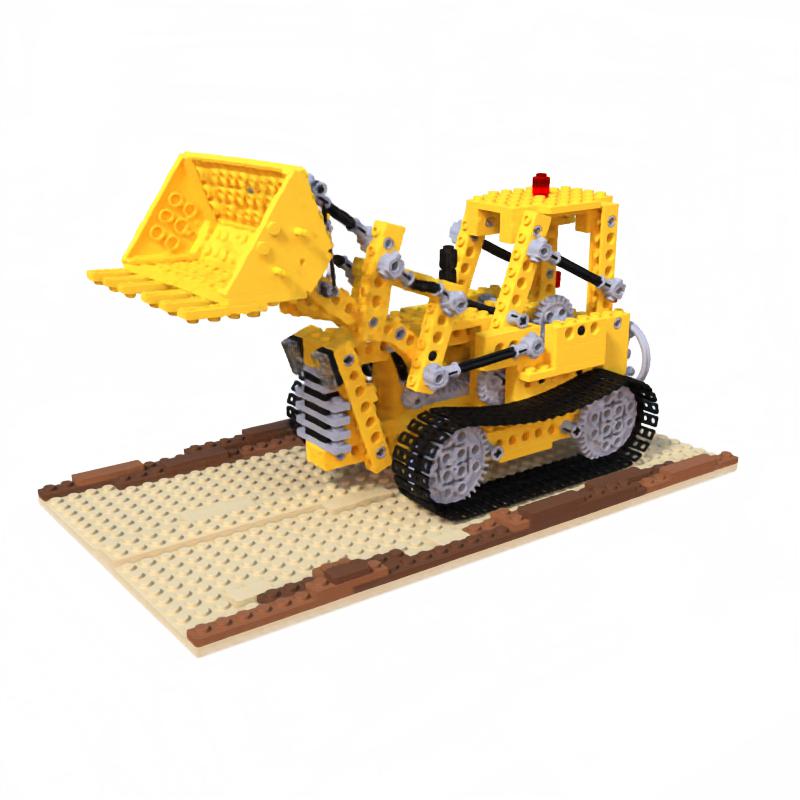}
        & \includegraphics[width=\hsize,valign=m]{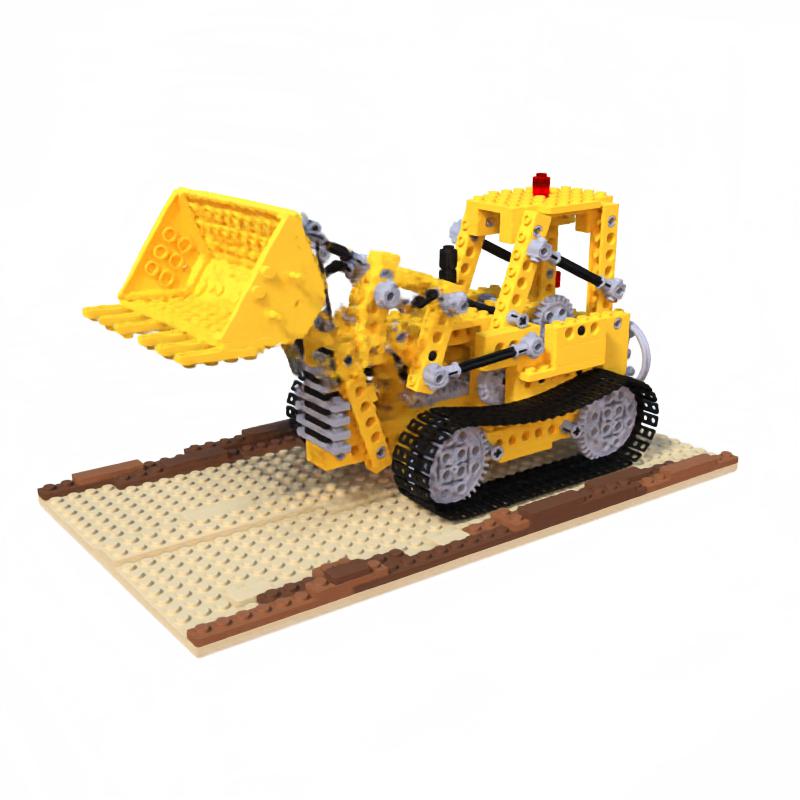}
        & \includegraphics[width=\hsize,valign=m]{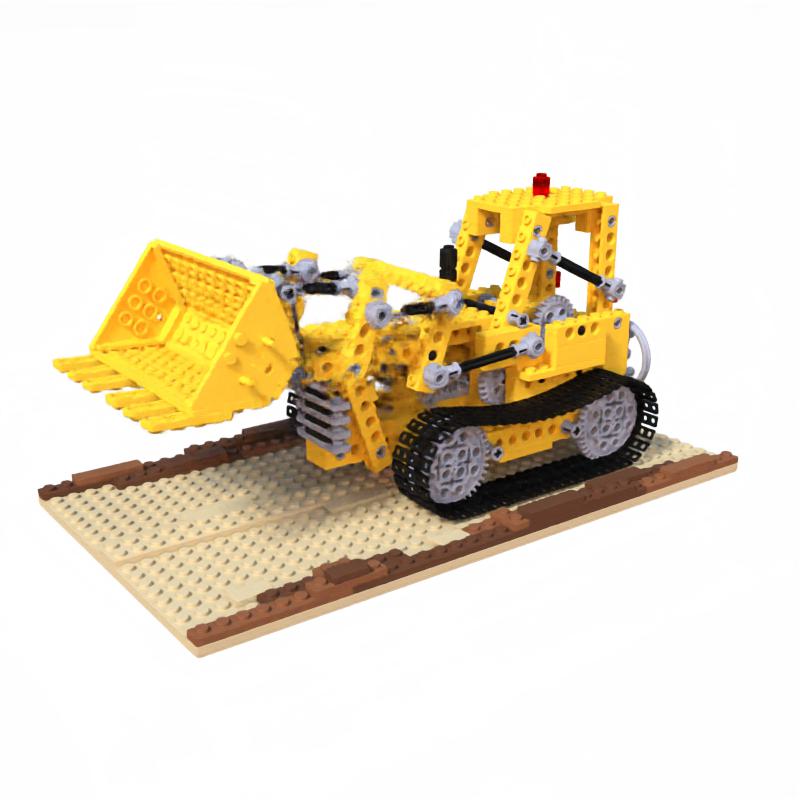}
        \\\midrule
\end{tabularx}
\caption{
    Sensitivity analysis on the effect of $m$, the interval size in terms of number of iterations to apply local displacement averaging step (LDAS).
}
\label{fig:sensitivity-supp}
\end{figure}

We also show the effect of varying the interval $m$ for the local displacement averaging step (LDAS). As shown in Figure~\ref{fig:sensitivity-supp}, too small a value of $m$ may slow down the geometry adaptation process. In practice, we choose a value of $m=100$ that best balances the adaptation speed and the quality of the intermediate renderings.

\section{Implementation Details}
\paragraph{Dataset Details}

We use multi-view RGB images from both the start and end states of each scene as our input data. The training set for each scene consists of 100 randomly selected views from the upper hemisphere for each state, while the evaluation set comprises 200 unseen test views. In synthetic scenes, all images are rendered at a resolution of $800\times 800$ pixels. The real-world tablet stand scene is rendered at a resolution of $1008\times 756$ pixels, and the lamp scene at $960\times 540$ pixels.

\paragraph{Training Details}
The duration of the intermediate scene interpolation process in our method takes 16 epochs. To enhance efficiency, we found that finetuning the model on target images downsampled by a factor of two is sufficient. This results in a training time of approximately an hour on a single NVIDIA A100 GPU. The end state appearance finetuning takes 16 epochs. We choose the number of nearest neighbours $k$ for each scene based on how rigid the object should be -- the more rigid it is, the higher the value of $k$. The specific values of $k$ for each scene are detailed in Table~\ref{tab:param}.

\begin{table*}[h]
    \centering
    \begin{tabular}{cccccc|cc}
    \toprule
    \multicolumn{6}{c}{Synthetic Scenes} & \multicolumn{2}{c}{Real-world Scenes}\\
    \midrule
         Butterfly & Crab & Dolphin & Giraffe & Lego Bulldozer & Lego Man & Stand & Lamp\\
         \midrule
         300 & 150 & 200 & 70 & 100 & 200 & 200 & 200\\
         \bottomrule
    \end{tabular}
    \caption{Different values of nearest neighbour $k$ for each scene.}
    \label{tab:param}
\end{table*}

For the baseline method, Dynamic Gaussian~\cite{Luiten2023Dynamic3G}, their original approach involves finetuning for 75 epochs at each subsequent time step after the initial one. To better accommodate larger scene changes in our context, we extend this significantly to 300 epochs.

\section{Additional Results}
\begin{figure*}[t]
\footnotesize
\begin{tabularx}{\linewidth}{lYYYY|YYYY}
& \multicolumn{2}{c}{Dynamic Gaussian~\cite{Luiten2023Dynamic3G}} & \multicolumn{2}{c}{PAPR in Motion (Ours)} & \multicolumn{2}{c}{Dynamic Gaussian~\cite{Luiten2023Dynamic3G}} & \multicolumn{2}{c}{PAPR in Motion (Ours)} \\
& Point Cloud & Rendering & Point Cloud & Rendering & Point Cloud & Rendering & Point Cloud & Rendering \\
    \rotatebox[origin=c]{90}{Start} 
    & \includegraphics[width=\hsize,valign=m]{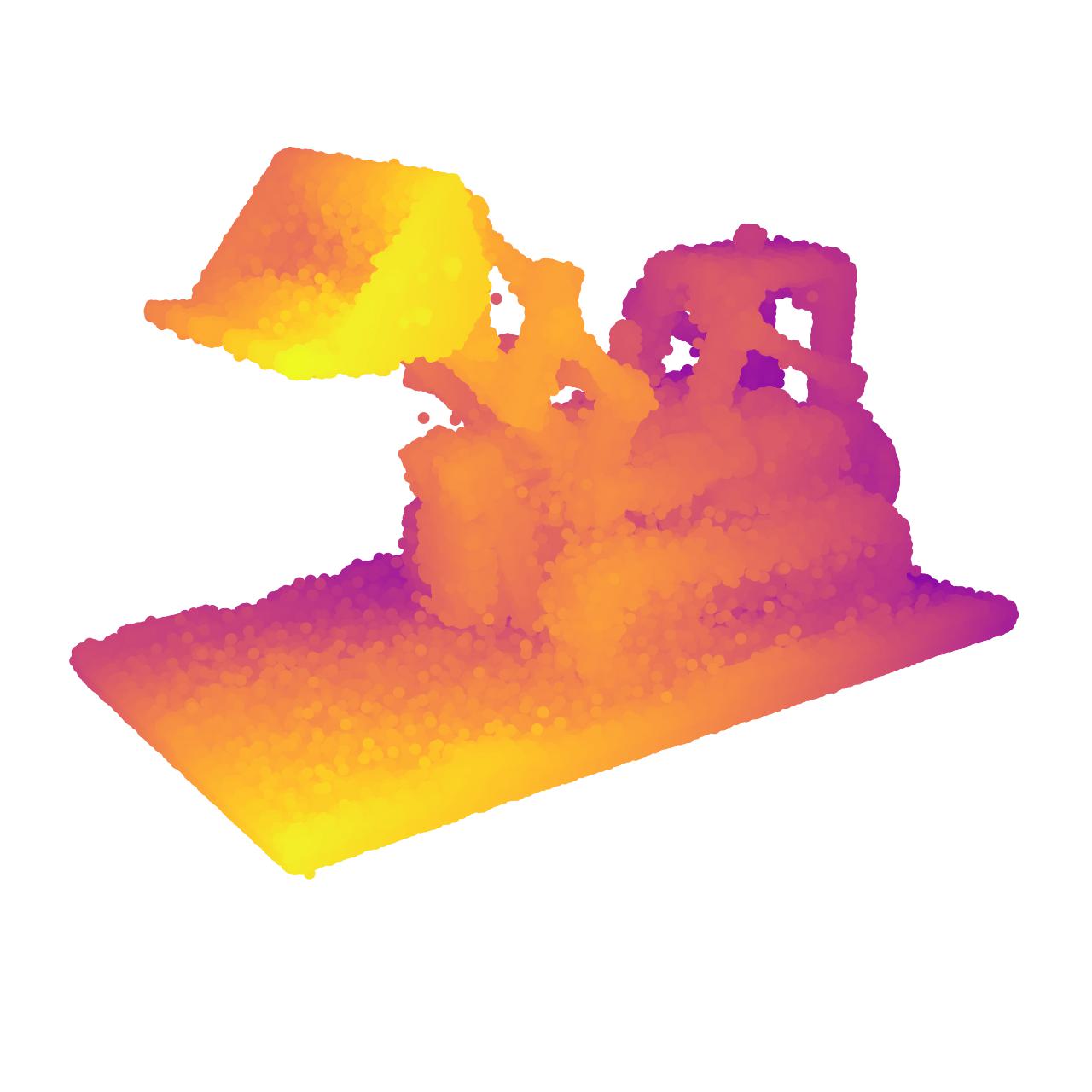}
    & \includegraphics[width=\hsize,valign=m]{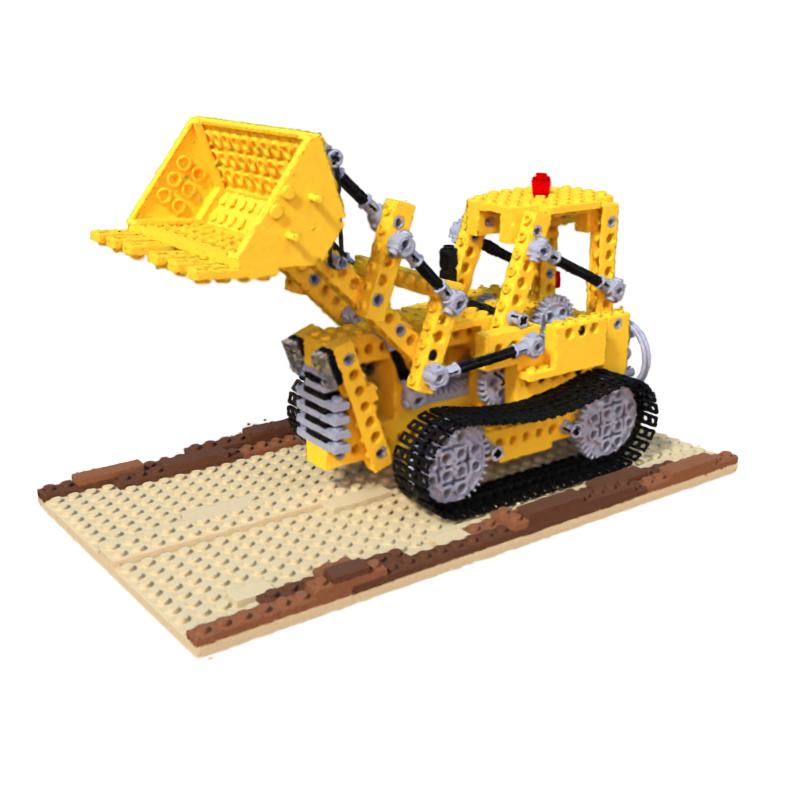}
    & \includegraphics[width=\hsize,valign=m]{figs/lego-pc-start.jpg}
    & \includegraphics[width=\hsize,valign=m]{figs/lego-rgb-start.jpg}
    & \includegraphics[width=\hsize,valign=m]{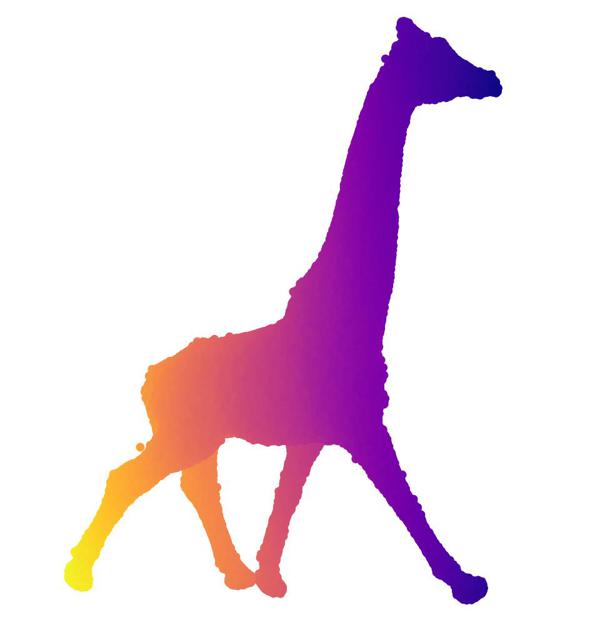}
    & \includegraphics[width=\hsize,valign=m]{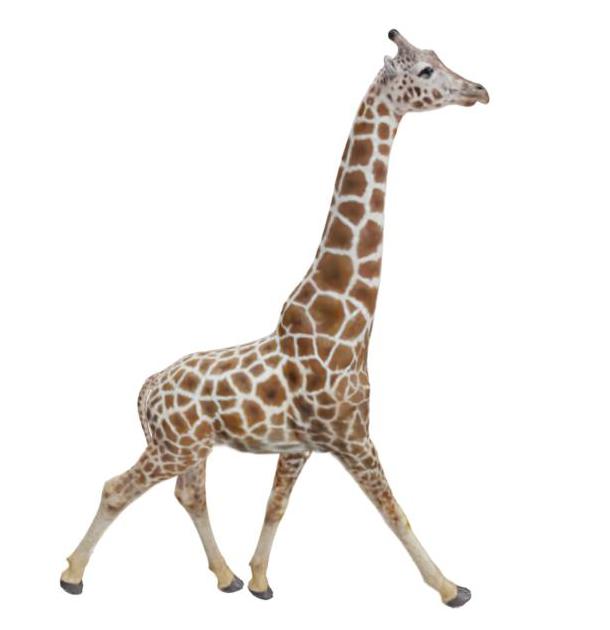}
    & \includegraphics[width=\hsize,valign=m]{figs/ab_giraffe_pc_start.jpg}
    & \includegraphics[width=\hsize,valign=m]{figs/ab_giraffe_rgb_start.jpg}
    \\
    \multirow{2}{*}{\rotatebox[origin=c]{90}{Intermediate}}
    & \includegraphics[width=\hsize,valign=m]{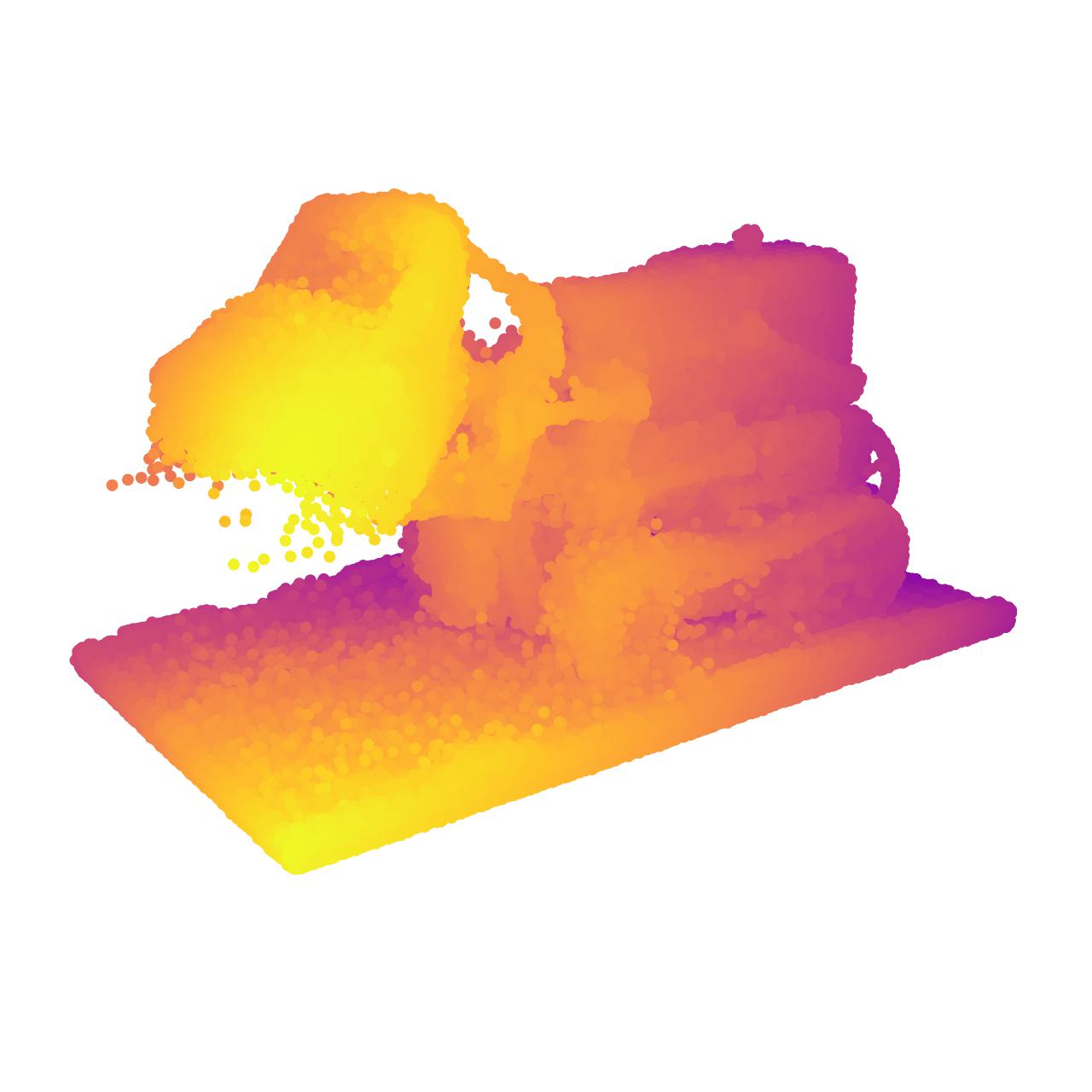}
    & \includegraphics[width=\hsize,valign=m]{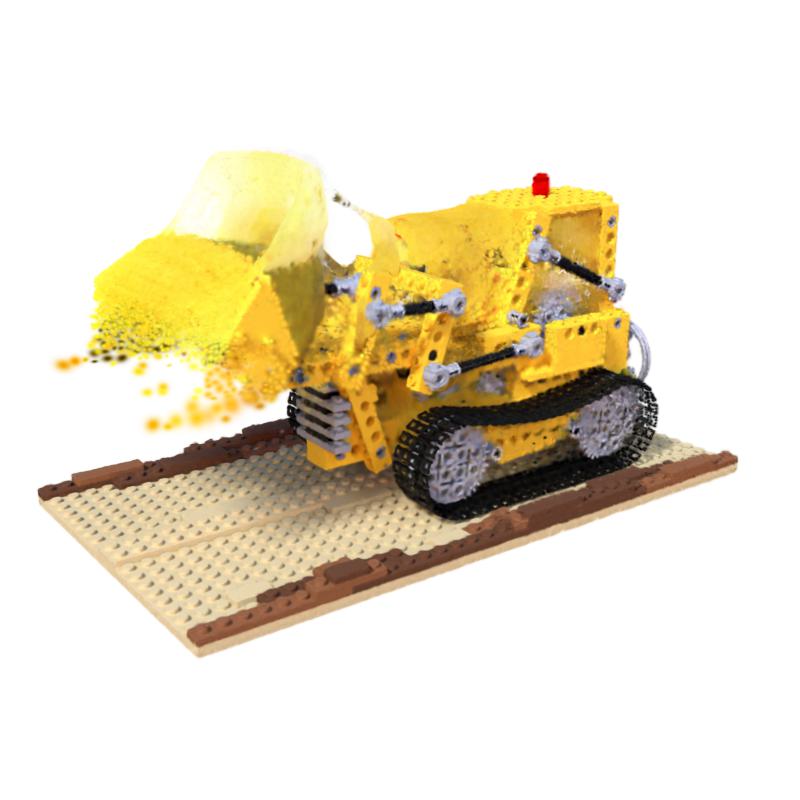}
    & \includegraphics[width=\hsize,valign=m]{figs/lego-pc-t1.jpg}
    & \includegraphics[width=\hsize,valign=m]{figs/lego-rgb-t1.jpg}
    & \includegraphics[width=\hsize,valign=m]{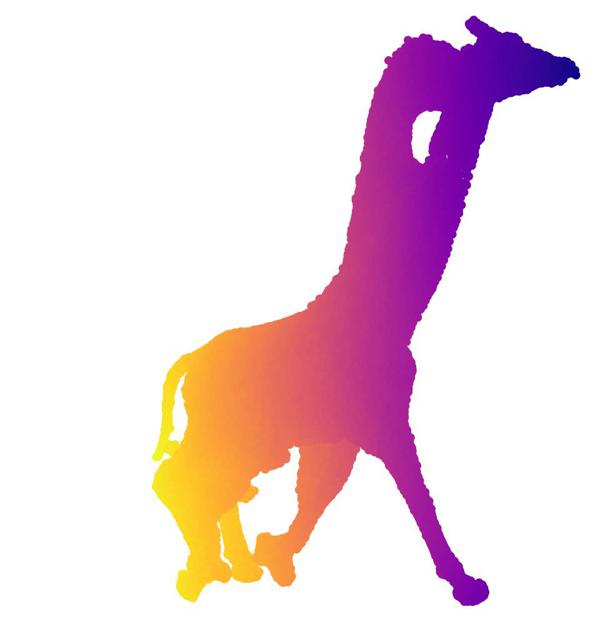}
    & \includegraphics[width=\hsize,valign=m]{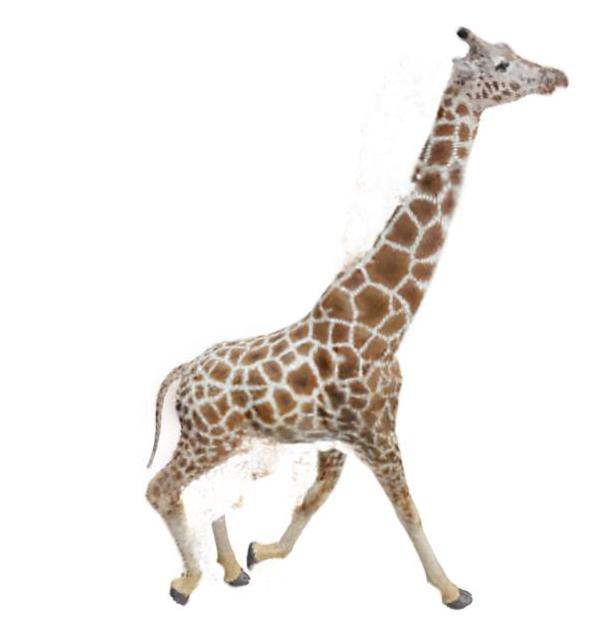}
    & \includegraphics[width=\hsize,valign=m]{figs/giraffe-pc-t1.jpg}
    & \includegraphics[width=\hsize,valign=m]{figs/giraffe-rgb-t1.jpg}
    \\
    & \includegraphics[width=\hsize,valign=m]{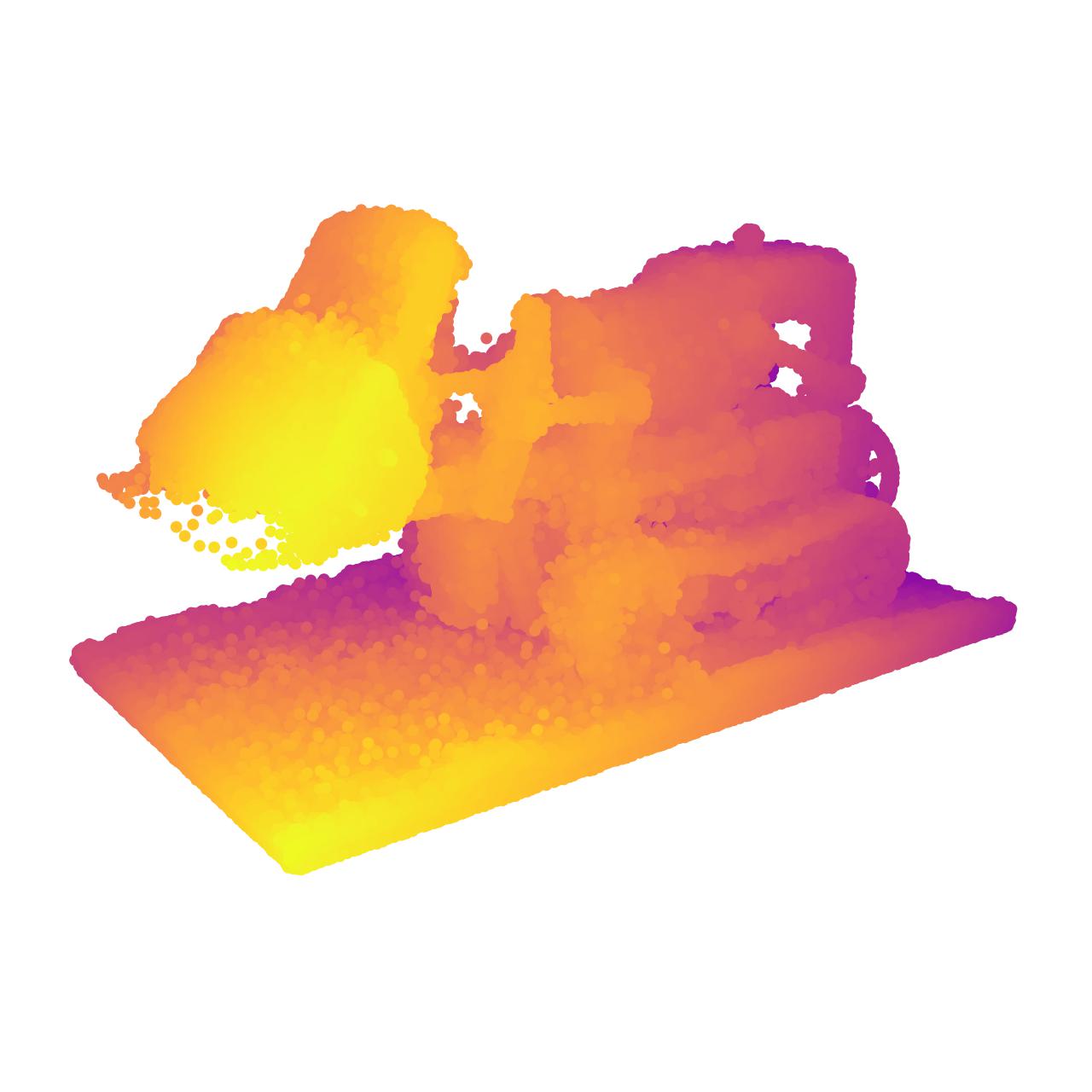}
    & \includegraphics[width=\hsize,valign=m]{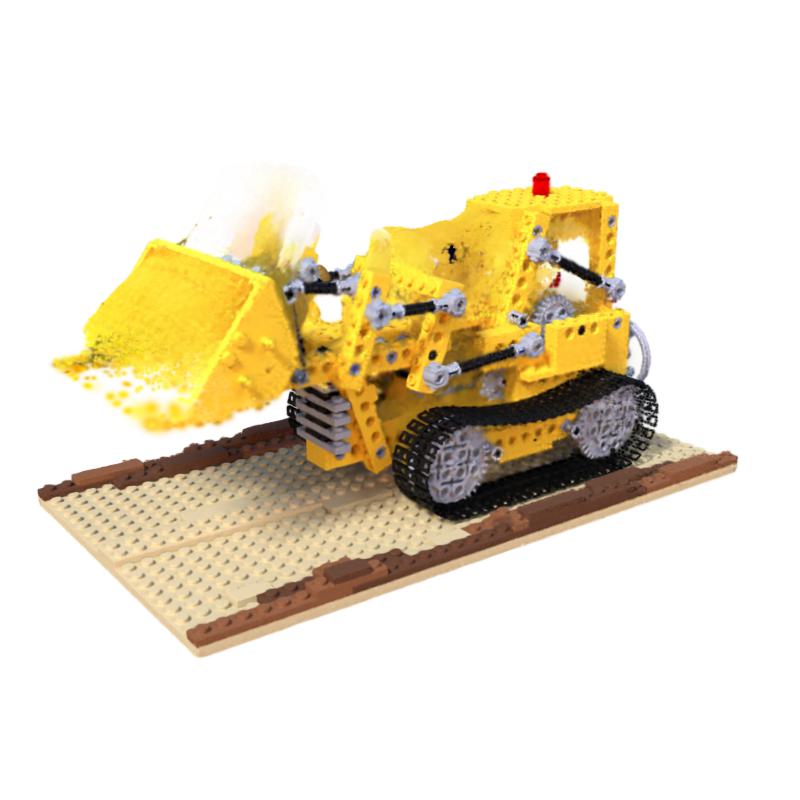}
    & \includegraphics[width=\hsize,valign=m]{figs/lego-pc-t2.jpg}
    & \includegraphics[width=\hsize,valign=m]{figs/lego-rgb-t2.jpg}
    & \includegraphics[width=\hsize,valign=m]{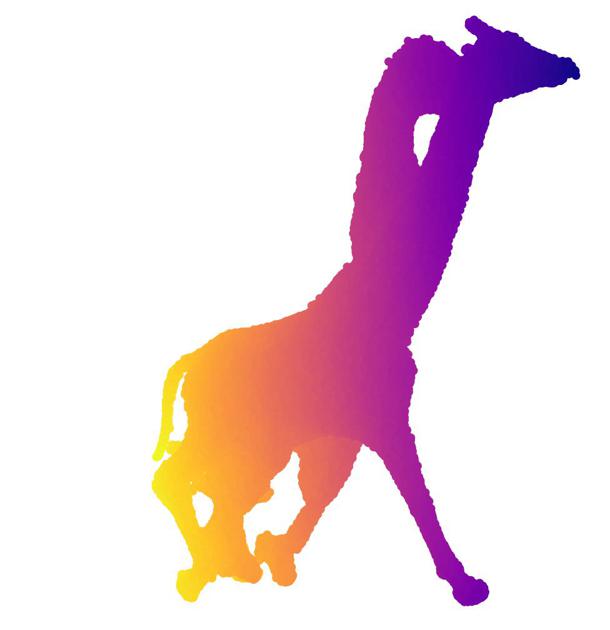}
    & \includegraphics[width=\hsize,valign=m]{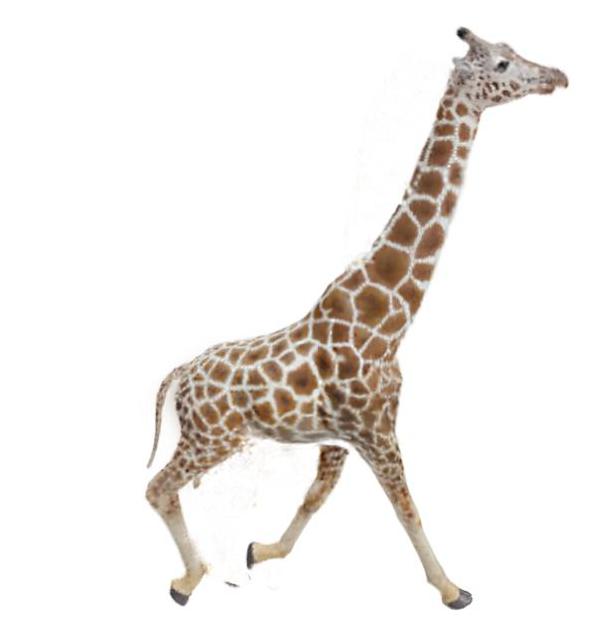}
    & \includegraphics[width=\hsize,valign=m]{figs/giraffe-pc-t2.jpg}
    & \includegraphics[width=\hsize,valign=m]{figs/giraffe-rgb-t2.jpg}
    \\

    \rotatebox[origin=c]{90}{End}
    & \includegraphics[width=\hsize,valign=m]{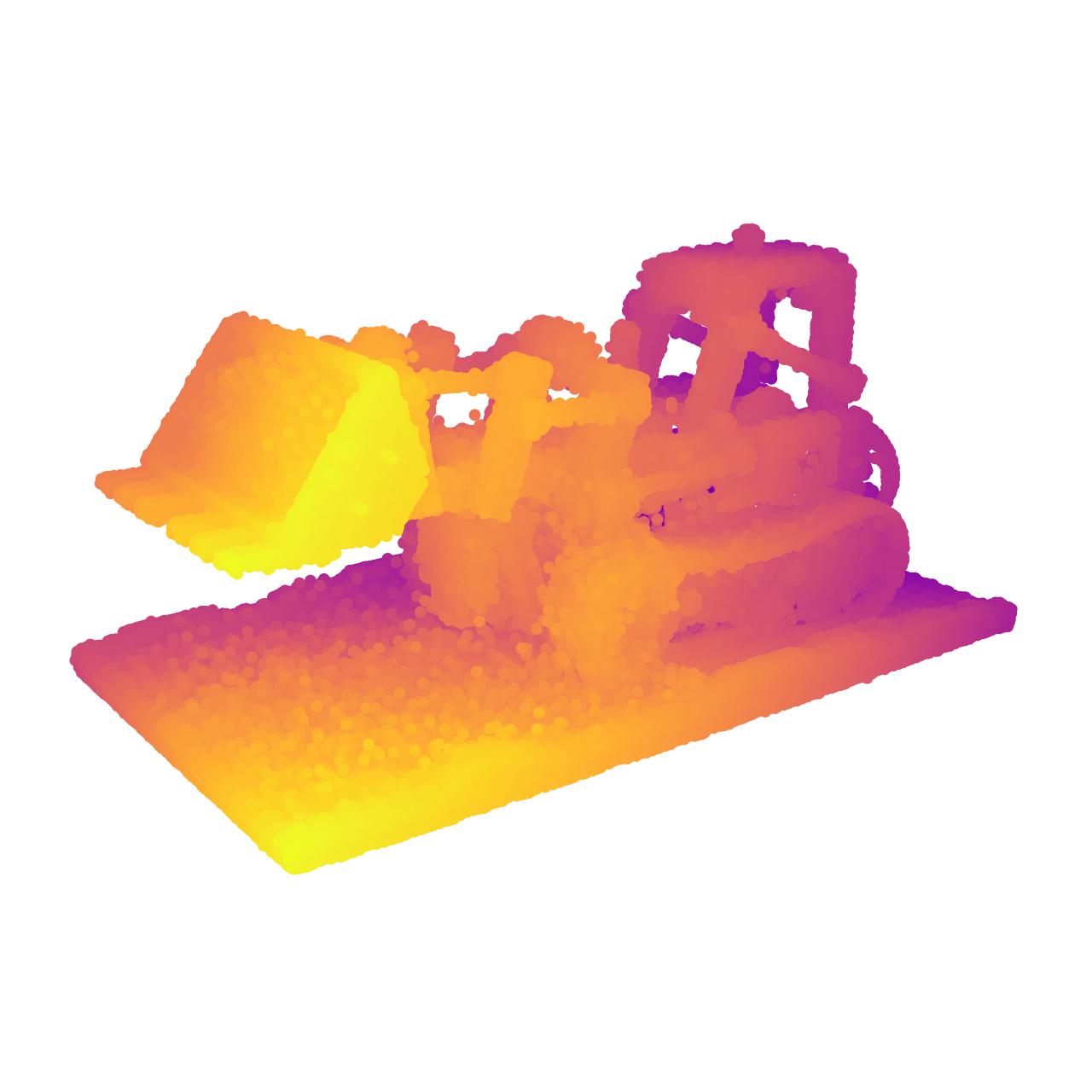}
    & \includegraphics[width=\hsize,valign=m]{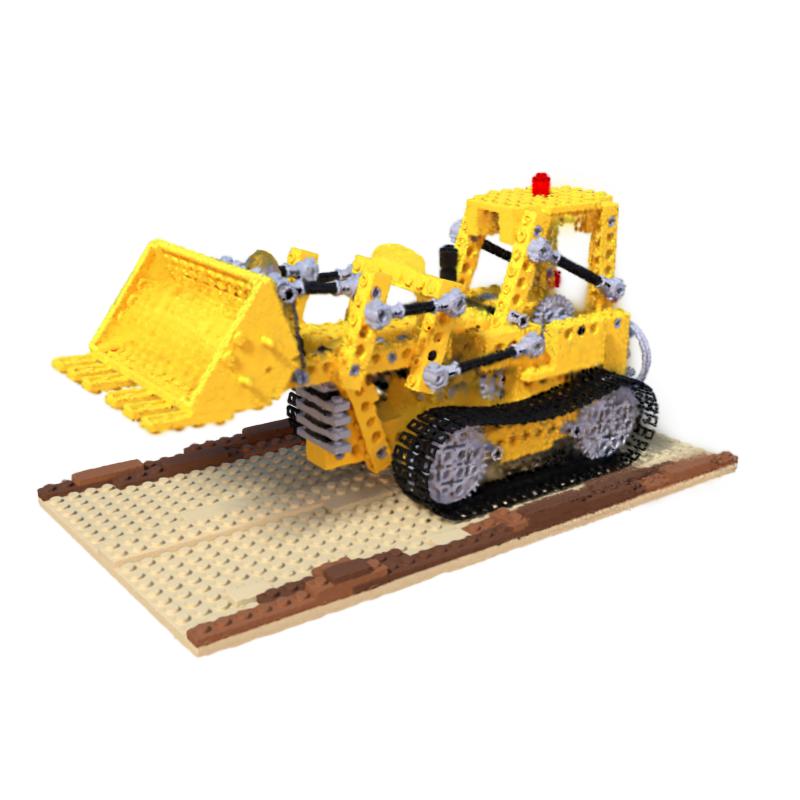}
    & \includegraphics[width=\hsize,valign=m]{figs/lego-pc-end.jpg}
    & \includegraphics[width=\hsize,valign=m]{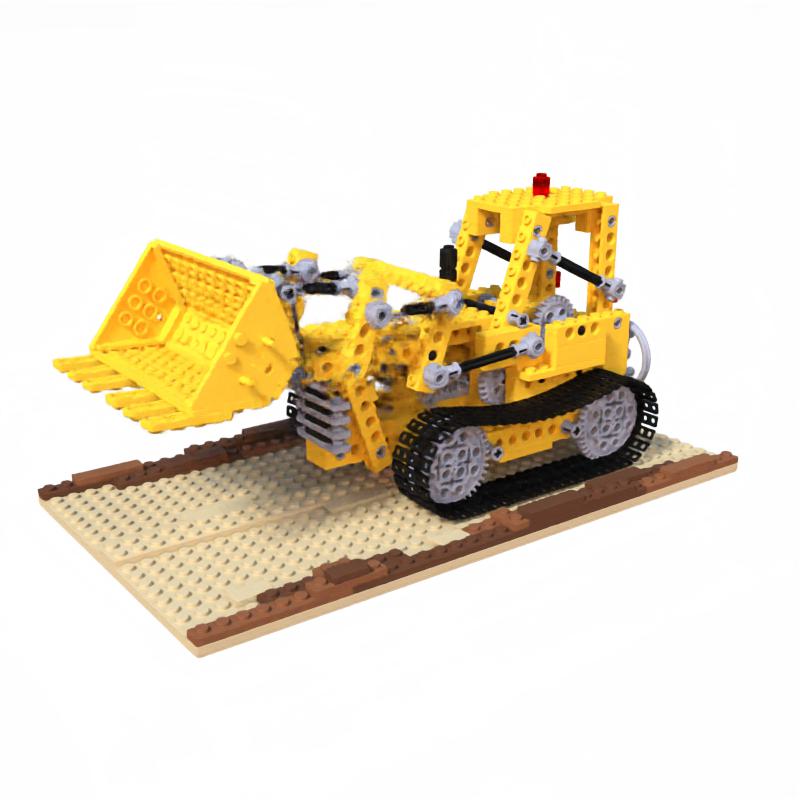}
    & \includegraphics[width=\hsize,valign=m]{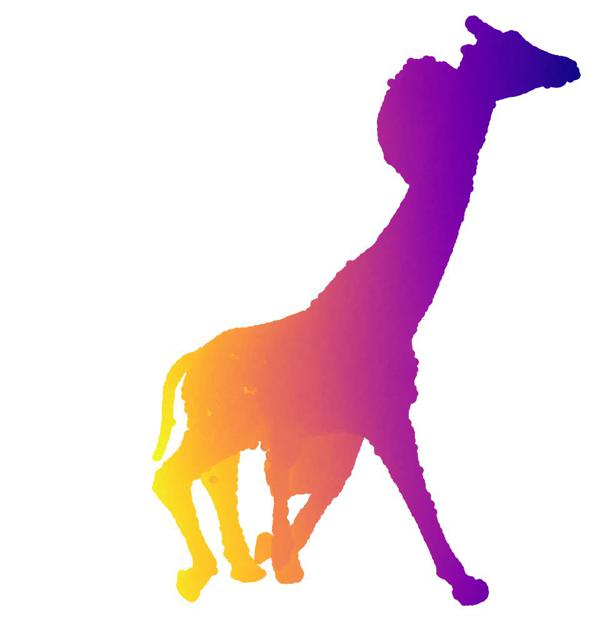}
    & \includegraphics[width=\hsize,valign=m]{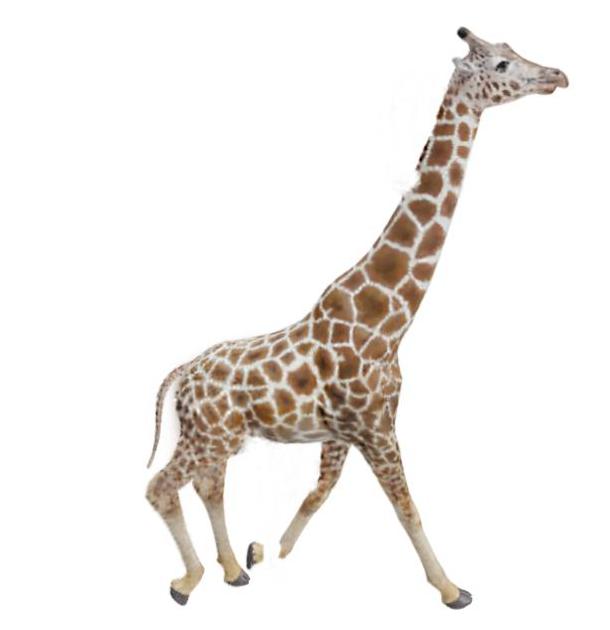}
    & \includegraphics[width=\hsize,valign=m]{figs/ab_giraffe_pc_end.jpg}
    & \includegraphics[width=\hsize,valign=m]{figs/giraffe-rgb-end.jpg}
    \\
    \rotatebox[origin=c]{90}{Trajectory}
    & \multicolumn{2}{c}{\includegraphics[width=0.14\linewidth,valign=m]{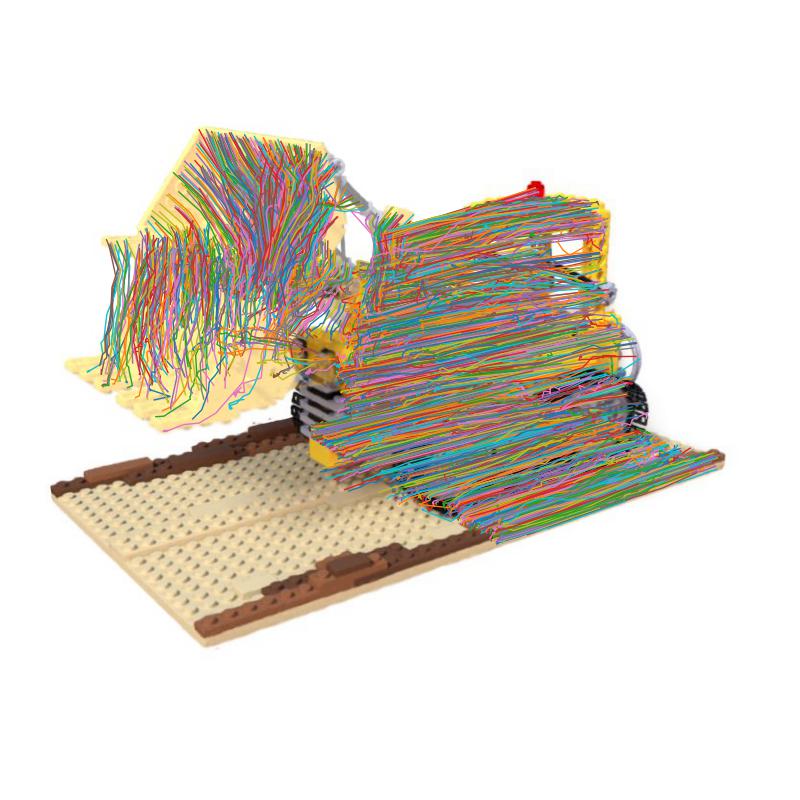}}
    & \multicolumn{2}{c}{\includegraphics[width=0.14\linewidth,valign=m]{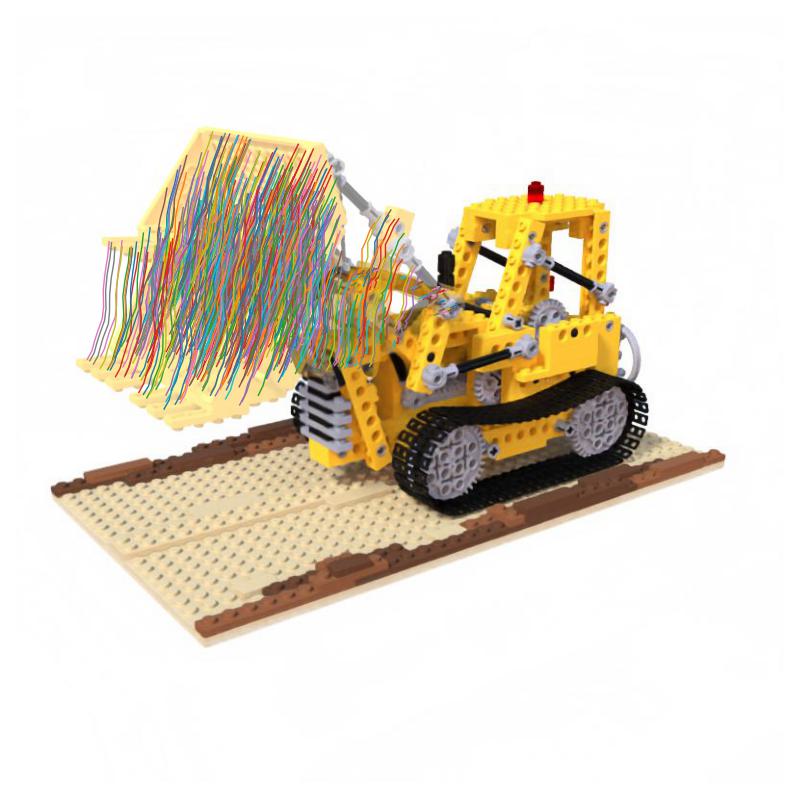}} 
    & \multicolumn{2}{c}{\includegraphics[width=0.1\linewidth,valign=m]{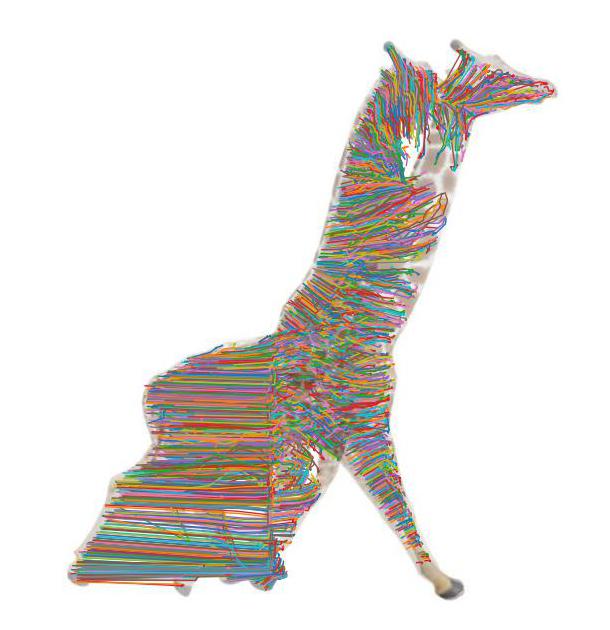}}
    & \multicolumn{2}{c}{\includegraphics[width=0.1\linewidth,valign=m]{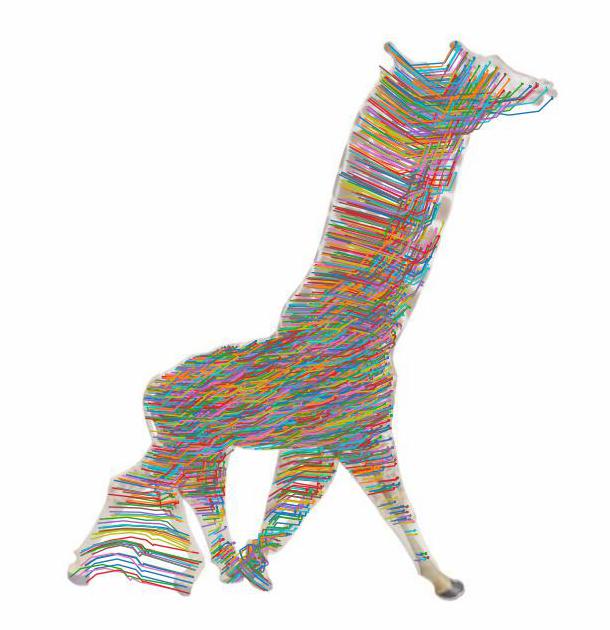}} 
\end{tabularx}
\caption{    
Qualitative comparison of 3D scene interpolation from start to end state using synthetic scenes. Both methods start by training a static model for the start state and subsequently finetune it towards the end state, all without any intermediate supervision. As shown, our PAPR in Motion method generates more plausible interpolation between states.
}
\label{fig:qualitative_supp}
\end{figure*}

\begin{figure*}[t]
\footnotesize
\begin{tabularx}{\linewidth}{lYYYY|YYYY}
& \multicolumn{2}{c}{Dynamic Gaussian~\cite{Luiten2023Dynamic3G}} & \multicolumn{2}{c}{PAPR in Motion (Ours)} & \multicolumn{2}{c}{Dynamic Gaussian~\cite{Luiten2023Dynamic3G}} & \multicolumn{2}{c}{PAPR in Motion (Ours)} \\
& Point Cloud & Rendering & Point Cloud & Rendering & Point Cloud & Rendering & Point Cloud & Rendering \\
    \rotatebox[origin=c]{90}{Start} 
    & \includegraphics[width=\hsize,valign=m]{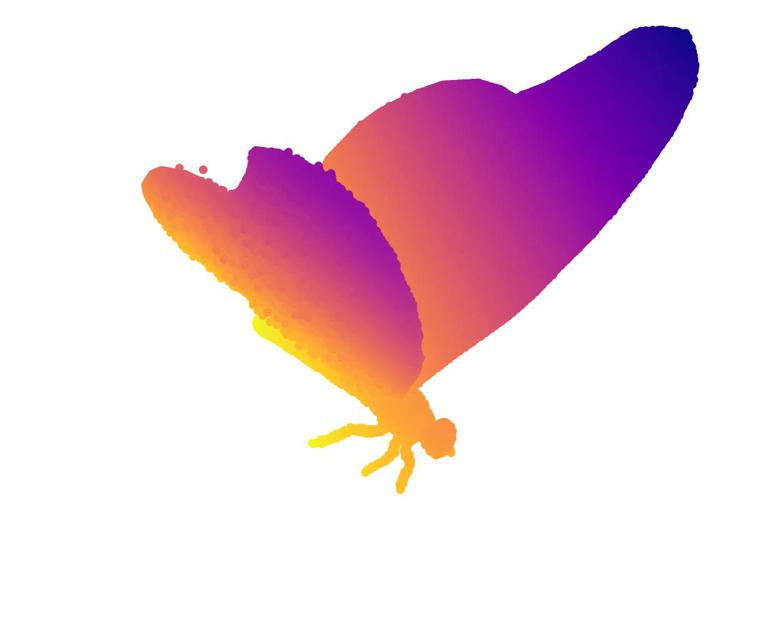}
    & \includegraphics[width=\hsize,valign=m]{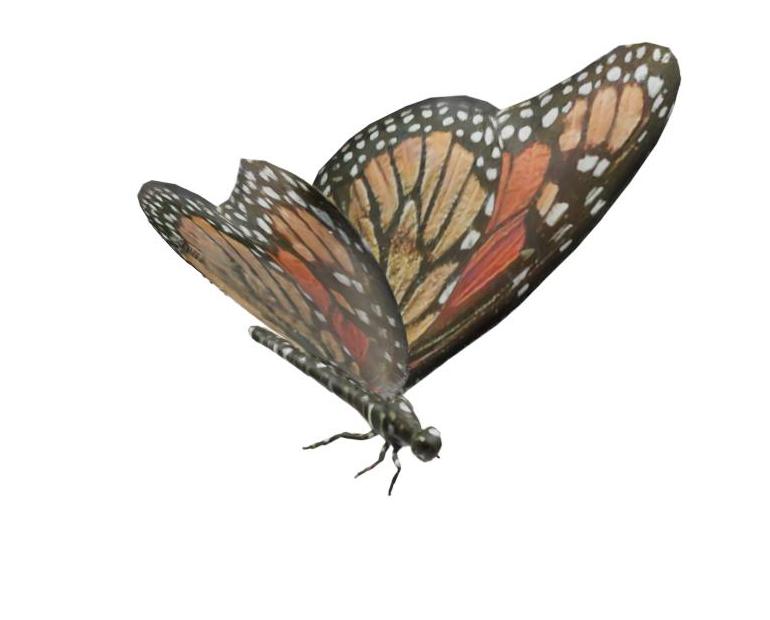}
    & \includegraphics[width=\hsize,valign=m]{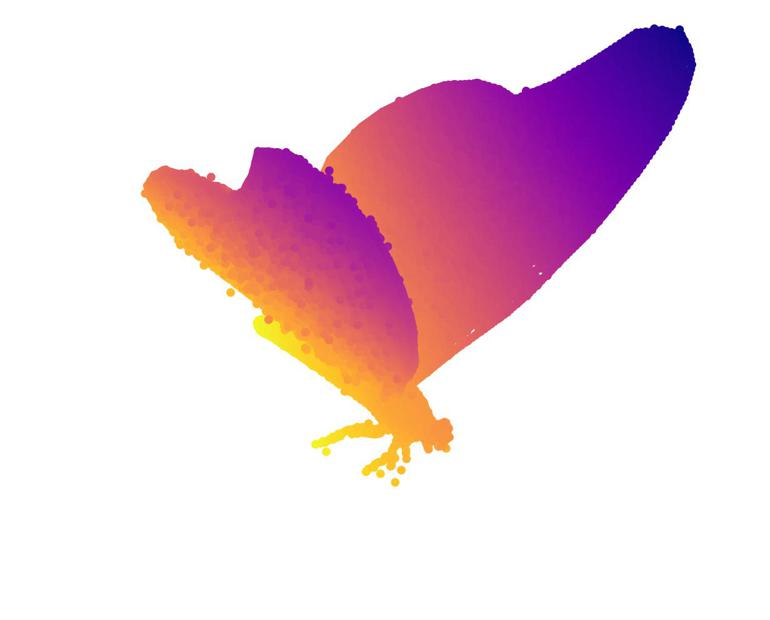}
    & \includegraphics[width=\hsize,valign=m]{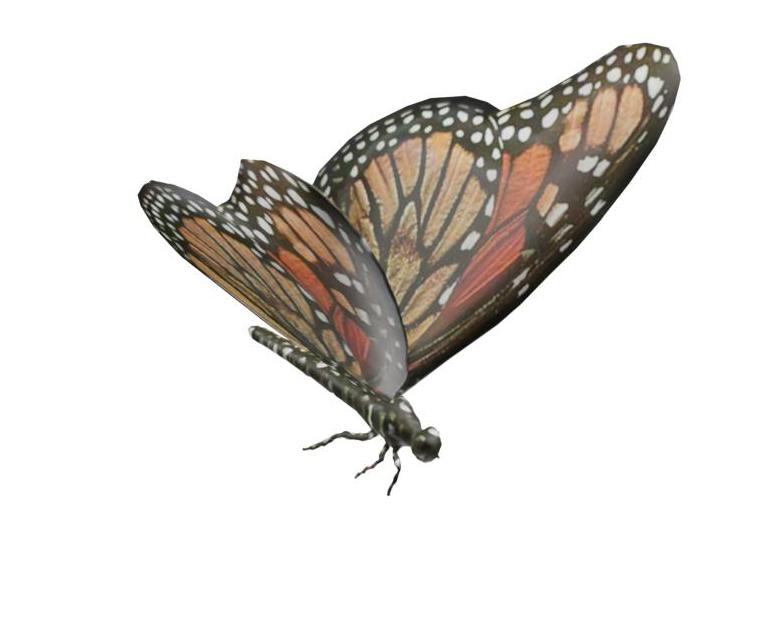}
    & \includegraphics[width=\hsize,valign=m]{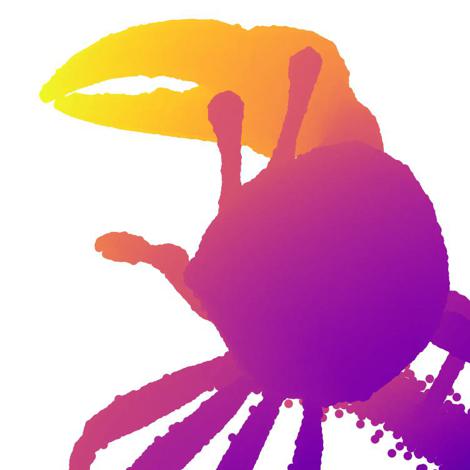}
    & \includegraphics[width=\hsize,valign=m]{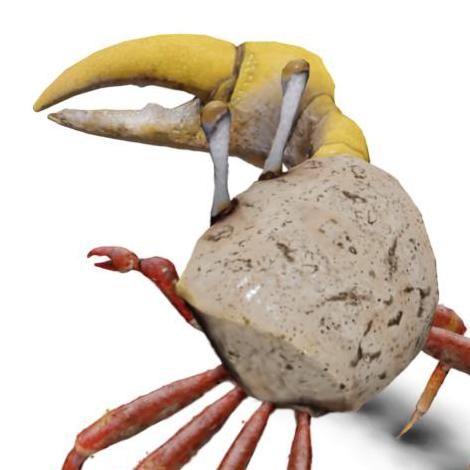}
    & \includegraphics[width=\hsize,valign=m]{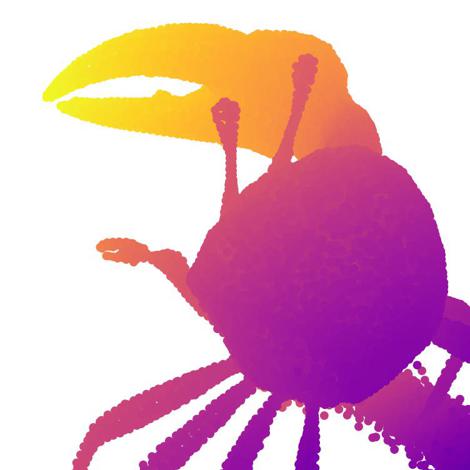}
    & \includegraphics[width=\hsize,valign=m]{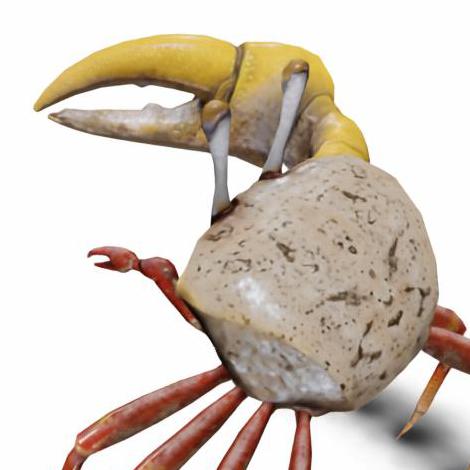}
    \\
    \multirow{2}{*}{\rotatebox[origin=c]{90}{Intermediate}}
    & \includegraphics[width=\hsize,valign=m]{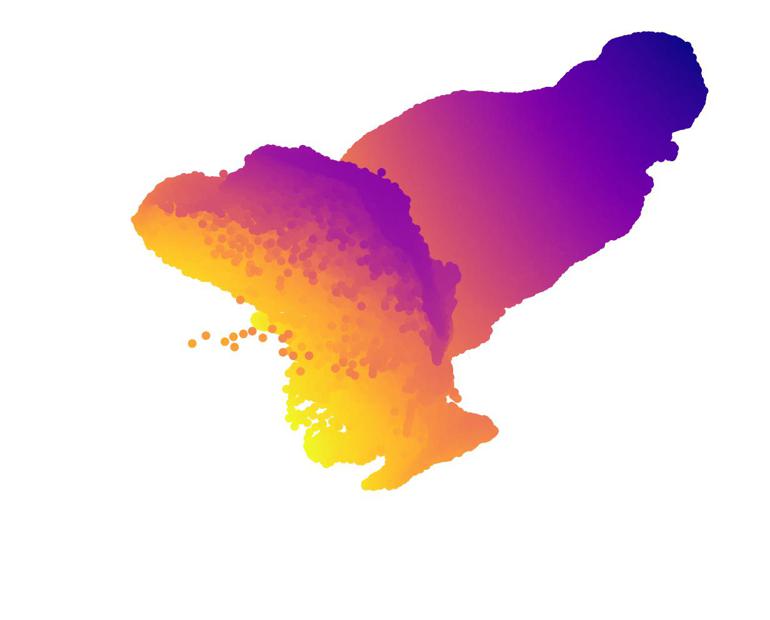}
    & \includegraphics[width=\hsize,valign=m]{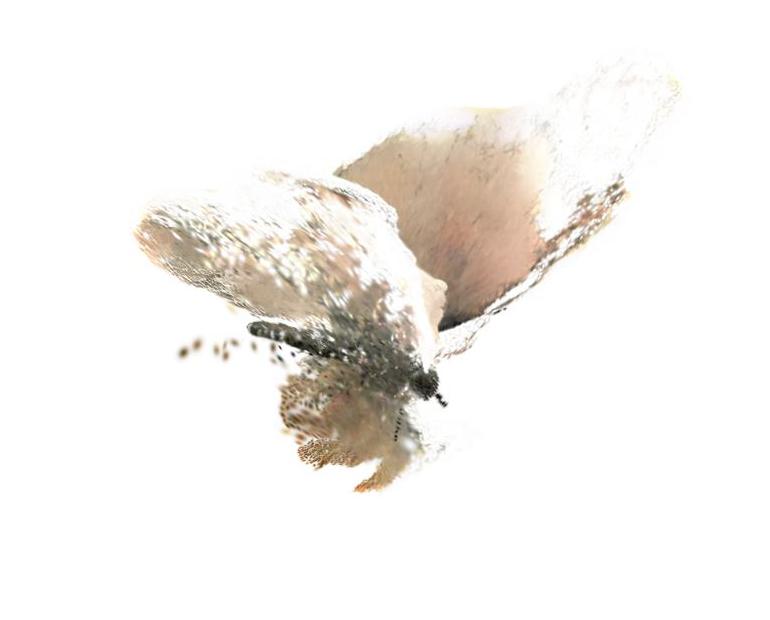}
    & \includegraphics[width=\hsize,valign=m]{figs/but-pc-t1.jpg}
    & \includegraphics[width=\hsize,valign=m]{figs/but-rgb-t1.jpg}
    & \includegraphics[width=\hsize,valign=m]{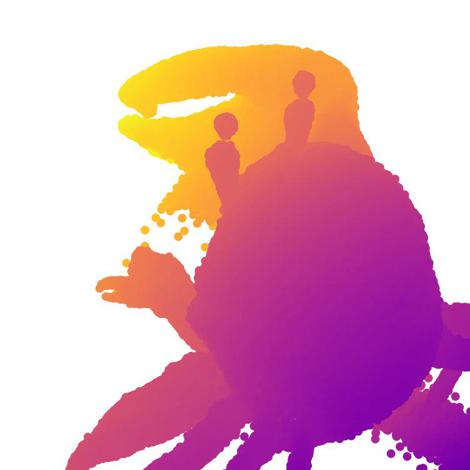}
    & \includegraphics[width=\hsize,valign=m]{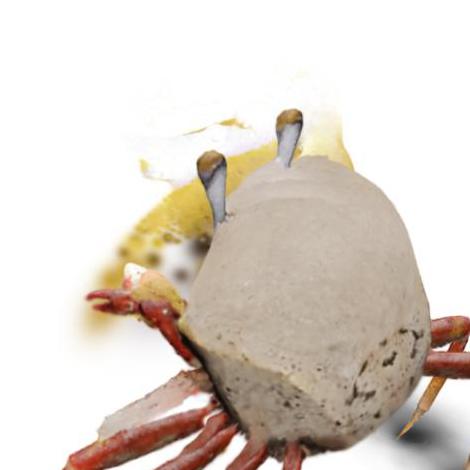}
    & \includegraphics[width=\hsize,valign=m]{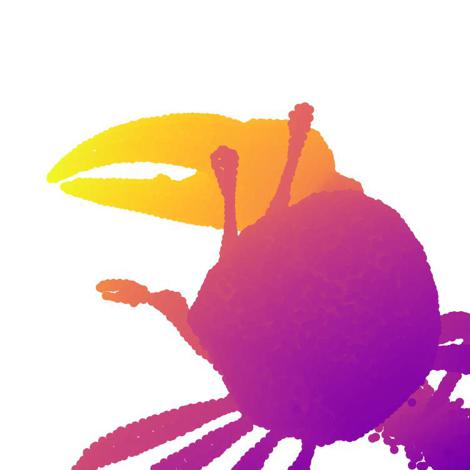}
    & \includegraphics[width=\hsize,valign=m]{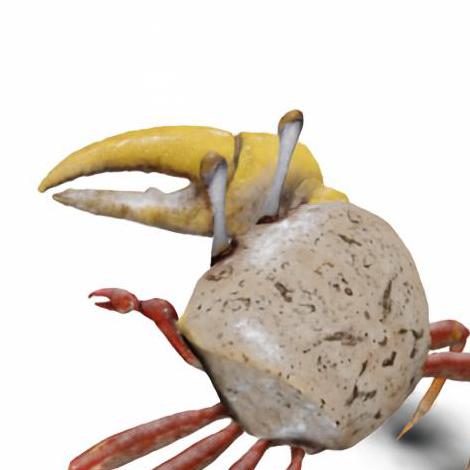}
    \\
    & \includegraphics[width=\hsize,valign=m]{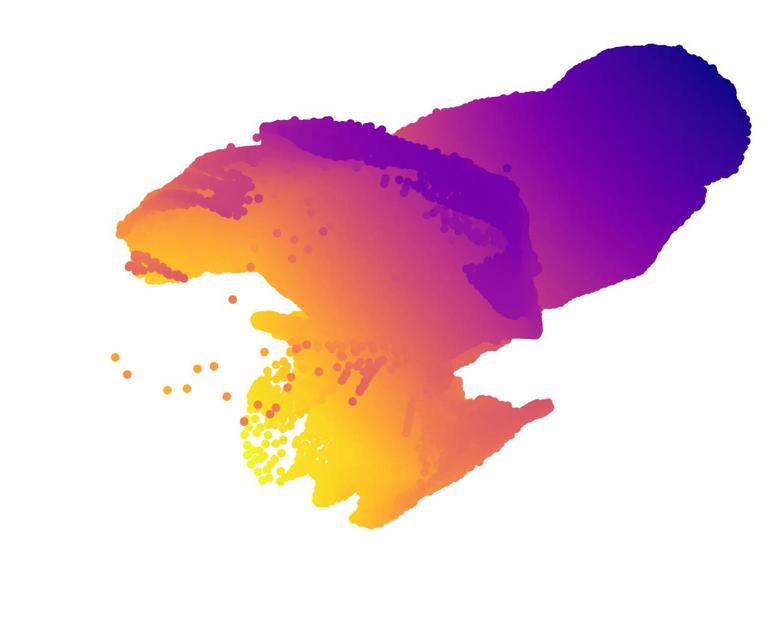}
    & \includegraphics[width=\hsize,valign=m]{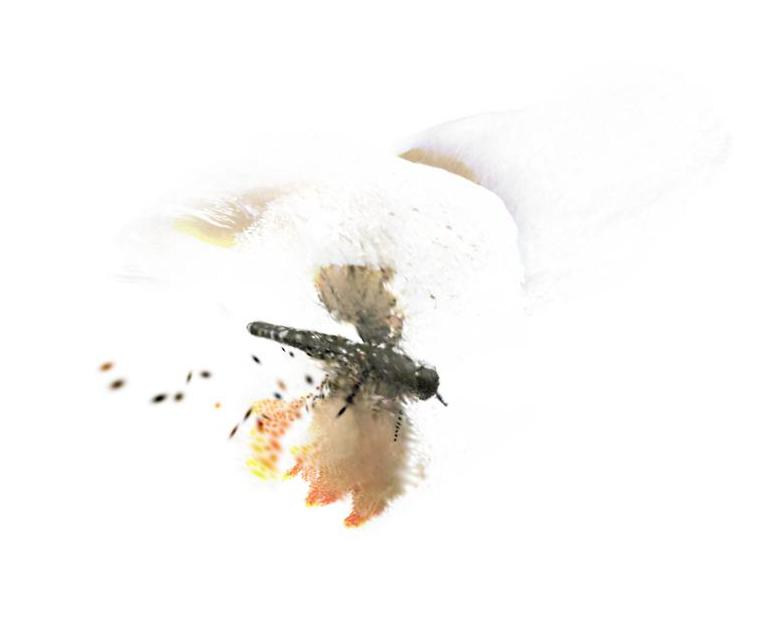}
    & \includegraphics[width=\hsize,valign=m]{figs/but-pc-t2.jpg}
    & \includegraphics[width=\hsize,valign=m]{figs/but-rgb-t2.jpg}
    & \includegraphics[width=\hsize,valign=m]{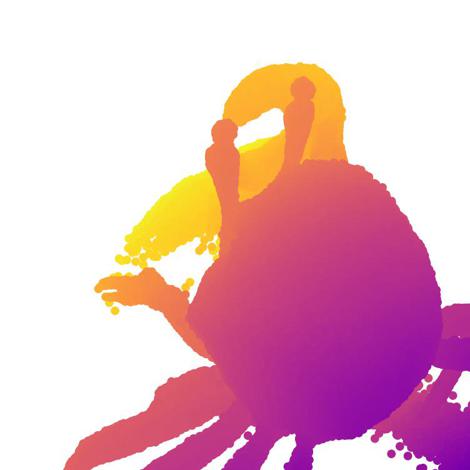}
    & \includegraphics[width=\hsize,valign=m]{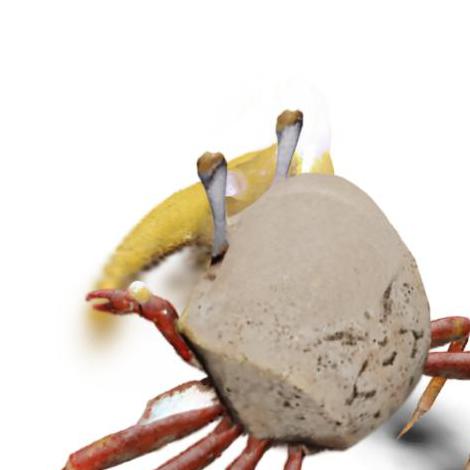}
    & \includegraphics[width=\hsize,valign=m]{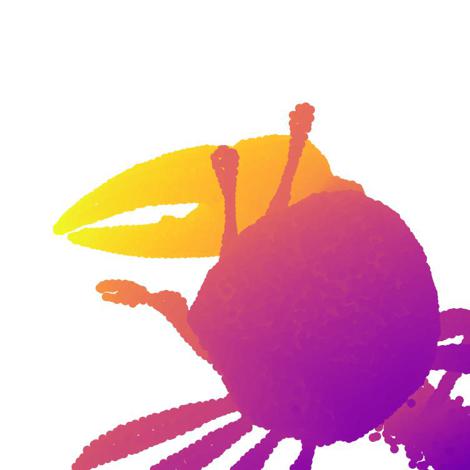}
    & \includegraphics[width=\hsize,valign=m]{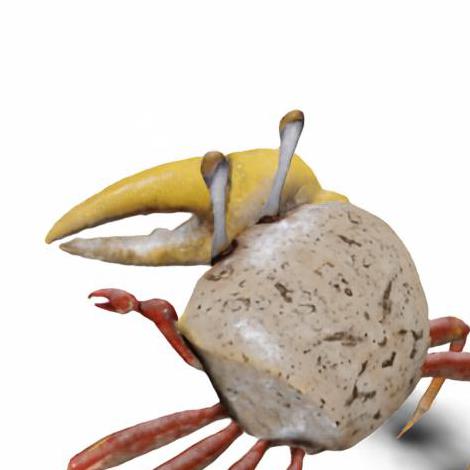}
    \\

    \rotatebox[origin=c]{90}{End}
    & \includegraphics[width=\hsize,valign=m]{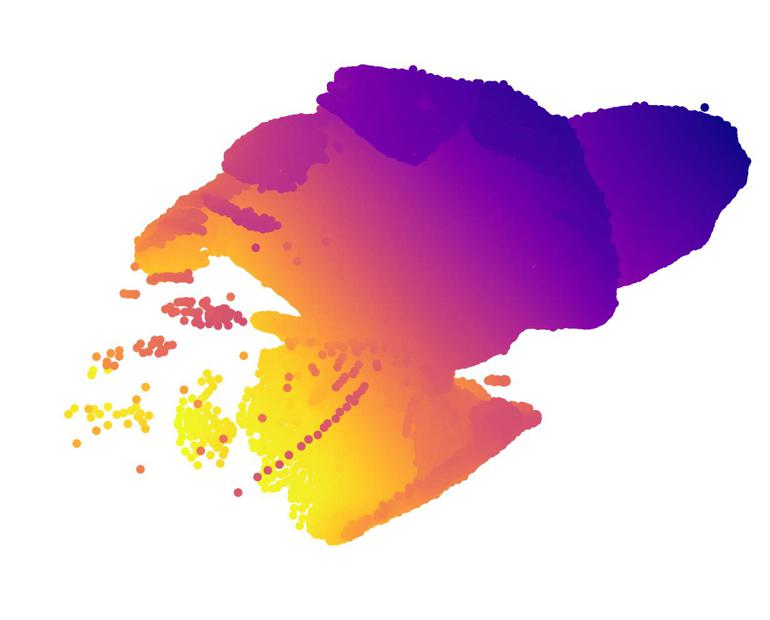}
    & \includegraphics[width=\hsize,valign=m]{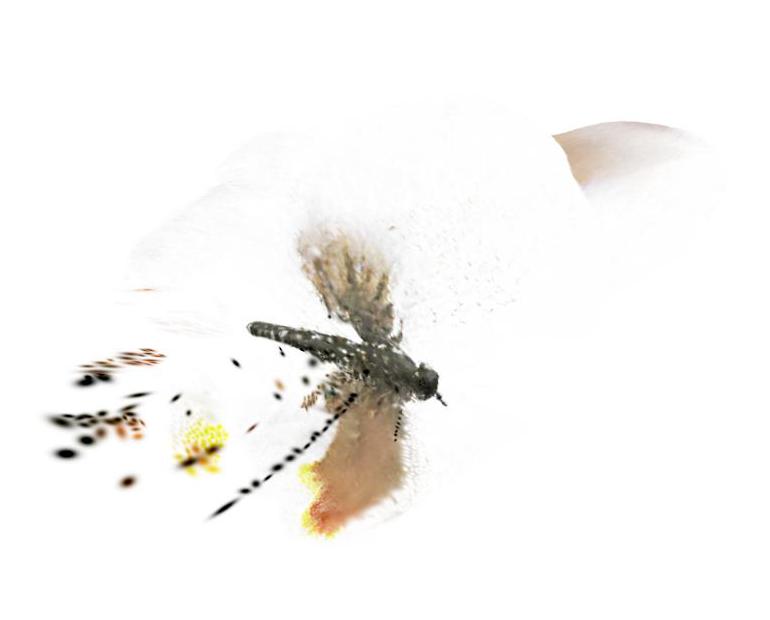}
    & \includegraphics[width=\hsize,valign=m]{figs/but-pc-end.jpg}
    & \includegraphics[width=\hsize,valign=m]{figs/but-rgb-end.jpeg}
    & \includegraphics[width=\hsize,valign=m]{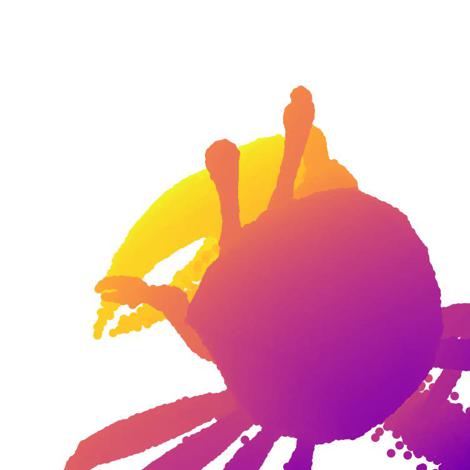}
    & \includegraphics[width=\hsize,valign=m]{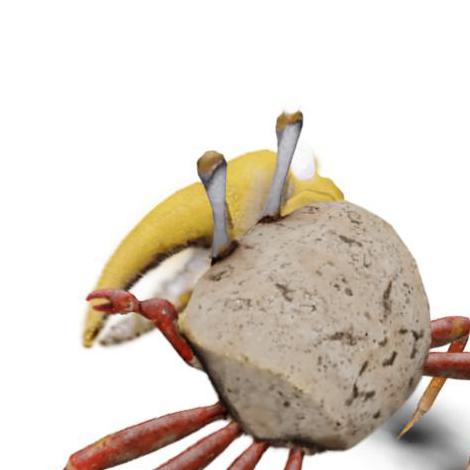}
    & \includegraphics[width=\hsize,valign=m]{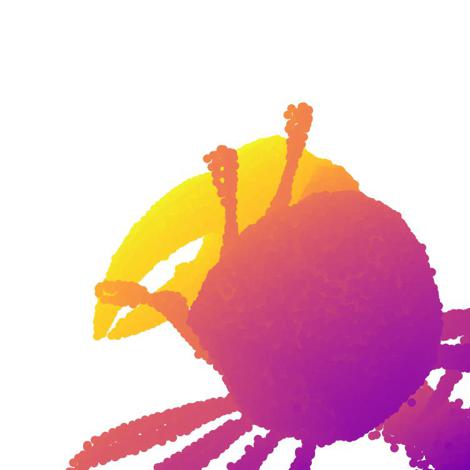}
    & \includegraphics[width=\hsize,valign=m]{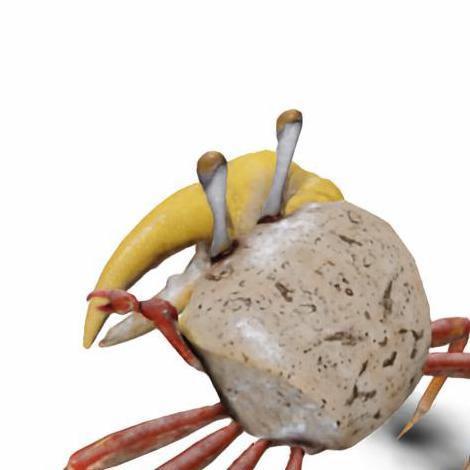}
    \\
    \rotatebox[origin=c]{90}{Trajectory}
    & \multicolumn{2}{c}{\includegraphics[width=0.14\linewidth,valign=m]{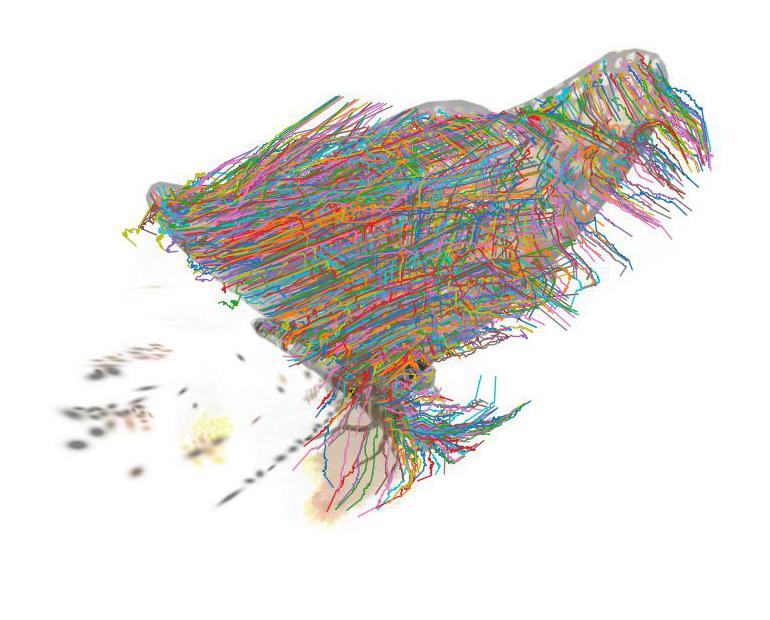}}
    & \multicolumn{2}{c}{\includegraphics[width=0.14\linewidth,valign=m]{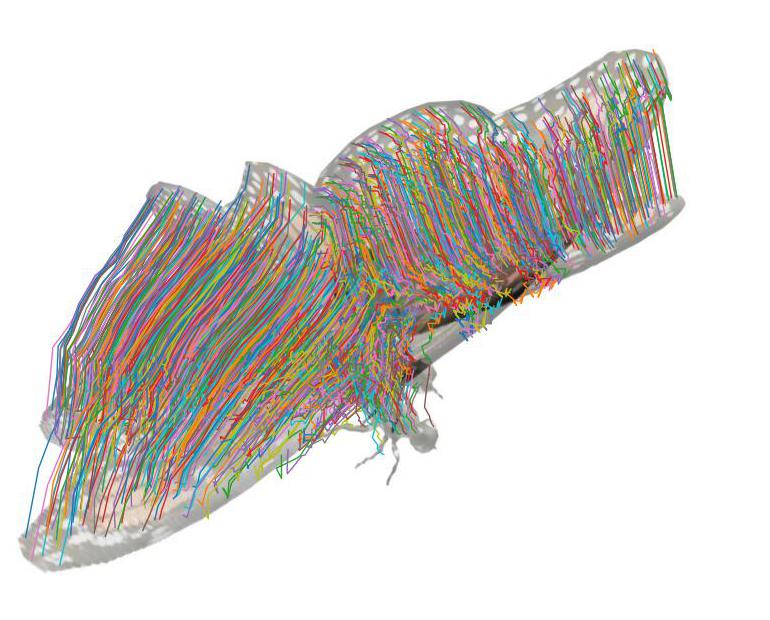}} 
    & \multicolumn{2}{c}{\includegraphics[width=0.1\linewidth,valign=m]{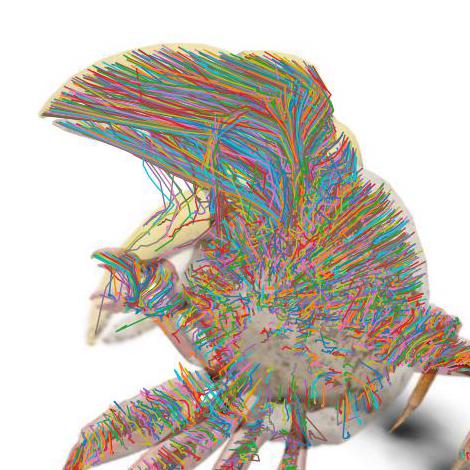}}
    & \multicolumn{2}{c}{\includegraphics[width=0.1\linewidth,valign=m]{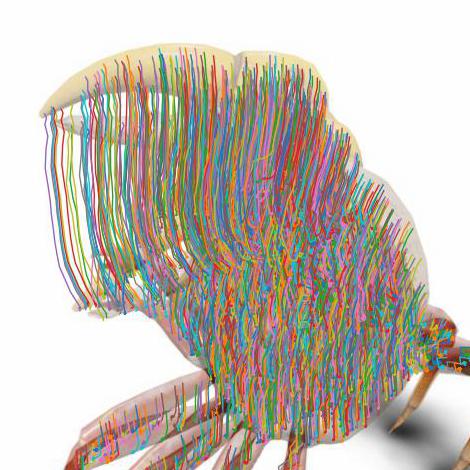}} 
    
\end{tabularx}
\caption{    
Qualitative comparison of 3D scene interpolation from start to end state using synthetic scenes. Both methods start by training a static model for the start state and subsequently finetune it towards the end state, all without any intermediate supervision. Dynamic Gaussian~\cite{Luiten2023Dynamic3G} fails to handle scene changes with large displacements, as shown by the butterfly example where the wings disappear, and in the crab scene, where the claw's geometry distorts during transition and fails to retain appearance details on the crab's shell. In contrast, PAPR in Motion produces smooth interpolations between states.
}
\label{fig:qualitative_supp_1}
\end{figure*}

Figure~\ref{fig:qualitative_supp} and \ref{fig:qualitative_supp_1} shows additional qualitative comparisons between our method, PAPR in Motion, and Dynamic Gaussian~\cite{Luiten2023Dynamic3G}. The results show that Dynamic Gaussian~\cite{Luiten2023Dynamic3G} struggles with maintaining object geometry integrity during the interpolation process. For instance, in the Lego Bulldozer scene, points on the arm notably drift, and points from the back of the cabin erroneously travel to the front. Similarly, in the giraffe scene, a portion of the points on the giraffe's neck and legs do not move cohesively, leading to disjointed transitions. In scenes with drastic changes like the butterfly and crab scenes, the baseline fails to preserve the original appearance.
 In contrast, PAPR in Motion successfully handles these challenging scenarios, producing smooth and natural interpolations between states.

\end{document}